\documentclass[10pt,twocolumn,letterpaper]{article}

\usepackage[pagenumbers]{cvpr} % To force page numbers, e.g. for an arXiv version

\usepackage{graphicx}
\usepackage{amsmath}
\usepackage{bm}
\usepackage{amssymb}
\usepackage{booktabs}
\usepackage[etex=true,export]{adjustbox}

\usepackage{graphicx, amsmath, amssymb, caption, subcaption, multirow, overpic, textpos}
\usepackage[table]{xcolor}
\usepackage{physics}
\definecolor{citecolor}{HTML}{0071BC}
\usepackage[pagebackref,breaklinks,colorlinks,citecolor=citecolor]{hyperref}

\definecolor{linkcolor}{HTML}{ED1C24}
\definecolor{my_blue}{HTML}{00B0F0}
\definecolor{my_purple}{HTML}{7030A0}
\definecolor{my_light}{HTML}{CAB2D6}

\newtheorem{lemma}{Lemma}
\newtheorem{proof}{Proof}
\newtheorem{definition}{Definition}

\newcommand{\y}{\mathbf{y}}
\newcommand{\z}{\mathbf{z}}
\newcommand{\f}{\mathbf{f}}

\newcommand{\w}{\mathbf{w}}

\newcommand{\W}{{\mathbf{W}}}
\newcommand{\U}{{\mathbf{U}}}
\newcommand{\M}{{\mathbf{M}}}

\newcommand{\ours}{CeCo}

\makeatletter
\newcommand*\bigcdot{\mathpalette\bigcdot@{.5}}
\newcommand*\bigcdot@[2]{\mathbin{\vcenter{\hbox{\scalebox{#2}{$\m@th#1\bullet$}}}}}
\makeatother

\usepackage[capitalize]{cleveref}
\crefname{section}{Sec.}{Secs.}
\Crefname{section}{Section}{Sections}
\Crefname{table}{Table}{Tables}
\crefname{table}{Tab.}{Tabs.}

\newlength\savewidth\newcommand\shline{\noalign{\global\savewidth\arrayrulewidth
  \global\arrayrulewidth 1pt}\hline\noalign{\global\arrayrulewidth\savewidth}}
\newcommand{\tablestyle}[2]{\setlength{\tabcolsep}{#1}\renewcommand{\arraystretch}{#2}\centering\footnotesize}
\renewcommand{\paragraph}[1]{\vspace{1.25mm}\noindent\textbf{#1}}

\newcolumntype{x}[1]{>{\centering\arraybackslash}p{#1pt}}
\newcolumntype{y}[1]{>{\raggedright\arraybackslash}p{#1pt}}
\newcolumntype{z}[1]{>{\raggedleft\arraybackslash}p{#1pt}}

\newcommand{\app}{\raise.17ex\hbox{$\scriptstyle\sim$}}

\definecolor{deemph}{gray}{0.6}

\definecolor{baselinecolor}{gray}{.9}

\newcommand{\cgaphl}[2]{
	\fontsize{8pt}{1em}\selectfont{\textcolor{Highlight}{($\textbf{#1}$\textbf{#2})}}
}
\newcommand{\cgaphll}[2]{
	\fontsize{8pt}{1em}\selectfont{\textcolor{my_red}{($\textbf{#1}$\textbf{#2})}}
}
\definecolor{my_red}{HTML}{FE4444}
\definecolor{Highlight}{HTML}{39b54a}  % green
\definecolor{Gray}{gray}{0.95}

\def\cvprPaperID{3196} % *** Enter the CVPR Paper ID here
\def\confName{CVPR}
\def\confYear{2023}

\begin{document}

\title{Understanding Imbalanced Semantic Segmentation Through Neural Collapse}

\author{
	\hspace{-30pt} Zhisheng Zhong$^{1, *}$ \hspace{4pt} Jiequan Cui$^{1, *}$ \hspace{4pt} Yibo Yang$^{2, *}$ \hspace{4pt} Xiaoyang Wu$^{3}$ \hspace{4pt} Xiaojuan Qi$^{3}$ \hspace{4pt} Xiangyu Zhang$^{4}$ \hspace{4pt} Jiaya Jia$^{1}$ \vspace{.3em}
	\\
	\hspace{-30pt} CUHK$^{1}$ \quad\quad\quad JD Explore Academy$^{2}$ \quad\quad\quad HKU$^{3}$ \quad\quad\quad MEGVII Technology$^{4}$ \\
	\vspace{-10pt}
	{\hspace{-30pt} \small $^{*}$equal  contribution \quad\quad code: \url{https://github.com/dvlab-research/Imbalanced-Learning}}
	%{\hspace{-30pt} \small Project: \url{https://github.com/dvlab-research/Imbalanced-Learning}}
}

\maketitle

%\pagestyle{empty}
%\thispagestyle{empty}

%%%%%%%%% ABSTRACT
\begin{abstract}

A recent study has shown a phenomenon called neural collapse in that the within-class means of features and the classifier weight vectors converge to the vertices of a simplex equiangular tight frame at the terminal phase of training for classification. In this paper, we explore the corresponding structures of the last-layer feature centers and classifiers in semantic segmentation. Based on our empirical and theoretical analysis, we point out that semantic segmentation naturally brings contextual correlation and imbalanced distribution among classes, which breaks the equiangular and maximally separated structure of neural collapse for both feature centers and classifiers. However, such a symmetric structure is beneficial to discrimination for the minor classes. To preserve these advantages, we introduce a regularizer on feature centers to encourage the network to learn features closer to the appealing structure in imbalanced semantic segmentation. Experimental results show that our method can bring significant improvements on both 2D and 3D semantic segmentation benchmarks. {Moreover, our method ranks {$\mathbf{1}^{\rm st}$} and sets a new record (+\textbf{6.8\%} mIoU) on the ScanNet200 test leaderboard}.

\end{abstract}

%%%%%%%%% BODY TEXT
\section{Introduction}
\label{sec:intro}

%% 暂时别删掉
%% 增加下在imbalance setting等角分类器的好处

%In real-world implementations, available data is prone to severe imbalanced distribution. Despite the huge success of deep learning in a range of application areas \cite{lecun2015deep}, data imbalance has not been completely tractable in the training of a deep neural network. It would easily cause a catastrophic performance for classes with only a limited number of training samples \cite{he2009learning}. In visual recognition, a lot of re-sampling and re-weighting based methods have been proposed to induce balanced accuracy among classes \cite{huang2016learning,buda2018systematic}. However, in semantic segmentation, how does data imbalance affect its learning behaviors and corresponding solutions remain largely unexplored.  

\begin{figure}[t!] 
	\centering
	\begin{subfigure}[t]{0.45\linewidth}
		\centering
		\includegraphics[width=0.99\linewidth]{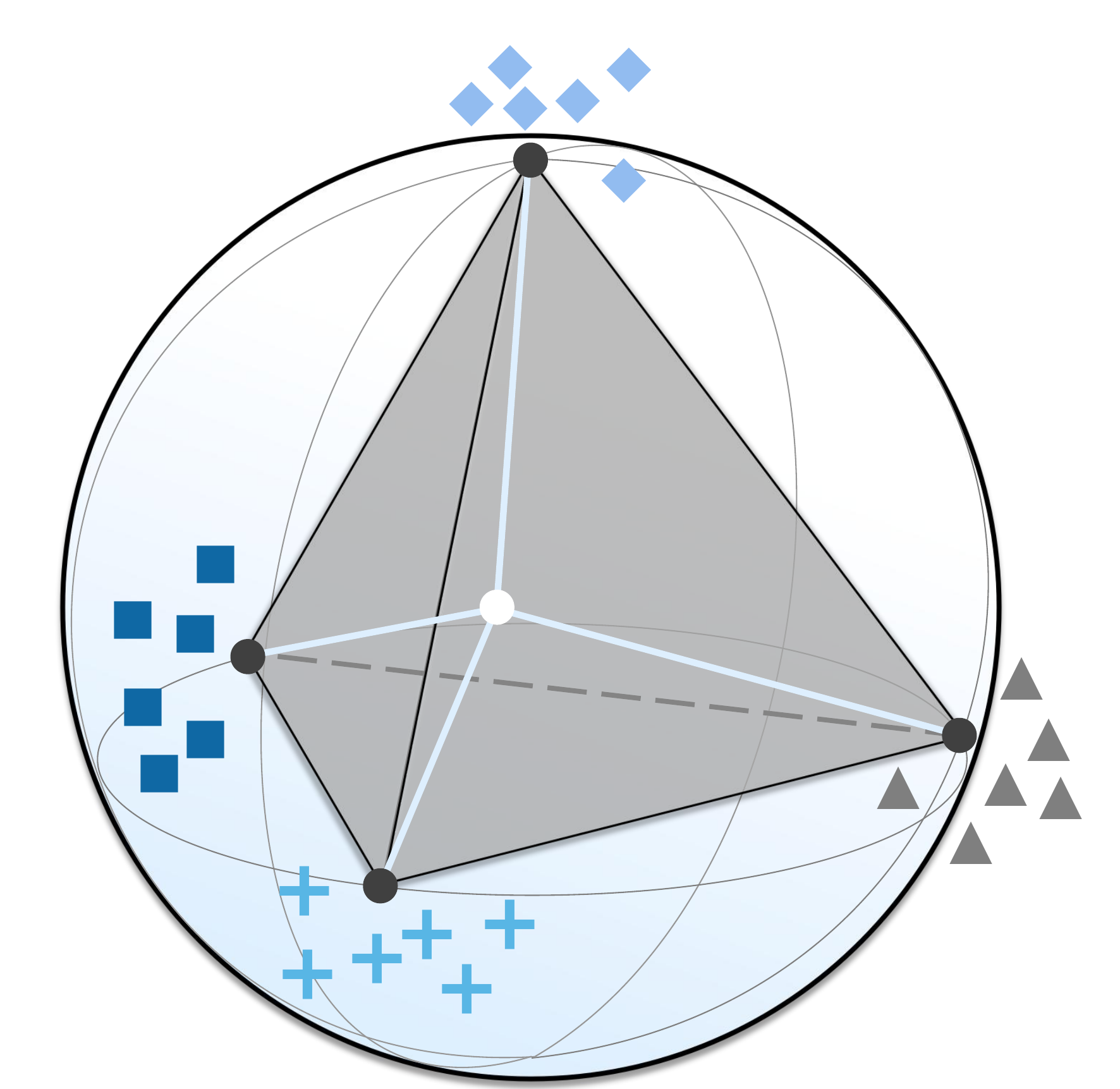}
		\caption{}
		\label{fig1_1}
	\end{subfigure}
	\hfill
	\begin{subfigure}[t]{0.45\linewidth}
		\centering
		\includegraphics[width=0.99\linewidth]{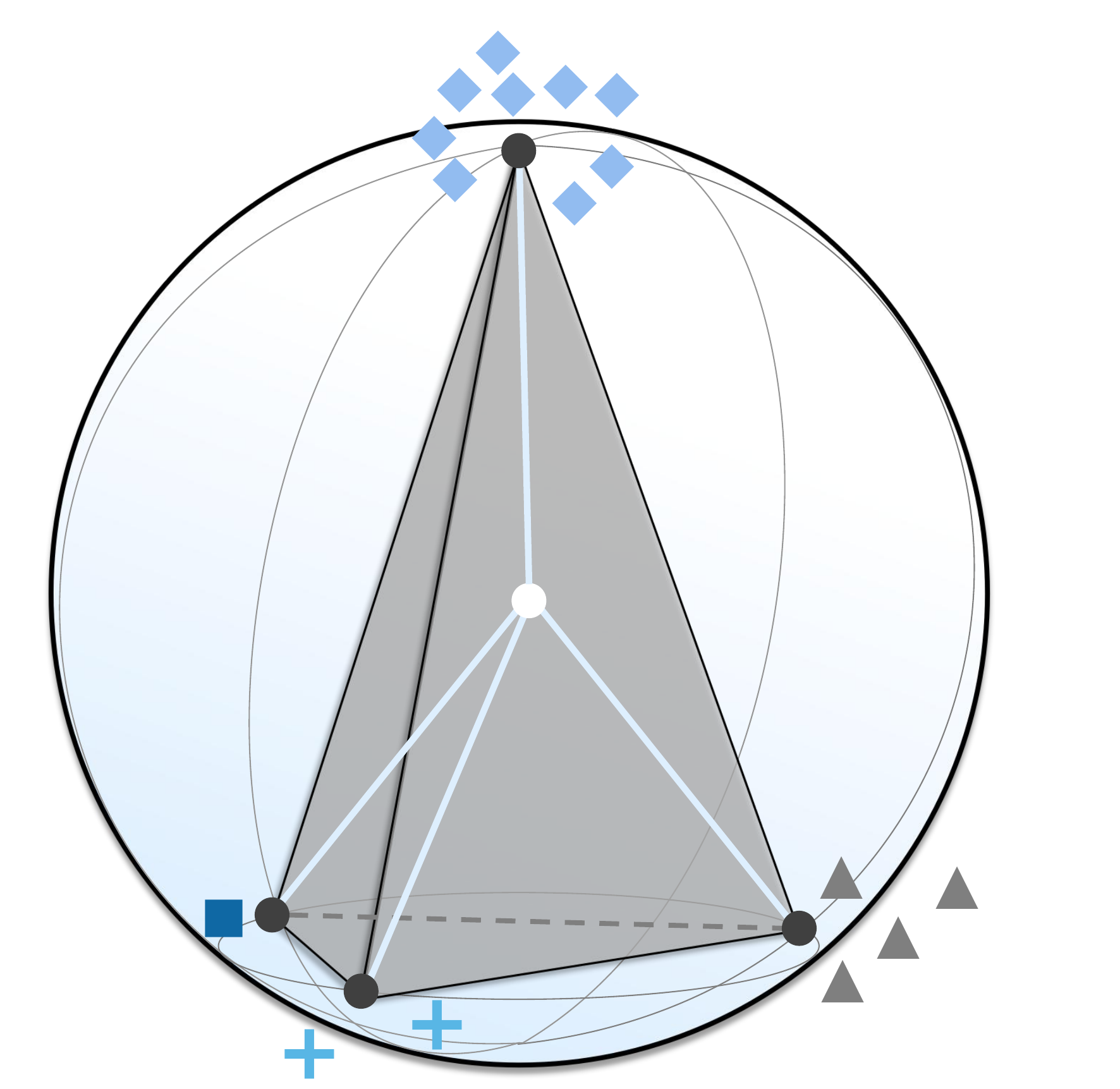}
		\caption{}
		\label{fig1_2}
	\end{subfigure}
	\hfill
	\caption{Illustration of equiangular separation (a) and non-equiangular separation (b) in a 3D space. Neural collapse reveals the structure in (a), where features are collapsed into their within-class centers with \textit{\textbf{maximal equiangular separation}} as a simplex ETF, and classifiers are aligned with the same structure. We observe that in semantic segmentation the feature centers and classifiers do not satisfy such a structure, as illustrated in (b) for an example. As some minor class features and classifier vectors lie in a close position, the discriminate ability of the network degrades.}
	\label{fig:intro}
\end{figure}

The solution structures of the last-layer representation and classifier provide a geometric perspective to delve into the learning behaviors in a deep neural network. 
%role of data imbalance. 
The neural collapse phenomenon discovered by Papyan et al. \cite{papyan2020prevalence} reveals that as a classification model is trained towards convergence on a balanced dataset,
%when training on a balanced dataset, as the model goes to convergence, 
the last-layer feature centers of all classes will be located on a hyper-sphere with maximal equiangular separation, as known as a simplex equiangular tight frame (simplex ETF), which means that any two centers have an equal cosine similarity, as shown in {Fig.~\ref{fig1_1}}. The final classifiers will be formed as the same structure and aligned with the feature centers. The following studies try to theoretically explain this elegant phenomenon, showing that neural collapse is the global optimality under the cross-entropy~(CE) and 
mean squared error~(MSE) loss functions in an approximated model \cite{weinan2020emergence, graf2021dissecting, lu2020neural, fang2021exploring, zhu2021geometric, ji2021unconstrained, mixon2020neural, poggio2020explicit, han2021neural, tirer2022extended}. 
%It is indicated that neural collapse will be broken when data imbalance emerges, which explains the deteriorated performance of training on imbalanced data \cite{fang2021exploring}. 
However, all the current studies on neural collapse focus on the training in image recognition, which performs classification for each image. 
%We notice that semantic segmentation naturally suffers from data imbalance because some semantic classes are prone to cover a large area with significantly more pixels. 
Semantic segmentation as an important pixel-wise classification problem receives no attention from the neural collapse perspective yet.

In this paper, we explore the solution structures of feature centers and classifiers in semantic segmentation. Surprisingly, it is observed that the symmetric equiangular separation as instructed by the neural collapse phenomenon in image recognition does not hold in semantic segmentation for both feature centers and classifiers. An example of non-equiangular separation is illustrated in Fig.~\ref{fig1_2}. We point out two reasons that may explain the difference. 

First, classification benchmark datasets usually have low correlation among classes. %\xjqi{what do you mean by low correspondence among classes?? low interferences?? correlations among different categories, I think correlation would be better than correspondence??}. 
%In contrast, semantic classes inevitably show much stronger dependencies %\xjqi{don't quite clear. Do you want to say: semantic segmentation is performed on scene-level images, different categories will have contextual dependencies and relationships, for example, the laptop is often on a desk??} %\xjqi{rewrite this sentence as: unlike image classification tasks that different categories xxx, different categories in the semantic segmentation task are contextually related}. 
In contrast, different classes in the semantic segmentation task are contextually related. In this case, the classifier needs to be adaptable to class correlation, so does not necessarily equally separate the label space. We conduct a simple experiment to verify it:
\begin{center}\vspace{-.2em}
\tablestyle{4pt}{1.05}
\begin{tabular}{y{60}y{60}y{60}}
Classifier  & ScanNet200 & ADE20K \\
\shline
Learned & \textbf{27.8}$_{\textcolor{white}{\downarrow}}$ & \textbf{44.5}$_{\textcolor{white}{\downarrow}}$ \\
Fixed &   26.5$_{\textcolor{red}{\downarrow}}$ & 43.6$_{\textcolor{red}{\downarrow}}$ \\
\end{tabular}\vspace{-.5em}
\end{center}
It is shown that a semantic segmentation model with the classifier fixed as a simplex ETF performs much worse than a learnable classifier. Although using a fixed classifier of the simplex ETF structure has been proven to be effective for image recognition \cite{zhu2021geometric, yang2022we, graf2021dissecting}, we hold that in semantic segmentation the classifier needs to be learnable and does not have to be equiangular.

%In this case, the optimal classifier does not necessarily equally separate the label space. {\color{red} For example, the semantic classes ``sofa'' and ``chair'' in COCO show a smaller distance between their classifiers due to their similar semantic contexts.} \xjqi{if you want to use this example, move it before ``in this case'' would be better.} 

Second, the neural collapse phenomenon observed in image recognition highly relies on a balanced class distribution of training samples. 
It is indicated that neural collapse will be broken when data imbalance emerges, which explains the deteriorated performance of training on imbalanced data \cite{fang2021exploring}. 
We notice that semantic segmentation naturally suffers from data imbalance because some semantic classes are prone to cover a large area with significantly more points/pixels. 
%Semantic segmentation as a naturally imbalanced pixel-wise classification problem, 
Under the point/pixel-wise classification loss, the gradients will be also extremely imbalanced with respect to the backbone parameters, which breaks the equiangular separation structure for feature centers. In this case, the network makes the feature and classifier of minor classes lie in a close position and does not have the ability to discriminate the minor classes. However, the simplex ETF structure in neural collapse renders feature centers \textit{\textbf{equiangular separation}} and the \textit{\textbf{maximal discriminative}} ability, which is able to effectively improve the performance of minor classes in imbalanced recognition \cite{yang2022we, li2022targeted, zhu2022balanced}.

%% neural collapse does not happen in segmentation

%% reasons: imbalance (break NC 2), dependencies among pixels (break NC 3)

%% methods on classification do not work on seg

%% our method 

Inspired by our observations and analyses, we propose to induce the simplex ETF structure for feature centers, but keep a learnable classifier to enable adaptive class correlation for semantic segmentation. 
%resolve the imbalance problem in semantic segmentation to induce a maximal equiangular separation structure for feature centers, but keep the classifiers in a learnable state to hold the class dependencies. 
To this end, we propose an accompanied center regularization branch that extracts the feature centers of each semantic class. We regularize them by another classifier layer that is fixed as a simplex ETF. The fixed classifier forces feature centers to be aligned with the appealing structure, which enjoys the equiangular separation and the maximal discriminative ability. It in turn helps the feature learning in the original branch to improve the performance of minor classes for better semantic segmentation quality.  
%, which enjoys the maximally equiangular separation structure. {\color{blue} The fixed classifier induces the appealing structure for feature centers, which helps to improve the performance of minor classes.}  \xjqi{please add an explanation sentence how this regularization helps you regularize feature learning} 
We also provide theoretical results for a rigorous explanation. Our method can be easily integrated into any segmentation architecture and experimental results also show that our simple method consistently brings improvements on multiple image and point cloud semantic segmentation benchmarks. 
 
Our overall contributions can be listed as follows:
\begin{itemize}
\vspace{-1mm}
\item We are the first to explore neural collapse in semantic segmentation. We show that semantic segmentation naturally brings contextual correlation and imbalanced distribution among classes, which breaks the symmetric structure of neural collapse for both feature centers and classifiers. 
\item We propose a center collapse regularizer to encourage the network to learn class-equiangular and class-maximally separated structured features for imbalanced semantic segmentation.
\vspace{-2mm}
\item Our method is able to bring significant improvements on both point cloud and image semantic segmentation. Moreover, our method ranks {$\mathbf{1}^{\rm st}$} and sets a new record (+\textbf{6.8} mIoU) on the ScanNet200 test leaderboard. 
\end{itemize}

\begin{figure*}[t]
	\begin{center}
		\includegraphics[width=1.0\linewidth]{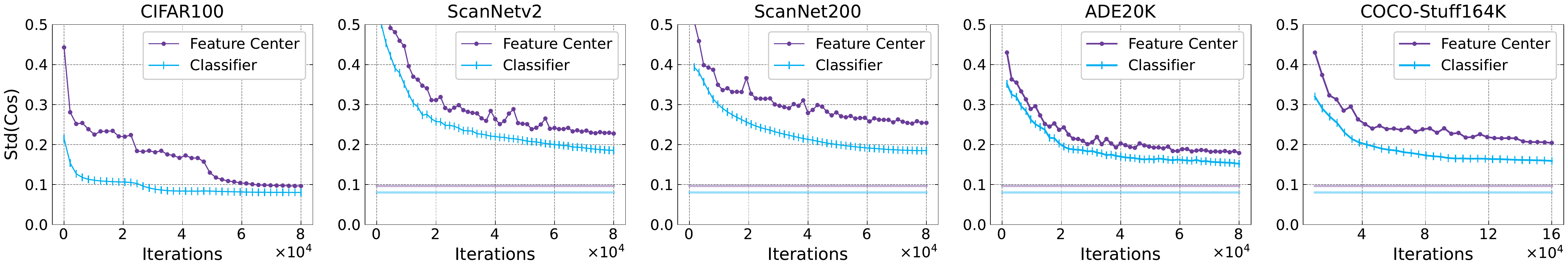}
            \vspace{-10pt}
	\begin{subfigure}[t]{0.196\linewidth}
		\centering
            \vspace{-14pt}
		\caption{}
		\label{fig3a}
	\end{subfigure}
	\hfill
	\begin{subfigure}[t]{0.196\linewidth}
		\centering
		\vspace{-14pt}
            \caption{}
		\label{fig3b}
	\end{subfigure}
	\hfill
	\begin{subfigure}[t]{0.196\linewidth}
		\centering
            \vspace{-14pt}		
            \caption{}
		\label{fig3c}
	\end{subfigure}
        \hfill
	\begin{subfigure}[t]{0.196\linewidth}
		\centering
		\vspace{-14pt}
            \caption{}
		\label{fig3d}
	\end{subfigure}
        \hfill
	\begin{subfigure}[t]{0.196\linewidth}
		\centering
		\vspace{-14pt}
            \caption{}
		\label{fig3e}
	\end{subfigure}
		\caption{Classifiers and train class-means approach equiangularity for  (a) \textit{recognition}, and worse equiangularity for (b-e) \textit{semantic segmentation}. The vertical axis shows the standard deviation of the cosines between pairs of centered class means (\textbf{{\color{my_purple}{purple}}} lines) and classifiers (\textbf{{\color{my_blue}{blue}}} lines) across all distinct pairs of classes $k$ and $k'$.  Mathematically, denote ${\rm Std}_{k \neq k'}(\cos(\hat{\z}_k,\hat{\z}_{k'}))$ and ${\rm Std}_{k \neq k'}(\cos(\hat{\w}_k,\hat{\w}_{k'}))$. As training progresses, the standard deviations of the cosines approach zero indicating equiangularity.}
        \vspace{-10pt}
		\label{fig3}
	\end{center}
\end{figure*}

%------------------------------------------------------------------------
\vspace{-3pt}
\section{Related Work}
\noindent\textbf{Neural collapse.}
Papyan et al. \cite{papyan2020prevalence} first discovered the neural collapse phenomenon that at the terminal phase of training, a classification model trained on a balanced dataset will have the last-layer features collapsed into their within class centers. These centers and classifiers will be formed as a simplex equiangular tight frame. Due to its elegant symmetry, later studies try to theoretically unravel such a phenomenon. It is proved that neural collapse is the global optimality of a simplified model with regularization \cite{zhu2021geometric, zhou2022optimization, tirer2022extended}, constraint \cite{fang2021exploring, graf2021dissecting, lu2020neural, weinan2020emergence}, or no explicit constraint \cite{ji2021unconstrained}, under the CE \cite{fang2021exploring, graf2021dissecting, ji2021unconstrained,lu2020neural, weinan2020emergence, zhu2021geometric} and the MSE loss functions \cite{han2021neural, mixon2020neural, poggio2020explicit, tirer2022extended, zhou2022optimization}. Some studies try to induce neural collapse in imbalanced learning for better accuracy of minor classes \cite{yang2022we, xie2022neural, thrampoulidis2022imbalance}. However, all these studies on neural collapse are limited in recognition. In contrast, we discover the solution structures of the last-layer feature centers and classifiers in semantic segmentation and propose to better induce the equiangular separation state accordingly.

\vspace{3pt}
\noindent\textbf{Semantic segmentation.}
Substantial progress was made for \textbf{2D semantic segmentation} with the introduction of FCN \cite{long2015fully}, which formulates the semantic segmentation task as per-point/pixel classification. Subsequently, many advanced methods have been introduced. Many approaches \cite{amirul2017gated, badrinarayanan2017segnet, lin2017refinenet, pohlen2017full, ronneberger2015u} combine up-sampled high-level feature maps and low-level feature maps to capture global information and recover sharp object boundaries. A large receptive field also plays an important role in semantic segmentation. To capture better global information, many studies \cite{chen2017deeplab, chen2014semantic,yu2015multi, chen2017rethinking, chen2018encoder} adopted spatial pyramid pooling to capture multi-scale and larger contextual information. On the other hand, with the introduction of large-scale annotated real-world 3D datasets~\cite{scannet, s3dis, semantickitti}, \textbf{3D semantic segmentation} has seen significant focus in recent years. Approaches for point cloud segmentation can be grouped into two categories, \ie, voxel-based and point-based methods. Voxel-based
solutions~\cite{3DSemanticSegmentationWithSubmanifoldSparseConvNet, SubmanifoldSparseConvNet, choy20194d} first divide the 3D space into regular voxels, and then apply sparse convolutions upon them.  Point-based methods~\cite{qi2017pointnet, qi2017pointnet++, zhao2020point, li2018pointcnn, lai2022stratified} directly adopt the point features and
positions as inputs, thus keeping the position information intact. However, all these studies on 2D \& 3D semantic segmentation mainly focus on network architecture and module design and ignore the impact of data distribution. As the datasets of semantic segmentation usually and naturally follow a heavily imbalanced distribution among classes, neural networks perform poorly when training on them~\cite{lws, logitsadj, balsfx, zhong2021improving, cui2021parametric}. With the available large-scale imbalanced segmentation datasets, such as ScanNet200~\cite{rozenberszki2022language}, ADE20K~\cite{zhou2017scene}, COCO~\cite{coco, Gupta2019LVIS}, exploring segmentation from an \textbf{imbalanced learning} view draws more attention recently.

%\noindent\textbf{Imbalanced Recognition.} 
%Deep neural networks usually perform poorly when training datasets are heavily class-imbalanced. In the area of imbalanced classification, recent methods can be roughly divided into five groups. Re-sampling~\cite{shen2016relay, zhong2016towards, buda2018systematic, byrd2019effect} and re-weighting~\cite{DBLP:conf/cvpr/HuangLLT16,huang2019deep,wang2017learning, ren2018learning, shu2019meta, jamal2020rethinking} approaches are based on the class frequencies and induce different penalties for different classes. Loss modification methods like LDAM~\cite{ldam}, Causal Norm~\cite{tang2020long}, Balanced Softmax~\cite{balsfx}, and LADE~\cite{lade} distribute class-wise decision boundaries. Architecture modification methods like BBN~\cite{bbn}, ELF~\cite{elf}, ResLT~\cite{cui2021reslt}, GistNet~\cite{liu2021gistnet}, RIDE~\cite{ride}, and DiVE~\cite{he2021distilling} design specific network modules for different categories. Decouple models~(cRT, LWS)~\cite{lws}, DisAlign~\cite{zhang2021distribution}, MiSLAS~\cite{mislas}, SSD~\cite{li2021self} and other multi-stage methods fix the backbone and fine-tune the classifier on re-sampling datasets.  Motivated by contrastive methods, Hybrid-PSC~\cite{wang2021contrastive}, RSG~\cite{wang2021rsg}, DRO-LT~\cite{samuel2021distributional}, and PaCo~\cite{paco} find an assisted contrastive learning process can enhance imbalanced recognition.

\begin{figure*}[!t]
	\begin{center}
		\includegraphics[width=0.99\textwidth]{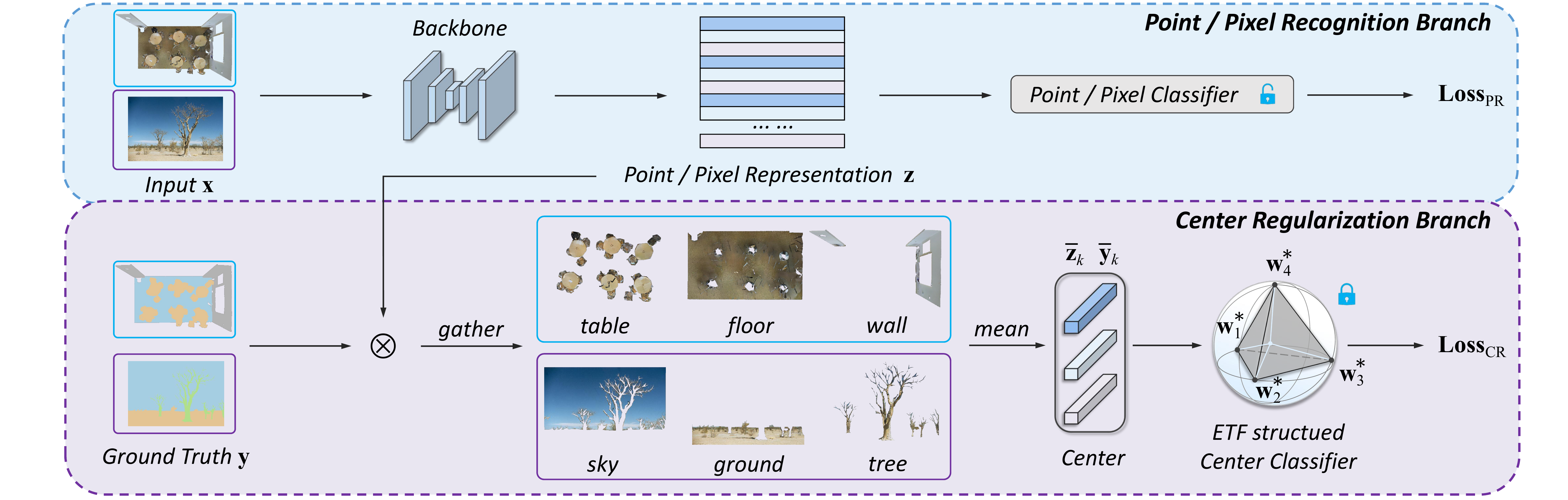}
		\caption{Framework of our method. The \textbf{{\color{my_blue}{blue}}} square part and \textbf{{\color{my_purple}{purple}}} square part represent 3D scene input illustration and 2D image input illustration, respectively. The point/pixel recognition branch is similar to the conventional segmentation model. We introduce a center regularization branch to make the feature center collapse to an ETF structure during training and remove it during evaluation.}
		\label{framework}
	\end{center}
	\vspace{-0.00in}
\end{figure*}

\section{Neural Collapse Observations}

\subsection{Neural Collapse in 
 Recognition}

% point / pixel 如何有效的统一起来？ element wise?

We consider a dataset, having $N$ training samples in total and the annotations $\y \in \mathbb{R}^N$. More concretely, it has $K$ classes and $n_k$ examples in $k$-th class. We refer to $\y_i \in \{1, ..., K\}$ as the label, $\mathbf{z}_i\in \mathbb{R}^d$ as the last-layer $d$-dimensional feature of the $i$-th sample. We define that ${\bar{\z}}_k=\mathrm{Avg}_{\y_i=k}\{\z_{i}\}$ is the within-class mean of the last-layer features in the $k$-th class. The linear classifier is specified by weights $\mathbf{W}=[\mathbf{w}_1, ...,\mathbf{w}_K]\in\mathbb{R}^{d\times K}$.

Papyan et al. \cite{papyan2020prevalence} revealed the neural collapse phenomenon for image recognition. It states that the last-layer features will converge to their within-class means, and the within-class means together with the classifiers will collapse to the vertices of a simplex equiangular tight frame at the terminal phase of training (after 0 training error rate).

\begin{definition}[Simplex Equiangular Tight Frame]
	\label{ETF}
	A collection of vectors $\mathbf{m}_k\in\mathbb{R}^d$, $k=1,2, ..., K$, $d\ge K$, is said to be a simplex equiangular tight frame if:
	\begin{equation}\label{ETF_M}
		\mathbf{M}=\sqrt{\frac{K}{K-1}}\mathbf{U}\left(\mathbf{I}_K-\frac{1}{K}\mathbf{1}_K\mathbf{1}_K^\top\right),
	\end{equation}
	where $\mathbf{M}=[\mathbf{m}_1, ...,\mathbf{m}_K]\in\mathbb{R}^{d\times K}$, $\mathbf{U}\in\mathbb{R}^{d\times K}$ allows a rotation and satisfies $\mathbf{U}^\top\mathbf{U}=\mathbf{I}_K$, $\mathbf{I}_K$ is the identity matrix, and $\mathbf{1}_K$ is an all-ones vector. 
\end{definition}

All vectors in a simplex ETF have an equal $\ell_2$ norm and the same pair-wise angle, \emph{i.e.,}
\begin{equation}\label{mimj}
	\mathbf{m}_i^\top\mathbf{m}_j=\frac{K}{K-1}\delta_{i,j}-\frac{1}{K-1}, \forall i, j \in \{1, ..., K\},
\end{equation}
where $\delta_{i,j}$ equals to $1$ when $i=j$ and $0$ otherwise. The pair-wise angle $-\frac{1}{K-1}$ is the maximal equiangular separation of $K$ vectors in $\mathbb{R}^d, d\ge K-1$ \cite{fang2021exploring, yang2022we}.

Then the two important geometric properties instructed by neural collapse can be formally described as:
%the neural collapse phenomenon can be formally described as:
(1) The normalized within-class centers converge to a simplex ETF, \emph{i.e.,} $\hat{\z}_k = ({\bar{\z}}_k-{\z}_G)/||{\bar{\z}}_k-{\z}_G||$ satisfies Eq.~\eqref{mimj}, where ${\z}_G=\mathrm{Avg}_{i}\{\z_{i}\}$ is the global mean of the last-layer features for all samples; (2) The normalized classifier vectors converge to the same simplex ETF as feature centers, \emph{i.e.,} $\hat{\w}_k=\w_k/||\w_k||=\hat{\z}_k$ and satisfies Eq.~\eqref{mimj}, where $\w_k$ is the classifier of the $k$-th class. 

\subsection{Neural Collapse in Semantic Segmentation}\label{nc_seg}

The elegant phenomenon has only been discovered and studied in image recognition, which performs image-wise classification. In this paper, we explore the corresponding structures of the last-layer feature centers and classifiers in image and point cloud semantic segmentation, which performs pixel- and point-wise classification, respectively. 
%view image and point cloud semantic segmentation as a pixel/point-level classification problem. 

Following \cite{papyan2020prevalence}, we calculate statistics during training to show the neural collapse convergence in semantic segmentation. We first compare the standard deviations of two cosine similarities, $\cos(\hat{\z}_{k}, \hat{\z}_{k'})$ and $\cos(\hat{\w}_k, \hat{\w}_{k'})$, for all pairs of different classes $k \ne k'$. The standard deviations of the cosines approach zero indicating equiangularity of the feature centers and classifier weights. As shown in Fig. \ref{fig3}, their standard deviations are much larger on semantic segmentation (Fig.~\ref{fig3b} to \ref{fig3e}, light-colored lines for CIFAR100 results), compared with them on image recognition (Fig.~\ref{fig3a}) as observed by Papyan et al. \cite{papyan2020prevalence}. It indicates that the equiangular structure of the feature centers and classifiers in semantic segmentation is not as valid as that in image recognition. Similarly, the feature centers and classifiers  approach the maximal-angle structure more closely in classification than in semantic segmentation. More analysis and experiments about the maximal separation structure of the feature centers and classifiers are shown in \textit{Appendix}~\ref{sec_nc_seg}.

We point out that semantic segmentation naturally brings contextual correlation and imbalanced distribution among classes, which may break the symmetric structure of neural collapse for both feature centers and classifiers. A learnable classifier will lead to a non-equiangular structure to be \textit{adaptable} to class contextual correlation. But the class imbalance will cause imbalanced gradients. As a result, the feature centers for minor classes will be closed after training, which impedes their performance. A rigorous analysis is conducted in Sec.~\ref{Math}.

% \begin{figure*}[!t]
% 	\begin{center}
% 		\includegraphics[width=1.0\textwidth]{fig/p_vs_center.pdf}
% 		\caption{\textbf{Comparison between point/pixel imbalance and class center imbalance.}
% Histogram statistics of point/center frequency over sorted class indexes on ScanNet-v2, ScanNet200 (left), and pixel/center frequency on ADE20K, COCO-Stuff164K (right). The imbalance factor (IF) is defined as $\frac{n_{\rm max}}{n_{\rm min}}$, where $n_{\rm max}$ and $n_{\rm min}$ are the numbers of samples (points, pixels, centers) for the most frequent class and the least frequent class of dataset. It can be seen that the center imbalance is greatly alleviated compared with the point/pixel imbalance.}
% 		\label{fig5}
% 	\end{center}
% 	\vspace{-0.00in}
% \end{figure*}

\section{Main Approach}\label{main_method}
\subsection{Motivation}

As said in Sec.~\ref{sec:intro}, although the classifier does not have to be equiangular, we note that the ETF structure in Eq.~\eqref{ETF_M} is appealing for feature centers in a classification problem, especially when the learning is imbalanced. First, such a structure has a balanced separation among all classes and the separation is maximally enlarged. We formally state this property in Lemma \ref{lemma}. Second, as suggested by prior studies, inducing neural collapse in imbalanced learning is able to improve the performance of minor classes \cite{yang2022we, xie2022neural, thrampoulidis2022imbalance}. 
\begin{lemma}
\label{lemma}
(\textbf{Equiangular \& Maximal Separated Property}) 

For any normalized matrix,  $\mathbf{W}=[\mathbf{w}_1, ...,\mathbf{w}_K]\in\mathbb{R}^{d\times K}, d \geq K, \mathbf{w}_k^\top\mathbf{w}_k=1, \forall k=1, ..., K$, the maximal separation value is $-\frac{1}{K-1}$, i.e., $\max_{k \not= k'} \cos(\w_k, \w_{k'}) \geq -\frac{1}{K-1}$. Equality holds if and if only the matrix is a simplex ETF, i.e., $\mathbf{W}$ satisfies Eq.~\eqref{ETF_M} and it enjoys the equiangular property, i.e., $\forall k \neq k', \cos(\w_k, \w_{k'})$ is a constant.
\end{lemma}
\begin{proof} 
Please refer to our Appendix~\ref{sec_proof} for proof. %\ref{app_1}
$\hfill\square$  
\end{proof}
Based on our observations and analyses, we propose to regularize the feature centers in a segmentation model into the simplex ETF structure in Eq.~\eqref{ETF_M} to relieve the imbalance dilemma for semantic segmentation.

%The above two reasons may lead to better generalization and performances of the minor classes in imbalanced semantic segmentation. To relieve the negative effect of data imbalance and the dependency in semantic segmentation, we propose a center classification branch to regularize the feature mean collapse to the ETF structure.

%It indicates that the two models are both trained towards the neural collapse state, while the one with our method converges more closely. 

%We further calculate the averages of $\cos\angle(\h_c-\h_g, \w_c)$, $\forall 1\le c\le K$, and $||\tilde{\W}-\tilde{\H}||_F^2$ in Figure \ref{diag} and \ref{F2} in Appendix \ref{exp}. It reveals that the model using our method generally has a higher $\cos\angle(\h_c-\h_g, \w_c)$ and a lower $||\tilde{\W}-\tilde{\H}||_F^2$, which indicates that the feature means and classifier vectors of the same class are better aligned. We observe no advantage of ResNet on STL-10 and DenseNet on CIFAR-100 in Figure \ref{diag} and \ref{F2}. In Table \ref{longtail}, we see that the two cases are right the failure cases, which shows consistency between neural collapse convergence and classification performance.

\subsection{Center Collapse Regularizer}\label{sec_c2r}
%\subsubsection{Operating Mechanism}
To take advantage of the equiangular and maximum separation properties for better performance on minor classes, we propose a \textbf{Ce}nter \textbf{Co}llapse Regularizer~(\textbf{\ours{}}) for imbalanced semantic segmentation problems. The overview of our method is shown in Fig.~\ref{framework}. The whole framework can be divided into two branches: the point/pixel recognition branch (upper part) and the center regularization branch (bottom part). The point/pixel recognition branch refers to a point cloud or image segmentation model, dealing with the semantic segmentation task in a point/pixel level manner. Therefore, our \ours{} can be easily integrated into any off-the-shelf segmentation architecture. 

For simplicity, we use a 3D scene (\textbf{{\color{my_blue}{blue}}} square part) and an image (\textbf{{\color{my_purple}{purple}}} square part) in Fig.~\ref{framework} as an example to illustrate the point cloud and image semantic segmentation, respectively. The input is represented as $\mathbf{x} \in \mathbb{R}^{N\times s}$, where $N=HW$ is the number of pixels and $s=3$ denotes the color dimension for the image case.
For point cloud, $N$ is the number of points, and $s=6$ is for both the color and the position dimension. As shown in the point/pixel recognition branch, we can get the feature representation $\mathbf{Z}=[\z_1, ...,\z_N]^\top \in \mathbb{R}^{N\times d}$ after the backbone feature extraction. The loss $\mathcal{L}_{\rm PR}$ for the point/pixel recognition branch is a point/pixel-wise CE for supervised learning.

%At the end of the point/pixel recognition branch, a point/pixel classifier and a point/pixel-wise cross-entropy loss $\mathcal{L}_{\rm PCE}$ are adopted for supervised learning. 

In the center regularization branch, we first gather $\z_i$ of the same class, compute the feature centers $\bar{\z}_k$, and generate center labels $\bar{\y}_k$ of all classes based on the ground truth $\mathbf{y}$:
\begin{equation}\label{center_compute}
	\bar{\z}_k = \frac{1}{n_k}\sum_{\mathbf{y}_i = k}^{n_k}\mathbf{z}_i, \ \ \bar{\y}_k=\y_i = k,
\end{equation}
where $n_k$ is the number of samples in $\mathbf{Z}$ belonging to the $k$-th class. We concatenate feature centers $\bar{\mathbf{Z}}=[\bar{\z}_1, ... , \bar{\z}_K] \in \mathbb{R}^{d \times K}$ and center labels $\bar{\y}=[1, ... , K] \in \mathbb{R}^{K}$ for the center regularization branch. To achieve center collapse, we introduce an ETF structured classifier for this branch. Another CE-based center collapse regularization loss $\mathcal{L}_{\rm CR}$ upon feature centers $\bar{\mathbf{Z}}$ and their center labels $\bar{\mathbf{y}}$ is used to measure the degree of feature center collapse. Concretely, we initialize the classifier $\W^*$ as a random simplex ETF by Eq.~\eqref{ETF_M}. During training, the classifier is fixed to make the maximal equiangular separation property satisfied all the time, \ie,
\begin{equation}
	{\w_k^*}^\top\w^*_{k'}=\alpha^2\left(\frac{K\delta_{k,k'}}{K-1}-\frac{1}{K-1}\right), \ \forall k,\  k'\in\{1, ..., K\}, \notag
\end{equation}
{where $\alpha$ is a hyper-parameter of weight scaling, and $\delta_{k,k'}$ equals to $1$ when $k=k'$ and $0$ otherwise. Then, a CE type of $\mathcal{L}_{\rm CR}$ loss can be written as follows:
\begin{equation}
   % \mathcal{L}_{\rm total} = \mathcal{L}_{\rm PR}({\mathbf{Z}},\ {\y}) + \lambda \mathcal{L}_{\rm CR}(\bar{\mathbf{Z}},\ \bar{\y}) =\mathcal{L}_{\rm PR}({\mathbf{Z}},\ {\y}) + \lambda \mathcal{L}_{\rm CR}(\bar{\mathbf{Z}},\ \bar{\y}) \label{eq:all},
   \mathcal{L}_{\rm CR}(\bar{\mathbf{Z}}, \W^*) = -\sum_{k=1}^K\log\left(\frac{\exp(\bar{\z}_k^\top\mathbf{w}^*_k)}{\sum_{k'=1}^{K}\exp(\bar{\z}_{k'}^\top\mathbf{w}^*_{k'})}\right).
\end{equation}
We define the total loss as the combination of two branches:
\begin{equation}\label{eq:all}
\mathcal{L}_{\rm total}\!=\!\mathcal{L}_{\rm PR}({\mathbf{Z}}, {\y}) +\lambda\mathcal{L}_{\rm CR}(\bar{\mathbf{Z}}, \W^*),
\end{equation}
where $\lambda$ is a loss weight hyper-parameter. 

Additionally, in the evaluation of our method, we only preserve the point/pixel recognition branch. It means that just the point/pixel classifier is preserved, while the center regularization branch is discarded. Therefore, the evaluation of \ours{} is very efficient: It is consistent with a conventional backbone like the vanilla ResNet~\cite{he2016deep}, without any additional computations.

%\subsubsection{Accuracy Correlation and Center Imbalance}
\subsection{Empirical Support}~\label{sec_emp}
In this subsection, we give some empirical evidence to show that our \ours{} can do better rebalance on semantic segmentation. Due to limited pages, the detailed results and evidence are shown in \textit{Appendix}~\ref{sec_emp_more}. We list the main conclusions in the following:

First, the center regularization branch in \ours{} transforms each point/pixel pair $(\z_i, \y_i)$ of the $k$-th class to a center pair $(\bar{\z}_k, \bar{\y}_k)$, which greatly decreases the imbalanced factor~\cite{cui2019class, liu2019large}: (The imbalanced factor is defined as $\frac{n_{\rm max}}{n_{\rm min}}$, where $n_{\rm max}$ and $n_{\rm min}$ are the maximal and minimal numbers of samples in all classes, respectively.) 
\begin{center}\vspace{-.2em}
\tablestyle{4pt}{1.05}
\begin{tabular}{y{20}y{22}y{20}y{22}y{20}y{22}y{20}y{22}}
\multicolumn{2}{l}{ScanNetv2}  & \multicolumn{2}{l}{ScanNet200} & \multicolumn{2}{l}{ADE20K} & \multicolumn{2}{l}{COCO-Stuff}\\
\shline
(point) & 116 & (point) & 37256& (pixel) &827 & (pixel) & 2612\\
(center)&  \textbf{11}$_{\textcolor{red}{\downarrow}}$ & (center) &\textbf{597}$_{\textcolor{red}{\downarrow}}$ & (center) &\textbf{282}$_{\textcolor{red}{\downarrow}}$ & (center) &\textbf{528}$_{\textcolor{red}{\downarrow}}$\\
\end{tabular}\vspace{-.2em}
\end{center}
As the imbalance severity of the dataset is significantly reduced, it becomes more friendly to the minor classes.

Second, in long-tailed classification, the class accuracy is \textit{positively correlated} with the class image number of the training dataset. However, in both point cloud and image semantic segmentation, the class accuracy has \textit{weak correlations} with class point/pixel numbers due to correlations among neighboring. 
The widely used correlation measurement, Pearson correlation coefficients~\cite{benesty2009pearson} between the class accuracy and the sample numbers in each class:
\begin{center}\vspace{-.2em}
\tablestyle{4pt}{1.05}
\begin{tabular}{y{20}y{40}y{20}y{35}y{20}y{35}}
\multicolumn{2}{l}{CIFAR100-LT-100}  & \multicolumn{2}{l}{ScanNet200} & \multicolumn{2}{l}{ADE20K} \\
\shline
(image) & 0.76 & (point) & 0.32 & (pixel) & 0.31 \\
  \hspace{2pt} -- &  & (center) &\textbf{0.51}$_{\textcolor[RGB]{0,150,0}{\uparrow}}$ & (center) &\textbf{0.49}$_{\textcolor[RGB]{0,150,0}{\uparrow}}$
\end{tabular}\vspace{-.2em}
\end{center}
However, most of the imbalanced recognition methods are using the sample numbers of classes for guidance, \eg, reweighting~\cite{cui2019class, huang2019deep, shu2019meta, jamal2020rethinking} and loss adjustment~\cite{Cao_nips, balsfx, logitsadj, zhong2021improving}. Due to the low correlations for the point and pixel cases, it is probably not feasible for imbalanced semantic segmentation. In contrast, \ours{} regularizes imbalance in a feature center space. Feature centers are more global representation and greatly eliminate the effects of correlations among neighboring. Thus, it has a \textit{better correlation} with class accuracy than the point/pixel frequency, further indicating its suitability for semantic segmentation rebalancing.

\subsection{Theoretical Support}~\label{Math}
In this part, we rethink the CE-based center collapse loss $\mathcal{L}_{\rm CR}$ from the perspective of gradients to analyze its imbalanced learning behaviors. 

\vspace{3pt}
\noindent\textbf{Gradient \emph{w.r.t} the center classifier.}
Recalling the center computation Eq.~\eqref{center_compute}, we first analyze the gradient of $\mathcal{L}_{\rm CR}$ \emph{w.r.t} the center classifier $\mathbf{W}=[\mathbf{w}_1,\cdots,\mathbf{w}_K]\in\mathbb{R}^{d\times K}$:
\begin{align}
    \label{eq_spm}
        \frac{\partial \mathcal{L}_{\rm CR}}{\partial\w_k} &= \left(p_k\left({\bar{\z}}_k\right)-1\right)\bm{{\bar{\z}}}_k+\sum_{k'\ne k}^{K-1}p_k\left(\bm{{\bar{\z}}}_{k'}\right)\bm{{\bar{\z}}}_{k'}, 
        \\
        &=\underbrace{\sum_{\mathbf{y}_i = k}^{n_k}\left(p_k\left(\bm{{\bar{\z}}}_k\right)\!-\!1\right)\frac{\mathbf{z}_i}{n_k}}_{\text{within-class}}+\underbrace{\sum_{k'\ne k}^{K-1}\sum_{\mathbf{y}_j =k'}^{n_k}p_k\left(\bm{{\bar{\z}}}_{k'}\right)\frac{\mathbf{z}_j}{n_{k'}}}_{\text{between-class}}, \notag
\end{align}
where $p_k(\bm{{\bar{\z}}})$ is the predicted probability that $\bm{{\bar{\z}}}$ belongs to the $k$-th class. It is calculated by the softmax transformation and takes the following form in the CE loss:
\begin{equation}\label{p_i}
	p_k(\bm{{\bar{\z}}}) = \frac{\exp(\bm{{\bar{\z}}}^\top\mathbf{w}_k)}{\sum_{k'=1}^{K}\exp(\bm{{\bar{\z}}}^\top\mathbf{w}_{k'})},\ k=1, 2, ..., K.
\end{equation}
It reveals that the gradient \emph{w.r.t}  $\w_k$ is also \textit{imbalanced} and can be decomposed into two parts. The ``within-class'' part contains $n_k$ terms and pulls $\w_k$ towards the directions of the same class feature center, \emph{i.e.}, $\bar{\z}_k$. While the ``between-class'' part contains $\sum_{k' \neq k}n_{k'}$  terms and pushes $\w_k$ away from the directions of the features of the other classes. Thus, the gradients of some minor classes can be more likely swallowed by the other classes, which is unfriendly to the decision boundaries of minor classes. If we fix the center classifier weights, it can avoid the imbalance gradient updating and benefit the minor classes discrimination.

\begin{figure}[t]
    \centering
    \includegraphics[width=0.99\linewidth]{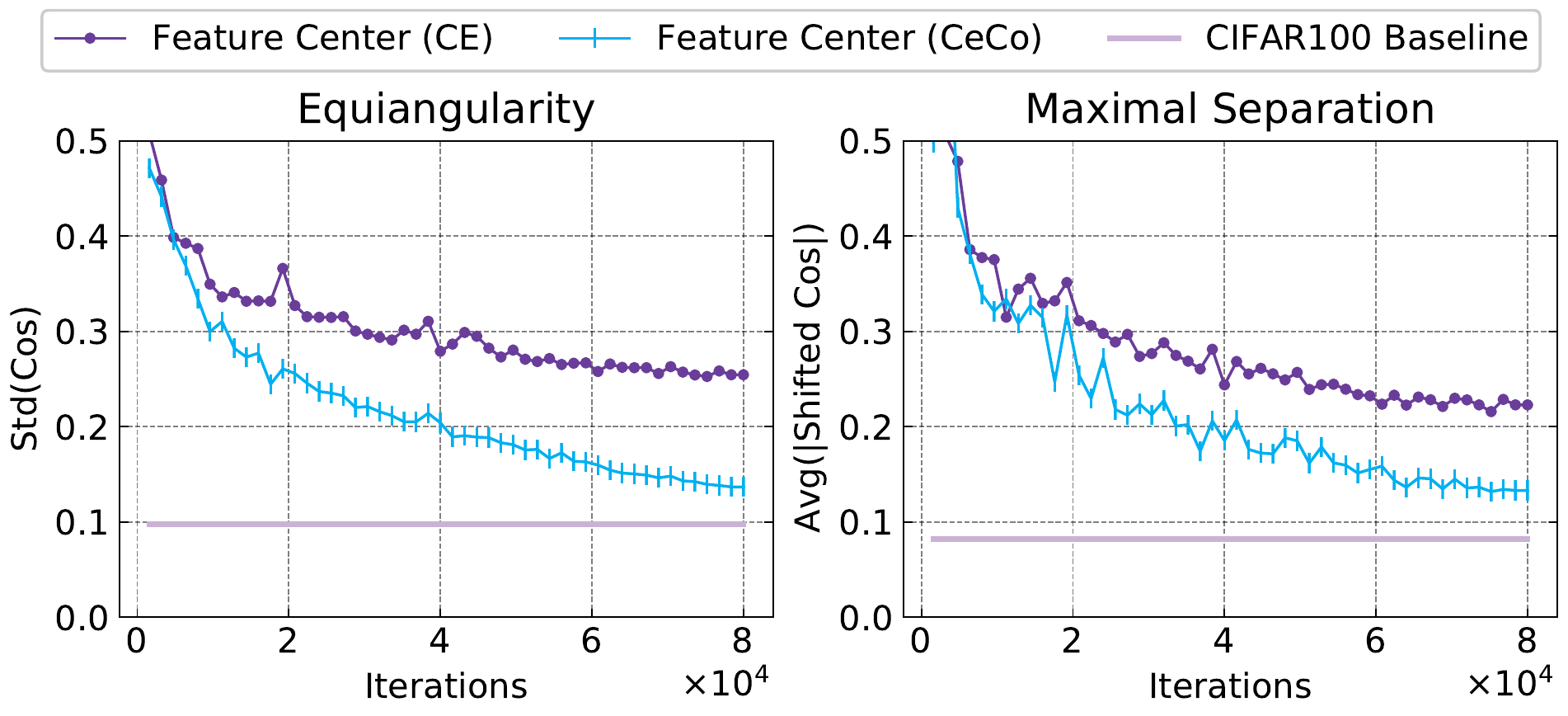}
    \begin{subfigure}[t]{0.49\linewidth}
		\centering
            \vspace{-14pt}
		\caption{}
		\label{fig4a}
	\end{subfigure}
	\hfill
	\begin{subfigure}[t]{0.49\linewidth}
		\centering
		\vspace{-14pt}
            \caption{}
		\label{fig4b}
	\end{subfigure}
	\hfill
    \vspace{-10pt}
    \caption{Ablation on the center equiangularity (a) and maximal separation (b) on ScanNet200. Adopting \ours{}, the feature centers can obtain a better equiangular and maximally separated structure. It is better to look together with Fig.~\ref{fig3c}.}
    \vspace{-10pt}
    \label{ce_vs_cr}
\end{figure}

\noindent\textbf{Gradient \emph{w.r.t} point/pixel features.}
Since the center collapse loss $\mathcal{L}_{\rm CR}$ is a regularization upon the point/pixel feature $\z_i$, we analyze how the center collapse loss influences the point/pixel feature also from the gradient view. Suppose the class label $\y_i=k$, $\z_i$ contributes to $\bar{\z}_k$. The gradient of $\mathcal{L}_{\rm CR}$ with respect to $\z_i$ is:
\begin{equation}\label{grad_z}
	\frac{\partial \mathcal{L}_{\rm CR}}{\partial \z_i}\!=\!\frac{\partial \mathcal{L}_{\rm CR}}{\partial{\bar{\z}}_k}
	\bigcdot\frac{\partial\bm{{\bar{\z}}}_{k}} {\partial \z_i}\!=\!\sum_{k'=1}^{K}p_{k'}(\bar{\z}_{k}){{(\w_{k'} - \w_{k})}}\bigcdot\frac{1}{n_k}.
\end{equation}
For Eq.\eqref{grad_z}, we mainly consider the medium term, \ie, $\w_{k'} - \w_{k}$. In \cite{fang2021exploring}, the authors describe a minority collapse phenomenon that the classifier weight vectors of minor classes converge in a similar direction under the extreme imbalance case, \ie, $\lim_{\frac{n_{\rm max}}{n_{k}} \to \infty, \frac{n_{\rm max}}{n_{k'}} \to \infty} \w_{k'} - \w_k = \boldsymbol{0}_{d}$.
% \begin{equation}
% 	\lim_{\frac{n_{\rm max}}{n_{k}} \to \infty, \frac{n_{\rm max}}{n_{k'}} \to \infty} \w_{k'} - \w_k = \boldsymbol{0}_{d}. \notag
% \end{equation}
If the classifier is fixed ETF structured, $\|\w_{k'} - \w_{k}\|=\frac{2K}{K-1}, \forall k \neq k',$ is always satisfied during training, according to the ETF property Eq.~\eqref{mimj}. It means that ETF structured classifier enables the gradient transfer between minor classes and obtains \textit{better discrimination among minor classes}. Thus it avoids the features of minor classes converging in a close position (see Fig.~\ref{fig1_2}).

\section{Experiments}

\subsection{Datasets and Implementation Details}
To evaluate the effectiveness of our method, we conduct experiments on the most popular benchmarks in semantic segmentation: \ie, ScanNet200~\cite{rozenberszki2022language} for point cloud semantic segmentation, ADE20K~\cite{zhou2017scene} and COCO-Stuff164K~\cite{caesar2018coco} for image semantic segmentation. All these three datasets contain more than \textit{150 classes}, which is more suitable to validate the imbalanced performance of the proposed method under severe imbalance cases.
Implementation details and dataset descriptions are available in \textit{Appendix}~\ref{sec_data}.

\subsection{Ablation Study}
In this subsection, we conduct ablation experiments to examine the effectiveness of \ours{}. The ablation results are reported on the ScanNet200 validation dataset with the MinkowskiNet~\cite{choy20194d} backbone for 3D semantic segmentation, and on ADE20K (single-scale inference mIoU) with the Swin-T~\cite{liu2021swin} backbone for 2D semantic segmentation.

\vspace{3pt}
\noindent\textbf{Equiangularity and maximal separation analysis.} 
 We first analyze the equiangularity property of the feature centers. Following \cite{papyan2020prevalence} and Sec.~\ref{nc_seg}, we calculate the standard deviation of the cosines between pairs of centered class means across all distinct pairs of classes $k$ and $k'$. Mathematically, we measure ${\rm Std}_{k \neq k'}(\cos(\hat{\z}_k,\hat{\z}_k))$ and ${\rm Avg}_{k \neq k'}|\cos(\hat{\z}_k,\hat{\z}_k) + 1/(K-1)|$. Fig.~\ref{fig4a} and \ref{fig4b} show the change of standard deviation and average at different iterations. As training progresses, the standard deviations of the cosines approach zero indicating equiangularity, and the shifted averages of the cosines approach zero indicating maximal angle. \ours{} (\textbf{{\color{my_blue}{blue}}} lines) introduces a center collapse regularization and achieves a much \textit{smaller} standard deviation and shifted average values than the vanilla CE model (\textbf{{\color{my_purple}{purple}}} lines), and are \textit{closer} to the classification neural collapse results (\textbf{\color{my_light}{light-colored}} lines for reference). 

\begin{table}[t]
	\centering
        \small
	\renewcommand\arraystretch{1.1}
	{
		\begin{tabular}{y{46}y{46}y{46}y{46}}
			\toprule
			PC  & CC  & ScanNet200 & ADE20K \\
			\midrule
			Learned & --  &27.8 &44.5\\
                Fixed & Fixed  &28.3 &44.8\\
                Fixed & Learned &27.2 &44.0\\
                \textbf{Learned}  & \textbf{Fixed}  &\textbf{30.3}\cgaphl{+}{2.5} &\textbf{45.7}\cgaphl{+}{1.2}\\
                Learned  & Learned &26.9 &43.7\\
			\bottomrule
		\end{tabular}}
        \caption{Ablation on the fixed ETF structured or learned classifier. PC: point/pixel classifier. CC: feature center classifier.}
        \label{fixed_table}
\end{table}

\begin{table}[t]
	\centering
        \small
	\renewcommand\arraystretch{1.1}
        
	{
		\begin{tabular}{y{72}y{59}y{65}}
			\toprule
			{\textbf{Loss Weight}} & ScanNet200 & ADE20K \\
			\midrule
			$\lambda$ = 0.0 (Baseline) &27.8 &44.5\\
                $\lambda$ = 0.1  &28.6 &44.8\\
                $\lambda$ = 0.2  &29.4 &45.0\\
                $\lambda$ = 0.3  &30.1 &45.2\\
                $\lambda$ = 0.4  &\textbf{30.3}\cgaphl{+}{2.5} &45.3\\
                $\lambda$ = 0.5  &29.7 &\textbf{45.7}\cgaphl{+}{1.2}\\
                $\lambda$ = 0.6  &28.7 &44.9\\
			\bottomrule
		\end{tabular}}
        \caption{Ablation on the loss weight hyper-parameter $\lambda$.}
        \label{lambda_table}
        \vspace{-5pt}
\end{table}

\vspace{3pt}
\noindent \textbf{Ablation on the fixed ETF structured center classifier.} As discussed in Sec.~\ref{main_method}, a fixed ETF structured center classifier can bring many benefits under severely imbalanced cases. In this part, we conduct experiments to answer the following question: For both the point/pixel classifier and center classifier, should we make them fixed or learnable? We list the performance results for four kinds of variants in Table~\ref{fixed_table}. The fixed ETF structured center classifier induces a strong regularization for feature center distribution (equiangular and maximal separated) and improves the mIoU performance by 2.6\%, 1.2\% on ScanNet200 and ADE20K compared with the based model, respectively. Consistent with our analysis in Sec.~\ref{sec:intro} and similar to the conventional semantic segmentation models, a learnable point/pixel classifier can achieve better results. it is \textit{adaptive} and dynamically updated and hence can better handle a finer point/pixel-level classification.

\vspace{3pt}
\noindent\textbf{Ablation on the loss weight hyper-parameter $\lambda$.} In Eq.~\eqref{eq:all}, we introduce the hyper-parameter $\lambda$ for the weighting between the loss function of the point/pixel recognition branch $\mathcal{L}_{\rm PR}$ and the loss function of the center regularization branch $\mathcal{L}_{\rm CR}$. 
To show the sensitivity of our \ours{} to different $\lambda$ values, we conduct an ablation on ScanNet200 and ADE20K. Table \ref{lambda_table} lists the experimental results, showing that the performance can be consistently improved with the value of $\lambda$ in a wide range. According to Table \ref{lambda_table}, $\lambda=0.4$ can achieve the best mIoU results among others on ScanNet200, and $\lambda=0.5$ can achieve the best mIoU results among others on ADE20K.

\begin{table}[t]
	\centering
        \small
	\renewcommand\arraystretch{1.1}
        
	{
		\begin{tabular}{y{80}y{60}y{60}}
			\toprule
 			\textbf{Loss Type} & ScanNet200 & ADE20K \\
			\midrule
			CE (Baseline) & 27.8 &44.5\\
                + Dice~\cite{milletari2016v}  &28.8 &44.8\\
                \cellcolor{Gray}+ Dice + \textbf{\ours{}}  &\cellcolor{Gray}\textbf{30.6}\cgaphl{+}{1.8} &\cellcolor{Gray}\textbf{45.8}\cgaphl{+}{1.0}\\
                + Lovász~\cite{berman2018lovasz}  &30.0 &45.0\\
                \cellcolor{Gray}+ Lovász + \textbf{\ours{}}  &\cellcolor{Gray}\textbf{32.0}\cgaphl{+}{2.0} &\cellcolor{Gray}\textbf{46.5}\cgaphl{+}{1.5}\\
			\bottomrule
		\end{tabular}}
        \caption{Ablations on orthogonality to the Dice and Lovász loss.}
        \vspace{-3pt}
        \label{loss_table}
\end{table}

\begin{table}[t!]
\centering
\renewcommand\arraystretch{1.1}
\resizebox{1.0\linewidth}{!}{
\begin{tabular}{y{70}y{40}y{40}y{40}y{40}}
%{y{80}|y{50}y{50}y{50}y{50}|y{50}y{50}y{50}y{50}}
\toprule
{\textbf{Method}} & Head & Comm. & Tail &  \textbf{All} \\
\midrule
DLV3P (R50)   &  67.7  &   48.3  &   36.4 &  44.9 \\
+ DisAlign    & 67.7 & 48.6 & 37.8 & 45.7 \\
\cellcolor{Gray} + \textbf{\ours{}}   & \cellcolor{Gray} 67.7 & \cellcolor{Gray} \textbf{48.7}\cgaphl{+}{0.1} & \cellcolor{Gray} \textbf{39.0}\cgaphl{+}{1.2} & \cellcolor{Gray} \textbf{46.4}\cgaphl{+}{0.7}\\
\midrule
DLV3P (R101) & 68.7  & 49.0 & 38.4 & 46.4\\
+ DisAlign & 68.7 & 49.4 & 39.6 &  47.1\\
\cellcolor{Gray} + \textbf{\ours{}}  &\cellcolor{Gray} \textbf{68.8}\cgaphl{+}{0.1} & \cellcolor{Gray} 49.4 &\cellcolor{Gray} \textbf{40.9}\cgaphl{+}{1.3} & \cellcolor{Gray}\textbf{48.0}\cgaphl{+}{0.9}\\
\bottomrule
\end{tabular}}
\caption{Imbalanced performance comparison of our method and the SOTA long-tailed segmentation framework~DisAlign~\cite{zhang2021distribution} on ADE20K. All compared methods are based on DLV3P~\cite{chen2018encoder} with the ResNet-50~(R50) and ResNet-101~(R101) backbone.}
\label{ade_imbalance}
\end{table}

\vspace{3pt}
\noindent\textbf{Imbalanced performance.}
To evaluate the imbalanced performance of our method, 
We also follow~\cite{rozenberszki2022language, zhang2021distribution} to split the categories of ScanNet200 and ADE20K
into three subsets and report the average IoU and accuracy in these three subsets: head-shot, medium-shot, and few-shot, which are also called the \textit{Head}, \textit{Common} and \textit{Tail} categories, respectively. We list the detailed results of ADE20K in Table~\ref{ade_imbalance} and ScanNet200 in Table~\ref{scannet200_val}. For ADE20K, we mainly compare our method with the plain DeepLabV3+~\cite{chen2018encoder} (DLV3P) model and the state-of-the-art~(SOTA) long-tailed segmentation framework DisAlign~\cite{zhang2021distribution}. In Table~\ref{ade_imbalance}, our method further outperforms the baseline and DisAlign on both the ResNet-50 (0.7\% mIoU improvement), and ResNet-101 (0.9\% mIoU improvement) backbone. For ScanNet200, 
our \ours{} achieves \textbf{3.1\%} mIoU improvement in mIoU using MinkowskiNet-34~\cite{choy20194d} compared with previous methods like Contrastive Scene Contexts (CSC)~\cite{scene_contrast} and CLIP~\cite{radford2021learning} based language grounded model (LG)~\cite{rozenberszki2022language}. The performance of the common (\textbf{3.6\%} mIoU improvement) and tail (\textbf{5.4\%} mIoU improvement) are improved significantly.

\vspace{3pt}
\noindent\textbf{Orthogonality to segmentation losses.} 
By now, many segmentation losses, \eg, Focal~\cite{focalloss}, Dice~\cite{milletari2016v}, SoftIOU~\cite{rahman2016optimizing}, SoftTversky~\cite{salehi2017tversky}, and Lovász~\cite{berman2018lovasz}, have been proposed for improving the segmentation results. As our method is a novel regularization for neural collapse convergence, in this part, we mainly verify its orthogonality to three famous segmentation losses, Focal loss, Dice loss, and Lovász loss. As shown in Table~\ref{loss_table} and Table~\ref{scannet200_val} (the third row),
\ours{} can be jointly trained with these three segmentation losses and further improve their performance by \textbf{1.0-3.0\%}. The above experiments clearly verify the orthogonality of \ours{} to these wide-used segmentation losses.

\begin{table}[t]
	\centering
        \renewcommand\arraystretch{1.1}
	\resizebox{\linewidth}{!}{\begin{tabular}{y{70}y{40}y{40}y{40}y{40}}
        \toprule
		\textbf{Method} & Head  & Comm. & Tail  & \textbf{All}    \\ 
  \midrule
		Minkowski.~\cite{choy20194d}            & 48.3 & 19.1  & 7.9 & 25.1  \\ 
		Ins. Samp.~\cite{rozenberszki2022language}         & 48.2 & 18.9  & 9.2 & 25.4  \\ 
		C-Focal~\cite{cui2019class, focalloss}        &   48.1    &   20.2     &   9.3   &   25.8   \\ 
		SupCon~\cite{supcontrast}    & 48.5 & 19.1  & 10.3 & 26.0 \\ 
		CSC~\cite{scene_contrast}    & 49.4 & 19.5  & 10.3 & 26.5 \\ 
		LG (CLIP)~\cite{rozenberszki2022language}            & 50.4 & 22.8  & 10.1 & 27.7\\ 
		LG~\cite{rozenberszki2022language}            & 51.5 & 22.7  & 12.5 & 28.9 \\
        \cellcolor{Gray}\textbf{\ours{}}          & \cellcolor{Gray}51.2 & \cellcolor{Gray}{22.9}\cgaphl{+}{0.2}  & \cellcolor{Gray}{17.1}\cgaphl{+}{4.6} & \cellcolor{Gray}{30.3}\cgaphl{+}{1.4} \\
        \cellcolor{Gray}\textbf{\ours{}} (Lovász)          & \cellcolor{Gray}\textbf{52.4}\cgaphl{+}{0.9} & \cellcolor{Gray}\textbf{26.2}\cgaphl{+}{3.5}  & \cellcolor{Gray}\textbf{17.9}\cgaphl{+}{5.4} & \cellcolor{Gray}\textbf{32.0}\cgaphl{+}{3.1} \\
        \bottomrule
		\end{tabular}}
        \caption{mIoU comparison on the ScanNet200 validation sets.}
	\label{scannet200_val}
\end{table}

\begin{table}[t]
	\centering
        \renewcommand\arraystretch{1.1}
	\resizebox{\linewidth}{!}{\begin{tabular}{y{70}y{40}y{40}y{40}y{40}}
        \toprule
		\textbf{Method} &  Head  & Comm. & Tail  & \textbf{All}  \\ 
  \midrule
		Minkowski. \cite{choy20194d}        &  46.3 & 15.4 & 10.2 & 25.3 \\ 
		CSC~\cite{scene_contrast}                &  45.5 & 17.1  & 7.9 & 24.9  \\ 
		LG ~\cite{rozenberszki2022language}  &  48.5 & 18.4  & 10.6 & 27.2  \\
        \cellcolor{Gray}\textbf{\ours{}} \hspace{0.0pt} (Lovász)          & \cellcolor{Gray}52.1\cgaphl{+}{3.6} & \cellcolor{Gray}{23.6}\cgaphl{+}{5.2} & \cellcolor{Gray}{15.2}\cgaphl{+}{4.6} & \cellcolor{Gray}31.7\cgaphl{+}{4.5} \\
        \cellcolor{Gray}\textbf{\ours{}}$^{*}$(Lovász)          & \cellcolor{Gray}\textbf{55.1}\cgaphl{+}{6.6} & \cellcolor{Gray}\textbf{24.7}\cgaphl{+}{6.3} & \cellcolor{Gray}\textbf{18.1}\cgaphl{+}{7.5} & \cellcolor{Gray}\textbf{34.0}\cgaphl{+}{6.8} \\
        % \cellcolor{Gray}\ours{}          & \cellcolor{Gray}\textbf{55.1}\cgaphl{+}{6.6} & \cellcolor{Gray}\textbf{24.7}\cgaphl{+}{6.3} & \cellcolor{Gray}\textbf{18.1}\cgaphl{+}{7.5} & \cellcolor{Gray}\textbf{34.0}\cgaphl{+}{6.8} \\
        \bottomrule
		\end{tabular}}
        \caption{Comparison with other methods on the ScanNet200 test set. All numbers are from the benchmark on 11$^{\rm th}$ November 2022.}
	\label{scannet200_test}
\end{table}

\subsection{Main Results}
\noindent\textbf{Comparison on ScanNet200}. To show the powerful rebalance performance of our method,
we compare with a SOTA point cloud pre-training approaches CSC~\cite{scene_contrast} and Supervised Contrastive Learning (SupCon) \cite{supcontrast}, along with SOTA LG~\cite{rozenberszki2022language} in Table~\ref{scannet200_val} and \ref{scannet200_test}. Following~\cite{rozenberszki2022language}, we use the same 3D MinkowskiNet~\cite{choy20194d} backbone and training setting for a fair comparison. 
For both the validation dataset and the test dataset, our \ours{}  consistently surpasses previous methods by a large margin on all split IoU measurements.
Concretely, \ours{} outperforms the previous best by 0.9\%, \textbf{3.6\%}, \textbf{5.4\%}, and \textbf{3.1\%} on the ScanNet200 validation dataset, \textbf{3.6\%}, \textbf{5.2\%}, \textbf{4.6\%}, and \textbf{4.5\%} on the ScanNet200 test dataset, under the head, common, tail, and all measurements, respectively. Following Mix3D~
\cite{nekrasov2021mix3d}, the SOTA 3D model on ScanNetv2, we also report the ensemble results of three \ours{} models (denoted as \ours{}$^*$) for the test set. Our \ours{} yields the highest mIoU and ranks {$\mathbf{1}^{\rm st}$} on the ScanNet200 test leaderboard. Because \ours{} enables the network to learn  equiangular and more separated feature distribution, which brings great improvements over baselines and across common and tail categories. 

\begin{table}[t]
        \small
	\centering
	\renewcommand\arraystretch{1.07}
	
	\resizebox{1.0\linewidth}{!}
	{
		\begin{tabular}{y{50}y{60}y{45}y{45}}
			\toprule
			\textbf{Method}  &Backbone  &{mIoU} (s.s.) &{mIoU} (m.s.) \\
			\midrule
			OCRNet             &HRNet-W18   &39.3 &40.8 \\
			+ \textbf{\ours{}}           &HRNet-W18   & \textbf{41.8}\cgaphl{+}{2.5} &\textbf{43.5}\cgaphl{+}{2.7} \\
			%UPerNet &            &ResNet-50  &42.0 &42.8 \\
			%UPerNet & + \ours{}          &ResNet-50  &43.3 &44.0\cgaphl{+}{1.2} \\
			DLV3P       &ResNet-50  &43.9 &44.9 \\
			+ \textbf{\ours{}}     &ResNet-50  &\textbf{45.0}\cgaphl{+}{1.1} &\textbf{46.4}\cgaphl{+}{1.5} \\
			\midrule
			OCRNet             &HRNet-W48   &43.2 &44.9 \\
			  + \textbf{\ours{}}           &HRNet-W48   &\textbf{44.5}\cgaphl{+}{1.3} &\textbf{46.1}\cgaphl{+}{1.2} \\
			UperNet             &ResNet-101 &43.8 &44.8 \\
			+ \textbf{\ours{}}          &ResNet-101 &\textbf{44.8}\cgaphl{+}{1.0} &\textbf{46.1}\cgaphl{+}{1.3} \\
			%PSPNet &            &ResNet-101 &44.39 &45.35 \\
			%PSPNet & + \ours{}           &ResNet-101 &44.65 &46.15\cgaphl{+}{0.80} \\
                DLV3P     &ResNet-101 &45.5 &46.4 \\
			+ \textbf{\ours{}}     &ResNet-101 &\textbf{46.7}\cgaphl{+}{1.2} &\textbf{48.0}\cgaphl{+}{1.6} \\
			\midrule
			UperNet      &Swin-T       &44.5 &45.8 \\
			+ \textbf{\ours{}}    &Swin-T       &\textbf{45.7}\cgaphl{+}{1.2} &\textbf{47.6}\cgaphl{+}{1.8} \\
			UperNet     &Swin-B  &50.0 &51.7 \\
			  + \textbf{\ours{}}    &Swin-B   &\textbf{51.2}\cgaphl{+}{1.2} &\textbf{52.9}\cgaphl{+}{1.2} \\
			\midrule
			UperNet     &BEiT-L &56.7 &57.0 \\
			 + \textbf{\ours{}}    &BEiT-L &\textbf{57.3}\cgaphl{+}{0.6} &\textbf{57.7}\cgaphl{+}{0.7} \\
			\bottomrule
		\end{tabular}
	}
        \caption{Comparison on ADE20K.}
        \vspace{-8.5pt}
	\label{tab:main_ade20k}
\end{table}

\vspace{3pt}
\noindent\textbf{Comparison on ADE20K}.
To show the flexibility of \ours{}, we experiment on ADE20K with various semantic segmentation head methods, \eg, UperNet~\cite{xiao2018unified}, OCRNet~\cite{yuan2020object}, and DLV3P, and different backbones, \eg, ResNet~\cite{he2016deep}, HRNet~\cite{wang2020deep}. We report the performance of both single-scale (s.s.) inference and multi-scale (m.s.) inference. As shown in Table \ref{tab:main_ade20k}, plugging our \ours{} into those methods leads to significant improvements. Specifically, for ResNet-101 or HRNet-W48, after training with \ours{}, there are 1.3\%, 1.2\%, and 1.6\% gains for UperNet, OCRNet, and DLV3P respectively. We also experiment with our method on awesome transformers like Swin~\cite{liu2021swin} and BEiT~\cite{bao2021beit}. For CNN-based models, we achieve 48.0\% mIoU with ResNet-101, surpassing the baseline by 1.6. With BEiT, the performance is improved from 57.0\% mIoU to 57.7\% mIoU. 

\begin{table}[t]
        \small
	\centering
	\renewcommand\arraystretch{1.1}
	\resizebox{1.0\linewidth}{!}
	{
		\begin{tabular}{y{50}y{60}y{45}y{45}}
			\toprule
			\textbf{Method} &Backbone  &{mIoU} (s.s.) &{mIoU} (m.s.) \\
			\midrule
			OCRNet             &HRNet-W18   &31.6 &32.4 \\
			+ \textbf{\ours{}}           &HRNet-W18   &\textbf{38.0}\cgaphl{+}{6.4} &\textbf{38.8}\cgaphl{+}{6.4} \\
			UPerNet             &ResNet-50  &39.9 &40.3 \\
			+ \textbf{\ours{}}          &ResNet-50  &\textbf{41.2}\cgaphl{+}{1.3} &\textbf{41.7}\cgaphl{+}{1.4} \\
			%PSPNet &             &ResNet-50  &40.53 &40.75 \\ 
			%PSPNet & + \ours{}           &ResNet-50  &41.92 &42.34\cgaphl{+}{1.95} \\ 
			DLV3P        &ResNet-50  &40.9 &41.5\\
			+ \textbf{\ours{}}     &ResNet-50  &\textbf{42.7}\cgaphl{+}{1.8} &\textbf{43.5}\cgaphl{+}{2.0} \\
			\midrule
			OCRNet              &HRNet-W48   &40.4 &41.7 \\
			+ \textbf{\ours{}}           &HRNet-W48   &\textbf{41.8}\cgaphl{+}{1.4} &\textbf{43.3}\cgaphl{+}{1.6} \\
			UperNet            &ResNet-101 &41.2 &41.5 \\
			+ \textbf{\ours{}}          &ResNet-101 &\textbf{42.3}\cgaphl{+}{1.1} &\textbf{42.8}\cgaphl{+}{1.3} \\
			%PSPNet &            &ResNet-101 &41.95 &42.96 \\
			%PSPNet & + \ours{}           &ResNet-101 &43.40 &43.85\cgaphl{+}{1.43} \\
                DLV3P       &ResNet-101 &42.4 &43.0 \\
			+ \textbf{\ours{}}     &ResNet-101 &\textbf{43.9}\cgaphl{+}{1.5} &\textbf{44.6}\cgaphl{+}{1.6} \\
			\midrule
			UperNet      &Swin-T       &43.8 &44.6 \\
			 + \textbf{\ours{}}    &Swin-T       &\textbf{44.5}\cgaphl{+}{0.7} &\textbf{45.2}\cgaphl{+}{0.6} \\
			UperNet      &Swin-B   &47.7 &48.6 \\
			  + \textbf{\ours{}}    &Swin-B   &\textbf{48.2}\cgaphl{+}{0.5} &\textbf{49.2}\cgaphl{+}{0.6} \\
			\bottomrule
		\end{tabular}
	}
	\caption{Comparison on COCOStuff-164K.}
        \vspace{-5pt}
        \label{tab:main_coco}
\end{table}

\vspace{3pt}
\noindent\textbf{Comparison on COCO-Stuff164K}.
On the large-scale COCO-Stuff$164$K dataset, we again demonstrate the flexibility of our \ours{}. The experimental results are summarized in Table~\ref{tab:main_coco}.
Equipped with \ours{} in training, CNN-based models, \ie, ResNets and HRNets, surpass their baselines by a large margin. Specifically, with HRNet-18 and OCRNet, our trained model outperforms the baseline by \textbf{6.4\%} mIoU. With the large CNN-based ResNet-101 and DLV3P, our model achieves 44.6\% mIoU, surpassing the baseline by 1.6\% mIoU. We also verify the effectiveness of \ours{} with the Swin transformer on COCO-Stuff164K. Experimental results with Swin-T and Swin-B show clear improvements after adopting our center collapse regularization in the training phase. Both Table~\ref{tab:main_ade20k} and Table~\ref{tab:main_coco} demonstrate the effectiveness and generalization of \ours{} on various backbones and segmentation head methods.

\vspace{3pt}
\noindent\textbf{Visual comparison.} More visual comparisons over the above three datasets are included in \textit{Appendix}~\ref{sec_vis}. We observe that \ours{} obtains more precise semantic segmentation masks for both common and tail classes.

% \section{Conclusion}
% In this paper, we explore the neural collapse structures of feature centers and classifiers for semantic segmentation. Semantic segmentation naturally brings contextual correlation and imbalanced distribution among classes. It breaks the equiangular and maximal separated structure of neural collapse for both feature centers and classifiers. However, minor classes benefit from such a structure. To preserve these properties, we introduce a center collapse loss to regularize the feature centers. {Adopting the new regularization, feature centers become more symmetric and class-separated and the performance of minor classes is greatly improved. We hope that our findings could advance future studies of 2D \& 3D imbalanced semantic segmentation. 

\section{Conclusion}
In this paper, we explore the neural collapse structures of feature centers and classifiers for semantic segmentation. Semantic segmentation naturally brings contextual correlation and imbalanced distribution among classes. It breaks the equiangular and maximal separated structure of neural collapse for both feature centers and classifiers. To preserve these properties for minor classes, we introduce a center collapse loss to regularize the feature centers. {Adopting the new regularization, feature centers become more symmetric and class-separated and the performance of minor classes is greatly improved. We hope that our findings could advance future studies of 2D \& 3D imbalanced semantic segmentation. Limitation analysis is provided in \textit{Appendix}~\ref{sec_limit}.

% \noindent\textbf{Limitation analysis.} This work presents a new regularization for imbalanced semantic segmentation. However, when the class number of the dataset is small enough or the imbalanced severity is negligible, the performance of \ours{} is quite incremental and even degraded. Detailed experiment results and analysis are shown in our appendix. 

\clearpage
%%%%%%%%% REFERENCES
{\small
\bibliographystyle{ieee_fullname}
\bibliography{egbib}
}

\clearpage
\onecolumn
\appendix 
\begin{center}
	\Large \textbf{Understanding Imbalanced Semantic Segmentation Through Neural Collapse\\ \vspace{8pt} \textit{Supplementary Material}}
\end{center}
\vspace{30pt}

\section{More Results of Neural Collapse in Semantic Segmentation}\label{sec_nc_seg}
In Sec.~\ref{nc_seg}, we discuss neural collapse in semantic segmentation. We mainly verify the equiangular property of the feature centers and classifier weights, \ie, the values of ${\rm Std}_{k \neq k'}(\cos(\hat{\z}_k,\hat{\z}_{k'}))$ and ${\rm Std}_{k \neq k'}(\cos(\hat{\w}_k,\hat{\w}_{k'}))$ for semantic segmentation are much larger than those for classification (as shown in Fig.~\ref{fig3}). In this part, we follow Papyan et al.~\cite{papyan2020prevalence} and analyze the maximally separated property of the feature centers. 
\vspace{5pt}

In~\cite{papyan2020prevalence}, feature centers approach maximal-angle equiangularity as training progresses. To measure the maximal-angle degree, Papyan et al. calculate the average of shifted cosine across all distinct classes during the whole training process. Mathematically, denote $\text{Avg}_{k,k^\prime} |(\cos(\hat{\z}_k,\hat{\z}_{k'}) + 1/(K-1)|$. As training progresses, the convergence of these values to zero implies that all cosines converge to $-1/(K-1)$. This corresponds to the maximum separation possible for globally centered, equiangular vectors. We carry on a similar experiment on various semantic segmentation datasets. As shown in Fig.~\ref{nc_avg}, the average values for semantic segmentation are two to three times larger than those for classification during the terminal phase of training. It means the feature centers in semantic segmentation are farther away from the maximal separation.

\begin{figure*}[h]
	\begin{center}
		\includegraphics[width=1.0\textwidth]{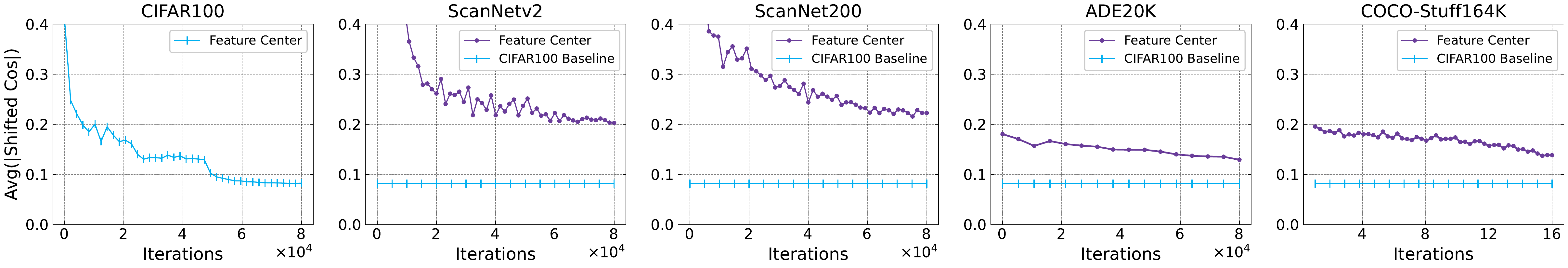}
		\vspace{-10pt}
		\begin{subfigure}[t]{0.196\linewidth}
			\centering
			\vspace{-14pt}
			\caption{}
			\label{fig5a}
		\end{subfigure}
		\hfill
		\begin{subfigure}[t]{0.196\linewidth}
			\centering
			\vspace{-14pt}
			\caption{}
			\label{fig5b}
		\end{subfigure}
		\hfill
		\begin{subfigure}[t]{0.196\linewidth}
			\centering
			\vspace{-14pt}		
			\caption{}
			\label{fig5c}
		\end{subfigure}
		\hfill
		\begin{subfigure}[t]{0.196\linewidth}
			\centering
			\vspace{-14pt}
			\caption{}
			\label{fig5d}
		\end{subfigure}
		\hfill
		\begin{subfigure}[t]{0.196\linewidth}
			\centering
			\vspace{-14pt}
			\caption{}
			\label{fig5e}
		\end{subfigure}
		\caption{We plot in the vertical axis of each cell the quantities $\text{Avg}_{k,k^\prime} |(\cos(\hat{\z}_k,\hat{\z}_{k'}) + 1/(K-1)|$ on a classification dataset (a) and different semantic segmentation datasets (b-e). The feature centers in semantic segmentation are more difficult to reach the maximally separated structure of neural collapse than classification.}
		\label{nc_avg}
	\end{center}
	\vspace{-5pt}
\end{figure*}

Taken together, Fig.~{\ref{fig3}} and Fig.~\ref{nc_avg} give evidence that both the feature centers and the classifier weights for semantic segmentation are harder than those for classification to converge to an equiangular and maximal separated structure. Moreover, the feature centers suffer a more difficult issue of converging to an ETF structure (neural collapse) than the classifier weights. We point out that, different from classification, semantic segmentation naturally brings contextual correlation and imbalanced distribution among classes, which may break the symmetric structure of neural collapse for both feature centers and classifiers. Noting that such an equiangular and maximally separated feature distribution will bring great benefit to the minor classes, we thus propose a feature center collapse regularizer to achieve this goal.

\vspace{5pt}

\paragraph{Experiment setting details of Fig.~{\ref{fig3}} and Fig.~\ref{nc_avg}.} For the CIFAR-100 classification task, we train a ResNet-101~\cite{he2016deep} model. For the ScanNetv2 and ScanNet200 point cloud semantic segmentation tasks, we train two MinkoswskiNet-34~\cite{choy20194d} models. For the ADE20K and COCO-Stuff164K image semantic segmentation tasks, we train two DeepLabv3+~\cite{chen2018encoder} ResNet-101 models. All training hyperparameters and optimizers are followed the conventional training setting. The total number of iterations for the above five tasks are 80K, 80K, 80K, 80K, and 160K, respectively. All neural collapse statistics computations are following~\cite{papyan2020prevalence}.

\clearpage
\section{Proof for Lemma \ref{lemma}: Equiangular \& Maximal Separated Property}\label{sec_proof}

\paragraph{Sufficiency.} Since $\mathbf{W}$ is a normalized matrix, \ie, $\mathbf{W}=[\mathbf{w}_1, ...,\mathbf{w}_K]\in\mathbb{R}^{d\times K}, d \geq K, \mathbf{w}_k^\top\mathbf{w}_k=1, \forall k=1, ..., K$, we have $\cos(\w_k, \w_{k'}) = \w_k^\top\w_{k'}$. We construct the following form:
\begin{equation}
	\label{eq2}	\left(\mathbf{W}\mathbf{1}_K\right)^\top\left(\mathbf{W}\mathbf{1}_K\right) = \mathbf{1}^\top_K\mathbf{W}^\top\mathbf{W}\mathbf{1}_K=\underbrace{\sum_{k=1}^{K}\sum_{k'=1}^{K}\w_k^\top\w_{k'}}_{K^2 \text{ terms}} = \underbrace{\sum_{k=1}^{K}\w_k^\top\w_{k}}_{K \text{ terms}} + \underbrace{\sum_{k \neq k'}\w_k^\top\w_{k'}}_{K(K-1) \text{ terms}} \geq 0.
\end{equation}
Recalling that $\mathbf{w}_k^\top\mathbf{w}_k=1, \forall k=1, ..., K,$ we have $\sum_{k \neq k'}\w_k^\top\w_{k'} \geq -K$. Thus we can get:
\begin{equation}
	\label{eq3}
	\max_{k \not= k'} \cos(\w_k, \w_{k'}) =\max_{k \not= k'} \w_k^\top\w_{k'} \geq -\frac{K}{K(K-1)} = -\frac{1}{K-1}.
\end{equation}
Thus, the maximal separation value, $\max_{k \not= k'} \cos(\w_k, \w_{k'})$, is greater or equals to $-\frac{1}{K-1}$. In the following, we will give a proof that $\max_{k \not= k'} \cos(\w_k, \w_{k'})=-\frac{1}{K-1}$. Here we let:
\begin{equation}
	\label{eq4} \W=\sqrt{\frac{K}{K-1}}\mathbf{U}\left(\mathbf{I}_K-\frac{1}{K}\mathbf{1}_K\mathbf{1}_K^\top\right), 
\end{equation}
where $\mathbf{U}$ is a rotation matrix satisfied $\mathbf{U}^\top\mathbf{U}=\mathbf{I}_K$. We can calculate:
\begin{equation}
	\label{eq5}
	\W^\top\W=\frac{K}{K-1}\left(\mathbf{I}_K-\frac{1}{K}\mathbf{1}_K\mathbf{1}_K^\top\right)^\top\left(\mathbf{I}_K-\frac{1}{K}\mathbf{1}_K\mathbf{1}_K^\top\right)=\begin{bmatrix}
		1  & -\frac{1}{K-1} & \cdots & -\frac{1}{K-1} \\
		-\frac{1}{K-1} & 1  & \cdots & -\frac{1}{K-1} \\
		\vdots                  & \vdots                  & \ddots & \vdots                  \\
		-\frac{1}{K-1} & -\frac{1}{K-1} & \cdots & 1
	\end{bmatrix}.
\end{equation}
According Eq.~\eqref{eq5}, it means that  $\mathbf{w}_k^\top\mathbf{w}_k=1, \forall k=1, ..., K,$  and  $\mathbf{w}_k^\top\mathbf{w}_{k'}=-\frac{1}{K-1}, \forall k\neq k'.$ When $\W$ satisfies Eq.~\eqref{eq4} and is simplex ETF structured, the equality in Eq.~\eqref{eq3} holds. Obviously, it enjoys the equiangular property, \ie, $\forall k \neq k', \cos(\w_k, \w_{k'})=-\frac{1}{K-1}$, where $\frac{1}{K-1}$ is a constant.

\paragraph{Necessity.} Since $\max_{k\ne k'} \cos(\mathbf{w}_k, \mathbf{w}_{k'})=-\frac{1}{K-1}$, we have $\cos(\mathbf{w}_k, \mathbf{w}_{k'})\le -\frac{1}{K-1}, \forall k\ne k'$, and then $\sum_{k\ne k'}\cos(\mathbf{w}_k, \mathbf{w}_{k'})\le -K$.
Given Eq. (\ref{eq2}), we have $\sum_{k\ne k'}\cos(\mathbf{w}_k, \mathbf{w}_{k'})\ge -K$. As a result, we have $\sum_{k\ne k'}\cos(\mathbf{w}_k, \mathbf{w}_{k'})=-K$, and the following equation,

\begin{equation}
	\W^\top\W=\begin{bmatrix}
		1  & -\frac{1}{K-1} & \cdots & -\frac{1}{K-1} \\
		-\frac{1}{K-1} & 1  & \cdots & -\frac{1}{K-1} \\
		\vdots                  & \vdots                  & \ddots & \vdots                  \\
		-\frac{1}{K-1} & -\frac{1}{K-1} & \cdots & 1
	\end{bmatrix}.
	\label{w_eq}
\end{equation}
We define $\M=\sqrt{\frac{K-1}{K}}\W$. We have:
$$\M^\top\M=\frac{K-1}{K}\W^\top\W=\frac{K-1}{K}\begin{bmatrix}
	1  & -\frac{1}{K-1} & \cdots & -\frac{1}{K-1} \\
	-\frac{1}{K-1} & 1  & \cdots & -\frac{1}{K-1} \\
	\vdots                  & \vdots                  & \ddots & \vdots                  \\
	-\frac{1}{K-1} & -\frac{1}{K-1} & \cdots & 1
\end{bmatrix}=\begin{bmatrix}
	\frac{K-1}{K}  & -\frac{1}{K} & \cdots & -\frac{1}{K} \\
	-\frac{1}{K} & \frac{K-1}{K}  & \cdots & -\frac{1}{K} \\
	\vdots                  & \vdots                  & \ddots & \vdots                  \\
	-\frac{1}{K} & -\frac{1}{K} & \cdots & \frac{K-1}{K}
\end{bmatrix}=\mathbf{I}_K-\frac{1}{K}\mathbf{1}_K\mathbf{1}_K.$$ 
Note that $\mathbf{I}_K-\frac{1}{K}\mathbf{1}_K\mathbf{1}_K$ is the centering matrix $\mathbf{C}_K$, which has the eigenvalue 1 of multiplicity $K-1$ and eigenvalue 0 of multiplicity 1.
Note that $\mathbf{C}_K^2=\mathbf{C}_K^\top\mathbf{C}_K=\mathbf{C}_K$, we conclude that, $\mathbf{C}_K = \mathbf{V} {\bm{\Sigma}} \mathbf{V}^\top$, where $\mathbf{V} = \left[\mathbf{V}', \frac{1}{\sqrt{K}}\mathbf{1}_K\right]$, and $\mathbf{V}'(\mathbf{V}')$ is the projector on the $(K-1)$-dimension subspace perpendicular to $\frac{1}{\sqrt{K}}\mathbf{1}_K$, \ie,
\[
\mathbf{V}'(\mathbf{V}')^\top = \mathbf{C}_K, \quad {\bm{\Sigma}} = \mqty[\dmat{1,1,\ddots,0}].
\]
Due to  the uniqueness of the orthogonal projector, $\M$ shares the same column space as $\mathbf{C}_K$, namely, we have,
\[
\M=\mathbf{U}_M{\bm{\Sigma}}\left[\mathbf{V}'\mathbf{Q}_{k-1}, \frac{1}{\sqrt{K}}\mathbf{1}_K\right]^\top = \mathbf{U}_M{\bm{\Sigma}} \mathbf{Q}^\top \mathbf{V}^\top, \quad \mathbf{Q}= \mqty[\dmat{\mathbf{Q}_{k-1},1}],
\]
where $\mathbf{Q}_{K-1} \in \mathbb{R}^{K-1 \times K-1}$ and $\mathbf{U}_M \in \mathbb{R}^{d \times K}$  are the arbitrary orthogonal matrices.
It is easy to verify that $\mathbf{Q}\mathbf{Q}^\top = \mathbf{I}$ and ${\bm{\Sigma}} \mathbf{Q}^\top  = \mathbf{Q}^\top {\bm{\Sigma}}$, thus we have,
\[
\M= \mathbf{U}_M{\bm{\Sigma}} \mathbf{Q}^\top \mathbf{V}^\top = \mathbf{U}_M \mathbf{Q}^\top {\bm{\Sigma}} \mathbf{V}^\top =\mathbf{U}_M \mathbf{Q}^\top \mathbf{V}^\top \mathbf{V} {\bm{\Sigma}} \mathbf{Q}^\top \mathbf{V}^\top = \mathbf{U}_M \mathbf{Q}^\top \mathbf{V}^\top \mathbf{C}_K.
\]
We let $\mathbf{U} = \mathbf{U}_M \mathbf{Q}^\top \mathbf{V}^\top$, and we can get,
\[
\mathbf{U}_M^\top \mathbf{U}_M =\mathbf{V} \mathbf{Q} \mathbf{U}^\top \mathbf{U} \mathbf{Q}^\top \mathbf{V}^\top = \mathbf{I}_K.
\]
Hence, we can conclude that $\M = \mathbf{U} \mathbf{C}_K$, where $\mathbf{U}$ is an orthogonal matrix. 
Finally, due to $\M=\sqrt{\frac{K-1}{K}}\W$,we prove the solution of Eq.~\eqref{w_eq} is 
$$\W = \sqrt{\frac{K}{K-1}}\M=\sqrt{\frac{K}{K-1}}\U\left(\mathbf{I}_K-\frac{1}{K}\mathbf{1}_K\mathbf{1}_K^\top\right).$$
In summary, a normalized matrix is maximal equiangular separated \textbf{\textit{if and if only}} the matrix is a simplex ETF. $\hfill\square$

\clearpage
\section{Detailed Empirical Evidence of Better Rebalance in \ours{}}\label{sec_emp_more}

In Sec.~\ref{sec_emp}, we list the main conclusions about better rebalance in \ours{}. Here, we give a more detailed discussion. 

\paragraph{Reduction the degree of imbalance.} In our center collapse regularizer, we get two important parameters, feature centers $\bar{\mathbf{Z}}$ and center's labels $\bar{\mathbf{y}}$. Semantic segmentation datasets naturally suffer a more severe class imbalance issue. However, We observe that the degree of feature center imbalance (the distribution of $\bar{\y}$) is greatly relieved than that of point/pixel imbalance (the distribution of $\y$). We collect these two statistics from the most popular semantic segmentation benchmarks. Following the convention \cite{cui2019class,liu2019large}, we calculate the imbalance factor (IF) as $\beta= \frac{n_{\rm max}}{n_{\rm min}}$, where $n_{\rm max}$ and $n_{\rm min}$ are the numbers of training samples for the most frequent class and the least frequent class. As shown in  Fig.~\ref{fig5}, for ADE20K, the point/pixel imbalance factor (PIF) is nearly three times the center imbalance factor (CIF). For COCO-Stuff$164$K, PIF is nearly five times RIF. For ScanNet-v2, PIF is nearly $10$ times CIF.  For ScanNet200, PIF is nearly $60$ times CIF. As our method relieves the class imbalance issue via the center collapse branch, it conveniently benefits from less imbalance than point/pixel would, and this in turn promotes more effective rebalancing for semantic segmentation.

\begin{figure*}[h]
	\begin{center}
		\includegraphics[width=1.0\textwidth]{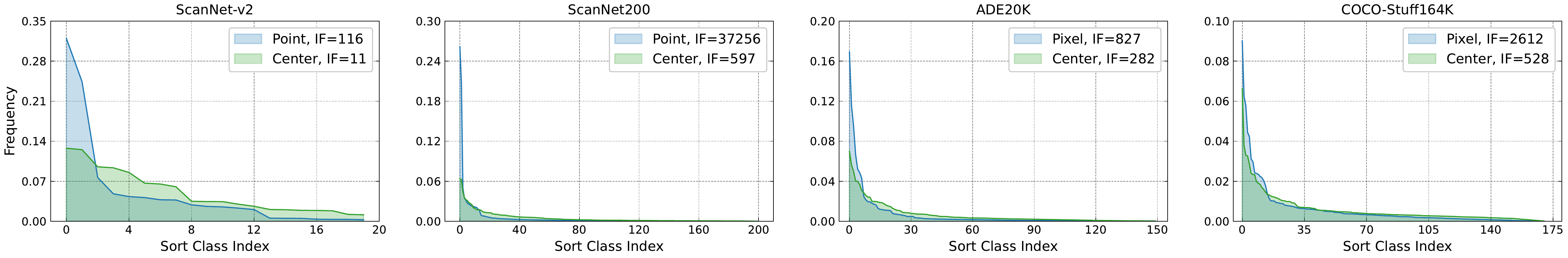}
		\caption{Comparison between point/pixel imbalance and class center imbalance.
			Histogram statistics of point/center frequency over sorted class indexes on ScanNet-v2, ScanNet200 (left), and pixel/center frequency on ADE20K, COCO-Stuff164K (right). The imbalance factor (IF) is defined as $\frac{n_{\rm max}}{n_{\rm min}}$, where $n_{\rm max}$ and $n_{\rm min}$ are the numbers of samples (points, pixels, centers) for the most frequent class and the least frequent class of dataset. It can be seen that the center imbalance is greatly alleviated compared with the point/pixel imbalance.}
		\label{fig5}
	\end{center}
	\vspace{-10pt}
\end{figure*}

\paragraph{Reducing the contextual correlation and improving the effective number.}
In this part, we will analyze the correlation influence in the center regularization branch. For any $\bar{\z}_k$, it relates to $n_k$ points/pixels from the original input in the point/pixel branch. Thus, $\bar{\z}_k$ is a higher-level semantic feature and contains more general information compared with the point/pixel-level feature $\z_i$. For the point/pixel-level pair $(\z, \y)$, as we mentioned in the introduction part and Fig.~\ref{fig2}, plenty of common information (similar color, close position) is shared among neighboring pixels/points. Neighboring points/pixels are highly correlated, which leads to the class accuracy having a lower correlation with the number of points/pixels. By contrast, the class center pair $(\bar{\z}, \bar{\y})$, is a higher and more global level semantic representation, which increases diversity between different samples. To proof this, we calculate the Pearson correlation coefficient~\cite{benesty2009pearson} between the class accuracy $\mathbf{a} \in \mathbb{R}^K$ and the class frequency $\f=[\f_1, ..., \f_K] \in \mathbb{R}^K, \f_k = \frac{n_k}{n_{\rm max}}$:
\begin{equation}
	\rho_{\mathbf{a}, \f} = \frac{{\rm Cov}(\mathbf{a}, \f)}{\sigma(\mathbf{a})\cdot\sigma(\f)} = \frac{\mathbb{E}[(\mathbf{a}-\mu(\mathbf{a}))(\f-\mu(\f))]}{\sigma(\mathbf{a})\cdot\sigma(\f)}, \notag
\end{equation}
where $\mu(\cdot)$, $\sigma(\cdot)$, ${\rm Cov}(\cdot)$ and $\mathbb{E}(\cdot)$ are the functions for mean, standard deviation, covariance, and expectation, respectively.

\begin{figure*}[h] 
	\centering
	\includegraphics[width=1.0\linewidth]{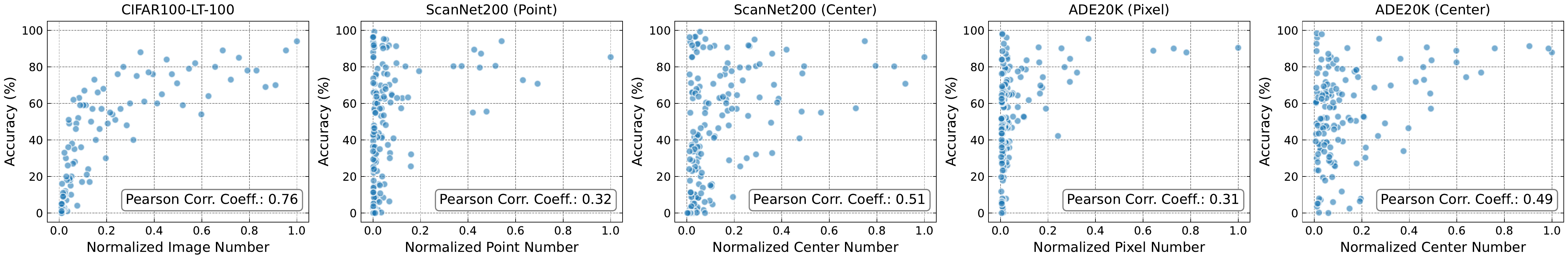}
	\vspace{-10pt}
	\begin{subfigure}[t]{0.196\linewidth}
		\centering
		\vspace{-14pt}
		\caption{}
		\label{fig2a}
	\end{subfigure}
	\hfill
	\begin{subfigure}[t]{0.196\linewidth}
		\centering
		\vspace{-14pt}
		\caption{}
		\label{fig2b}
	\end{subfigure}
	\hfill
	\begin{subfigure}[t]{0.196\linewidth}
		\centering
		\vspace{-14pt}		
		\caption{}
		\label{fig2c}
	\end{subfigure}
	\hfill
	\begin{subfigure}[t]{0.196\linewidth}
		\centering
		\vspace{-14pt}
		\caption{}
		\label{fig2d}
	\end{subfigure}
	\hfill
	\begin{subfigure}[t]{0.196\linewidth}
		\centering
		\vspace{-14pt}
		\caption{}
		\label{fig2e}
	\end{subfigure}
	\caption{Illustration of effective number for recognition and semantic segmentation. (a): Class accuracy is \textbf{positively} correlated with the class image number (normalized by the maximum image number among classes). (b) and (d): Class accuracy is \textbf{weakly} correlated with the class point/pixel number (normalized by the maximum point/pixel number among classes). (c) and (e): Class accuracy is \textbf{relatively} correlated with the class center number (normalized by the maximum center number among classes).}
	\label{fig2}
\end{figure*}

In the bottom right part of Fig.~\ref{fig2}, we list the Pearson coefficients between class accuracy and image frequency on CIFAR-100-LT, pixel frequency on ADE20K, point frequency on ScanNet200, and feature center frequency on ScanNet200. The correlation between class accuracy and image frequency is the largest. Many previous studies proposed class-balanced strategies~\cite{cui2019class, huang2019deep, Cao_nips, balsfx, logitsadj} guided by class frequency, and achieved good results In long-tailed image recognition. Feature center frequency has a stronger correlation with class accuracy than point/pixel frequency, as it eliminates the effects of correlations among neighboring pixels, further indicating its greater suitability for semantic segmentation rebalancing.
\vspace{4pt}

Different from the image classification task (CIFAR100-LT, see Fig.~\ref{fig2a}), the category accuracy has a \textbf{low correlation} with the number of pixels/points (ScanNet200 and ADE20K, Fig.~\ref{fig2b} and \ref{fig2d}). The reason may lie in the effective number of training samples \cite{cui2019class}, which is defined as the number of samples that contribute. Unlike image classification where each image is a unique instance, two pixels in a similar context may contribute similarly in training semantic segmentation models, and thus the actual number of pixels may not necessarily be the number of effective pixels. Any two different images of the same class offer significantly different training signals, whereas in semantic segmentation two pixels of the same semantic class may contribute similarly. By contrast, the center regularization branch significantly reduces the imbalance severity to increase the effective number, as shown in Fig.~\ref{fig2c} and \ref{fig2e}.

\section{Datasets Description and Implementation Details}\label{sec_data}
\subsection{Datasets}

\paragraph{ScanNet200.}
The ScanNet~\cite{scannet} Benchmark has provided an active online benchmark evaluation for 3D semantic segmentation, but only considers 20 class categories, which is insufficient to capture the diversity of many real-world environments. Thus, Rozenberszki et al. presented the ScanNet200~\cite{rozenberszki2022language} Benchmark Benchmark for 3D semantic segmentation with 200 class categories, an order of magnitude more than the previous. In order to better understand performance under the natural class imbalance of the ScanNet200 benchmark, they further split the 200 categories into sets of 66, 68, and 66 categories, based on the frequency number of labeled surface points in the train set: head, common, and tail respectively.
\vspace{4pt}

\paragraph{ADE20K.}
ADE20K \cite{zhou2017scene} is a challenging dataset often used to validate transformer-based neural networks on downstream tasks such as semantic segmentation. It contains 22K densely annotated images with 150 fine-grained semantic concepts. The training and validation sets consist of 20K and 2K images, respectively.
\vspace{4pt}

\paragraph{COCO-Stuff164K.}
COCO-Stuff164K \cite{caesar2018coco} is a large-scale scene understanding benchmark that can be used for evaluating semantic segmentation, object detection, and image captioning. It includes all 164K images from COCO 2017. The training and validation sets contain 118K and 5K images, respectively. It covers 171 classes: 80 thing classes and 91 stuff classes.

\subsection{Implementation Details}

\paragraph{3D semantic segmentation.} We implement \ours{} based on the CSC~\cite{scene_contrast} and LG~\cite{rozenberszki2022language} codebase.
We use the SGD optimizer with a batch size of 32 for a total of
20K steps. The initial learning rate is around 0.1, with polynomial
decay with a power of 0.9. For all experiments, we use data
parallel on 8 GPUs. For the backbone, we follow~\cite{scene_contrast, rozenberszki2022language} use the same and widely-used voxel-based network, MinkowskiNet-34~\cite{choy20194d}.
\vspace{5pt}

\paragraph{2D semantic segmentation.} 
We implement the proposed method in the mmsegmentation codebase~\cite{mmseg2020} and follow the commonly used training settings for each dataset. More details are described in the following.
\vspace{4pt}

For backbones, we use CNN-based ResNet-50c and ResNet-101c, which replace the first $7 \times 7$ convolution layer in the original ResNet-50 and ResNet-101 with three consecutive $3 \times 3$ convolutions. Both are popular in the semantic segmentation community \cite{DBLP:conf/cvpr/ZhaoSQWJ17}. For OCRNet, we adopt HRNet-W18 and HRNet-W$48$ \cite{wang2020deep}. For transformer-based neural networks, we adopt the popular Swin transformer \cite{liu2021swin} and BEiT \cite{bao2021beit}. \textsc{BEiT} achieves the most recent state-of-the-art performance on the ADE20K validation set. It is worth noting that Swin-B is pre-trained on ImageNet-22K.
\vspace{4pt}

With CNN-based models, we use SGD and the poly learning rate schedule \cite{DBLP:conf/cvpr/ZhaoSQWJ17} with an initial learning rate of 0.01 and a weight decay of 0.001. If not stated otherwise, we use a crop size of $512 \times 512$, a batch size of $16$, and train all models for 160K iterations on ADE20K and 320K iterations on COCO-Stuff164K. For the Swin transformer and BEiT, we use their default optimizer, learning rate setup, and training schedule.  
In the training phase, the standard random scale jittering between 0.5 and 2.0, random horizontal flipping, random cropping, as well as random color jittering are used as data augmentations \cite{mmseg2020}. For inference, we report the performance of both single-scale (s.s.) inference and multi-scale (m.s.) inference with horizontal flips and scales of 0.5, 0.75, 1.0, 1.25, 1.5, 1.75.

\clearpage
\section{Visual Comparison}\label{sec_vis}
In this section, we demonstrate the advantages of \ours{} with quantitative visualizations on ScanNet200, ADE20K, and COCOStuff164K shown in Figures \ref{scannet200}, \ref{fig:visual_comparison_ade20k_v1},  and \ref{fig:visual_comparison_cocostuff_v1}, respectively.  We observe that \ours{} well captures the contextual and geometry information and obtains more precise semantic segmentation masks for both common and tail classes.

\begin{figure*}[h]
	\begin{center}
		\begin{subfigure}[t]{0.24\linewidth}
			\centering
			\includegraphics[width=0.99\linewidth]{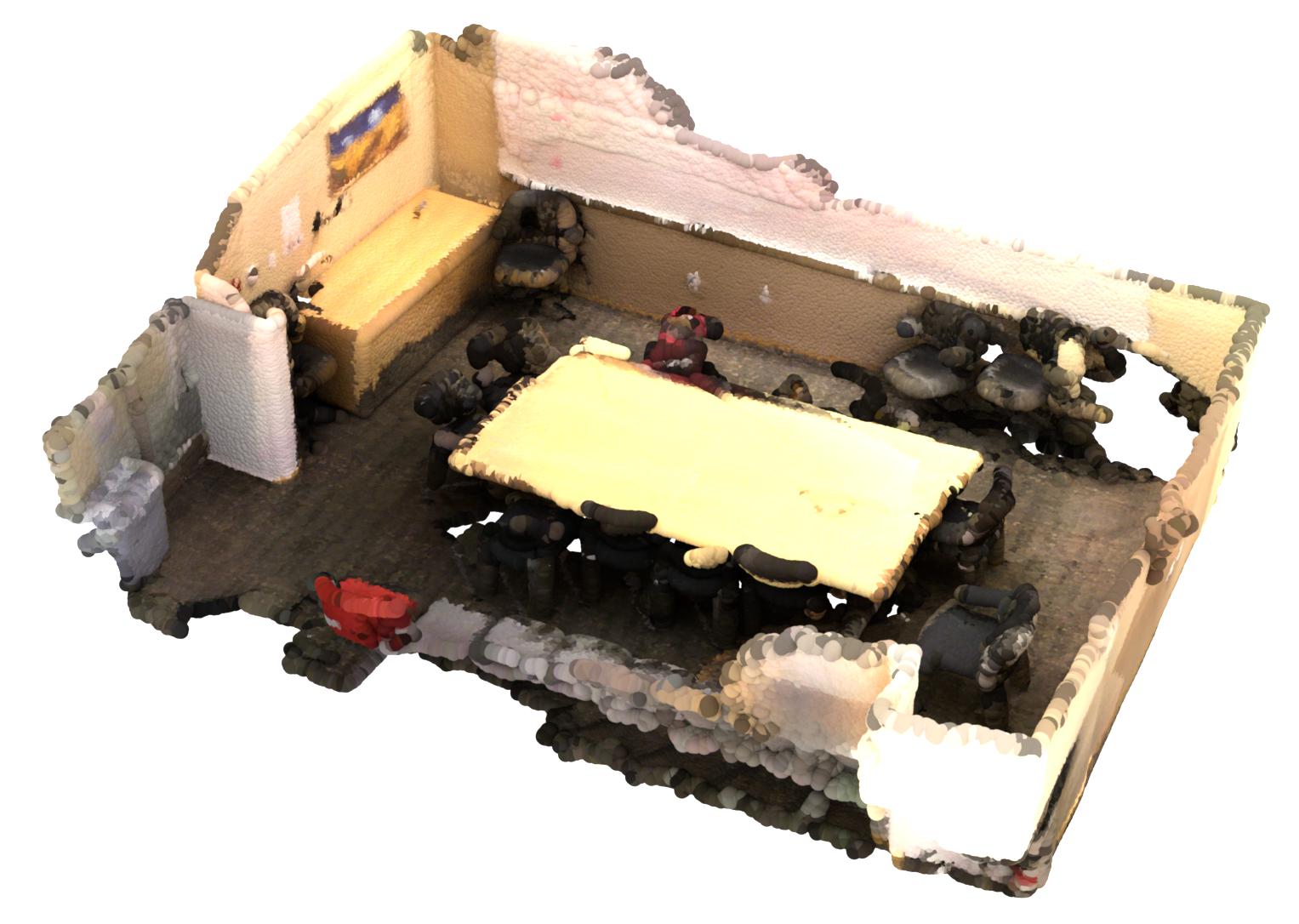}
			\caption*{}
		\end{subfigure}
		\hfill
		\begin{subfigure}[t]{0.24\linewidth}
			\centering
			\includegraphics[width=0.99\linewidth]{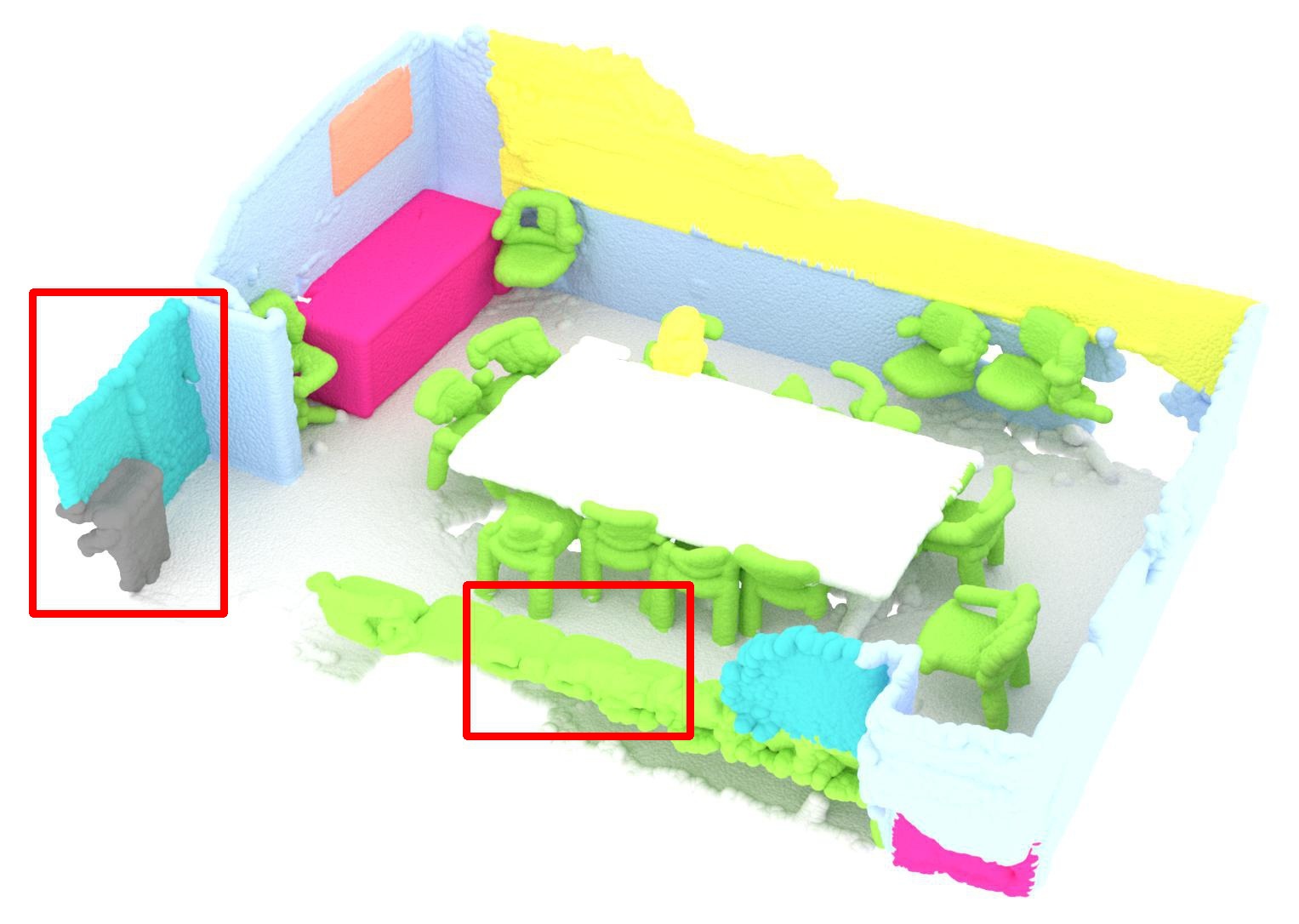}		
			\caption*{}
		\end{subfigure}
		\hfill
		\begin{subfigure}[t]{0.24\linewidth}
			\centering
			\includegraphics[width=0.99\linewidth]{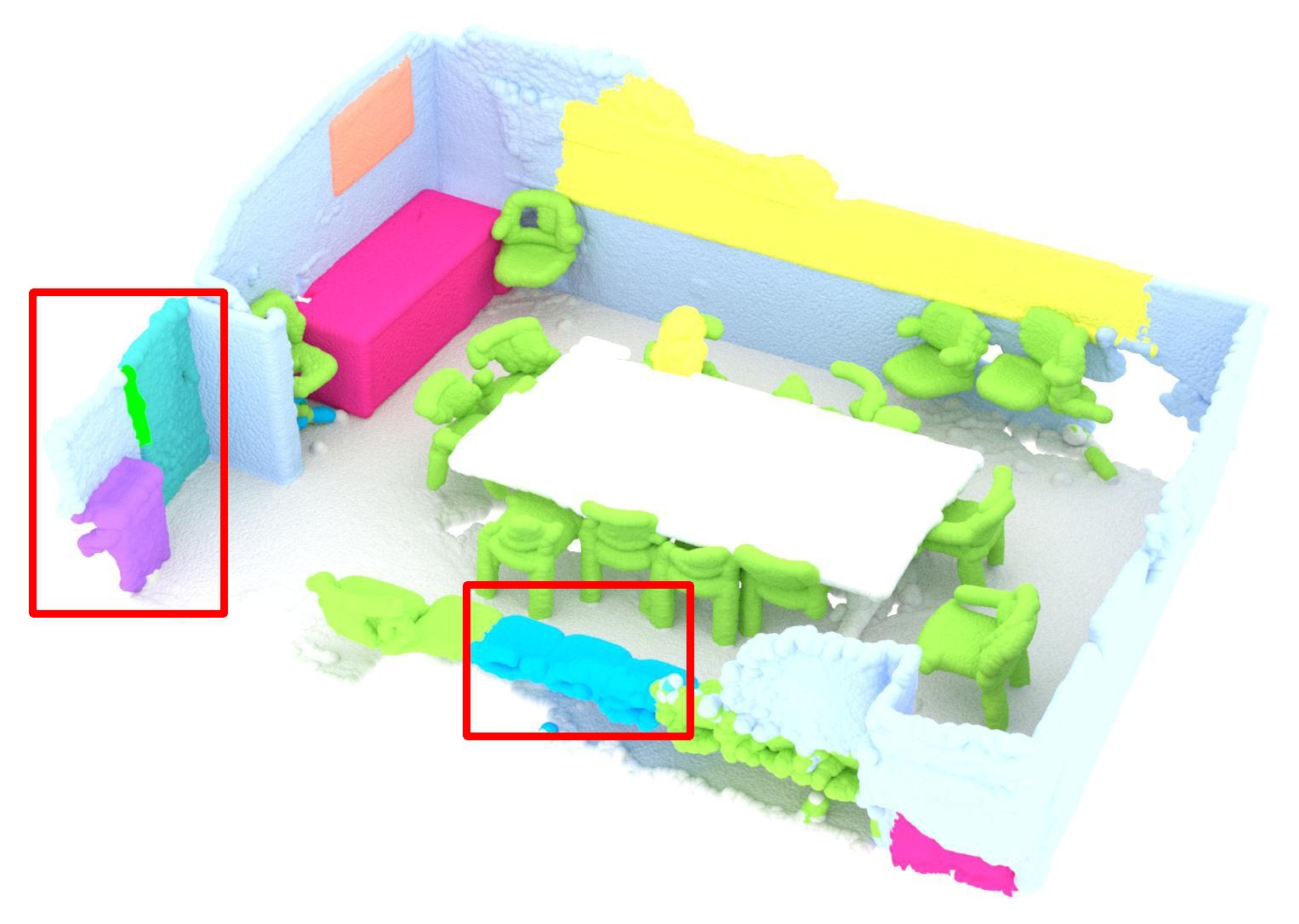}
			\caption*{}
		\end{subfigure}
		\hfill
		\begin{subfigure}[t]{0.24\linewidth}
			\centering
			\includegraphics[width=0.99\linewidth]{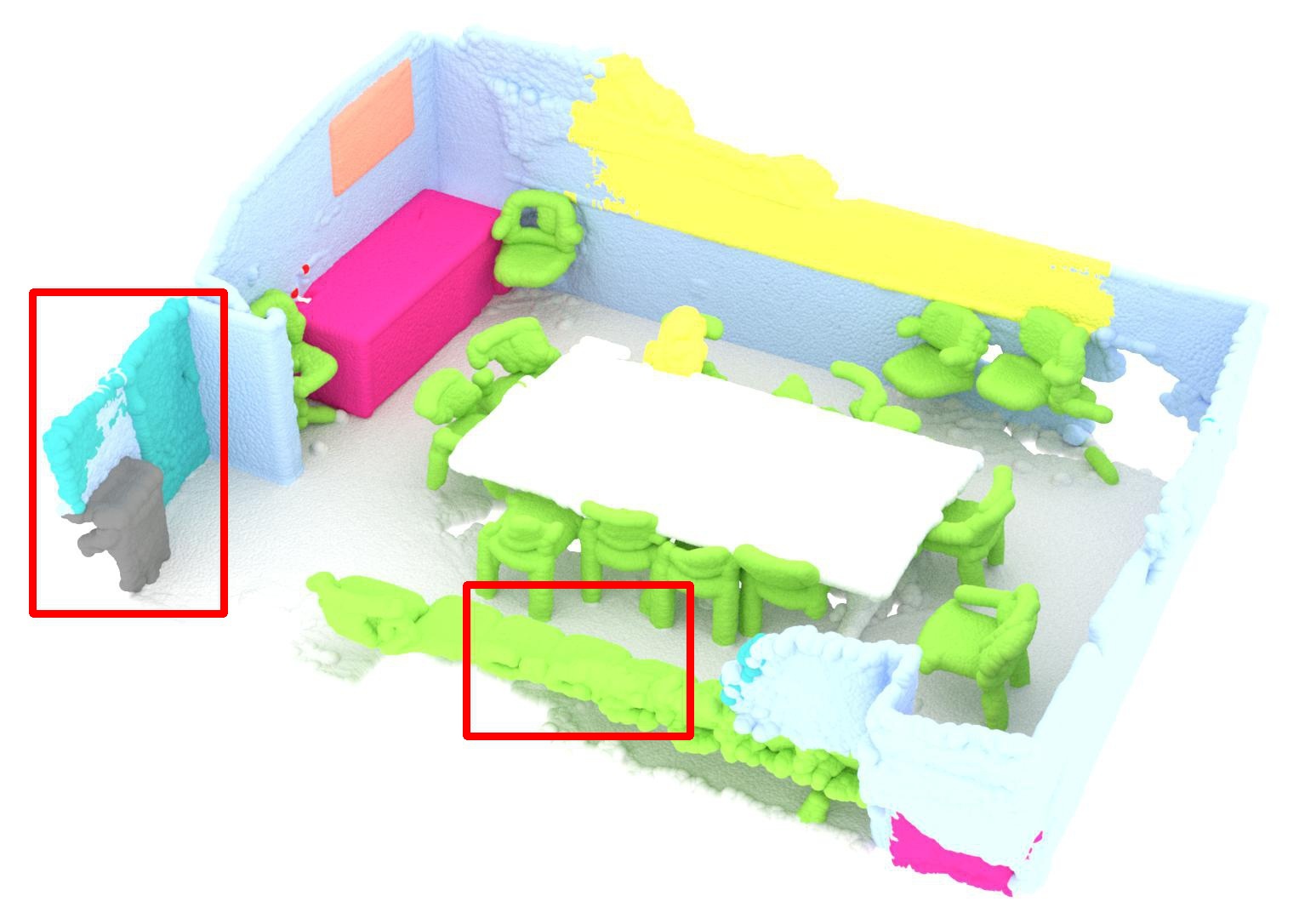}
			\caption*{}
		\end{subfigure}

		\begin{subfigure}[t]{0.24\linewidth}
			\centering
			\includegraphics[width=0.99\linewidth]{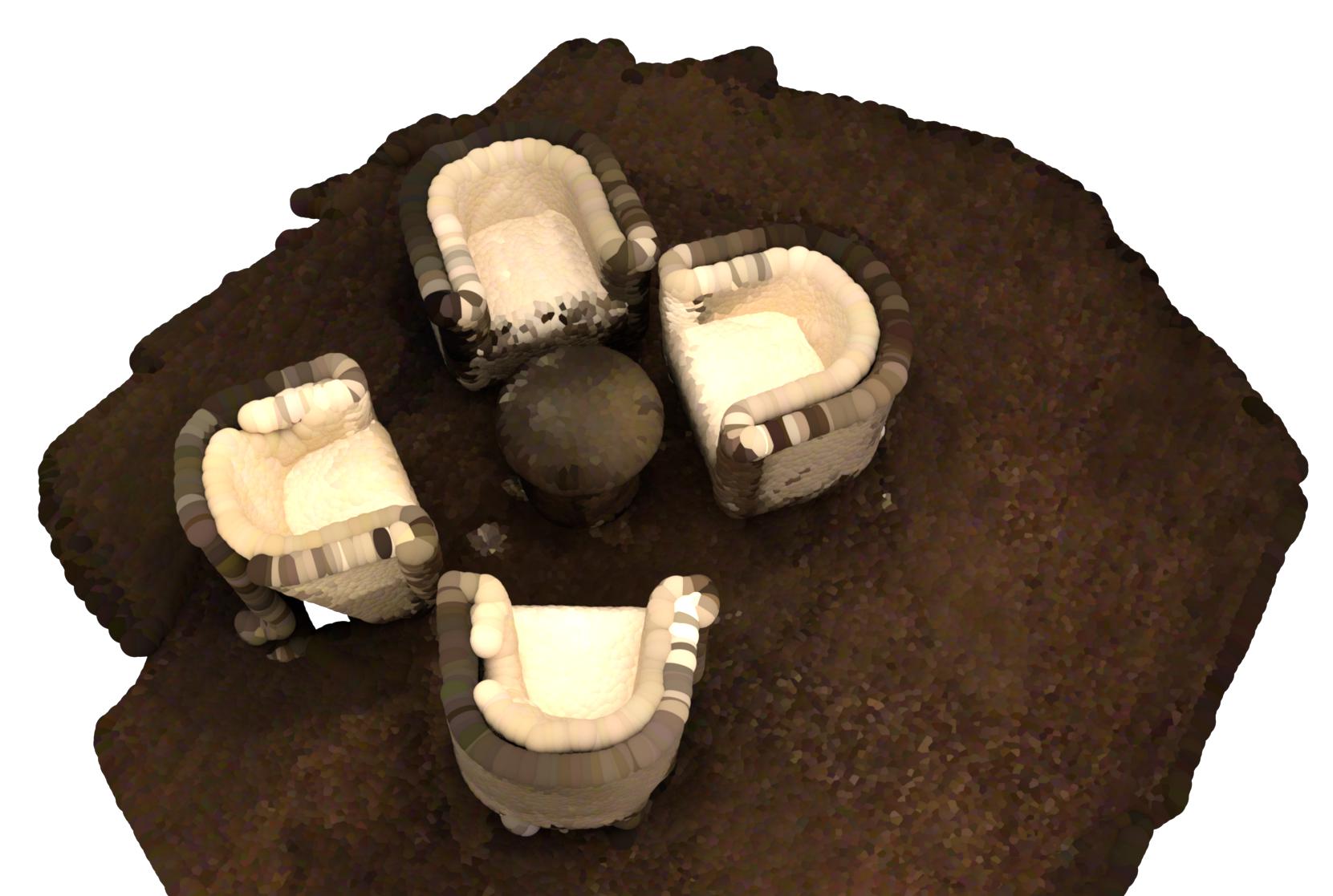}
			\caption*{}
		\end{subfigure}
		\hfill
		\begin{subfigure}[t]{0.24\linewidth}
			\centering
			\includegraphics[width=0.99\linewidth]{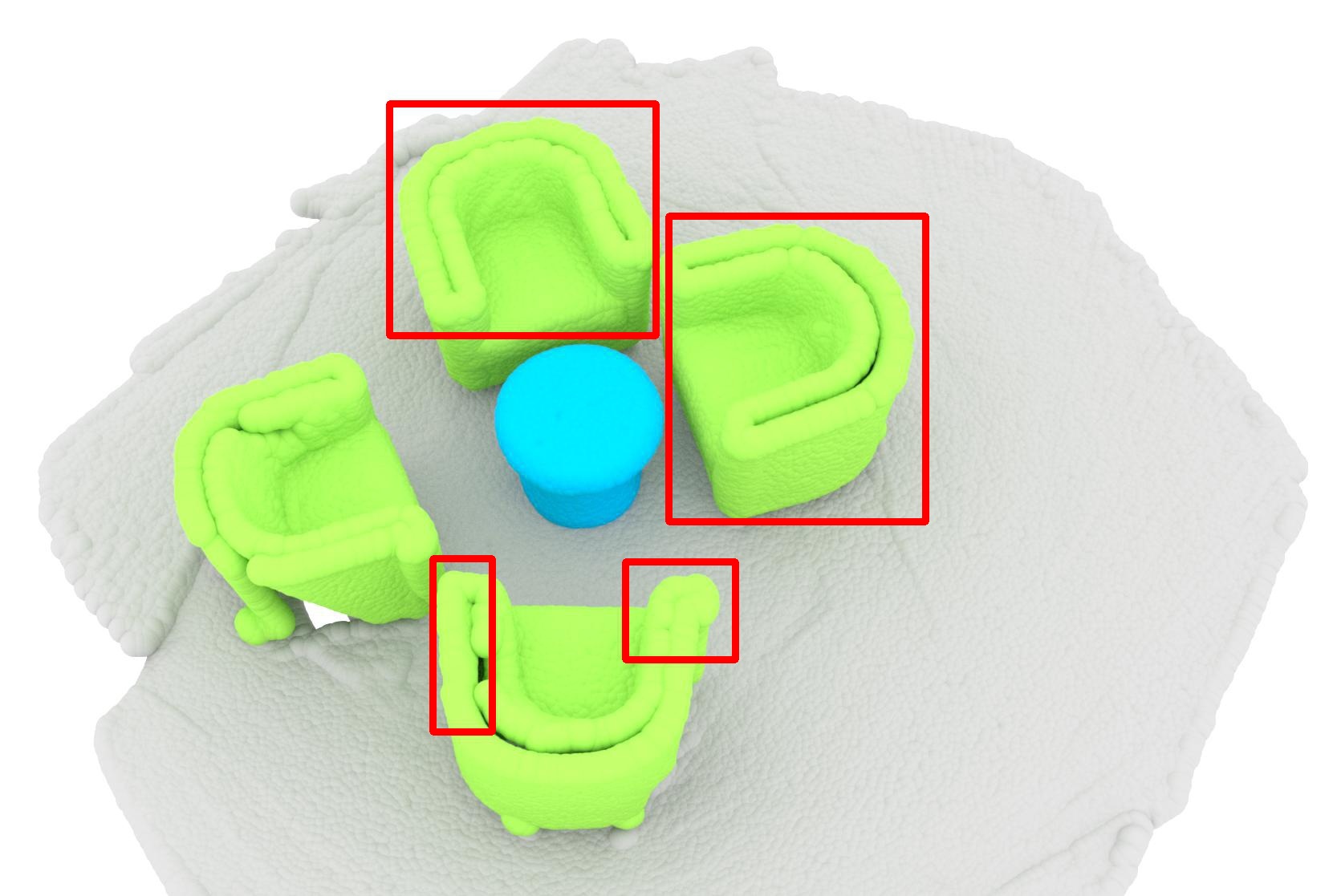}		
			\caption*{}
		\end{subfigure}
		\hfill
		\begin{subfigure}[t]{0.24\linewidth}
			\centering
			\includegraphics[width=0.99\linewidth]{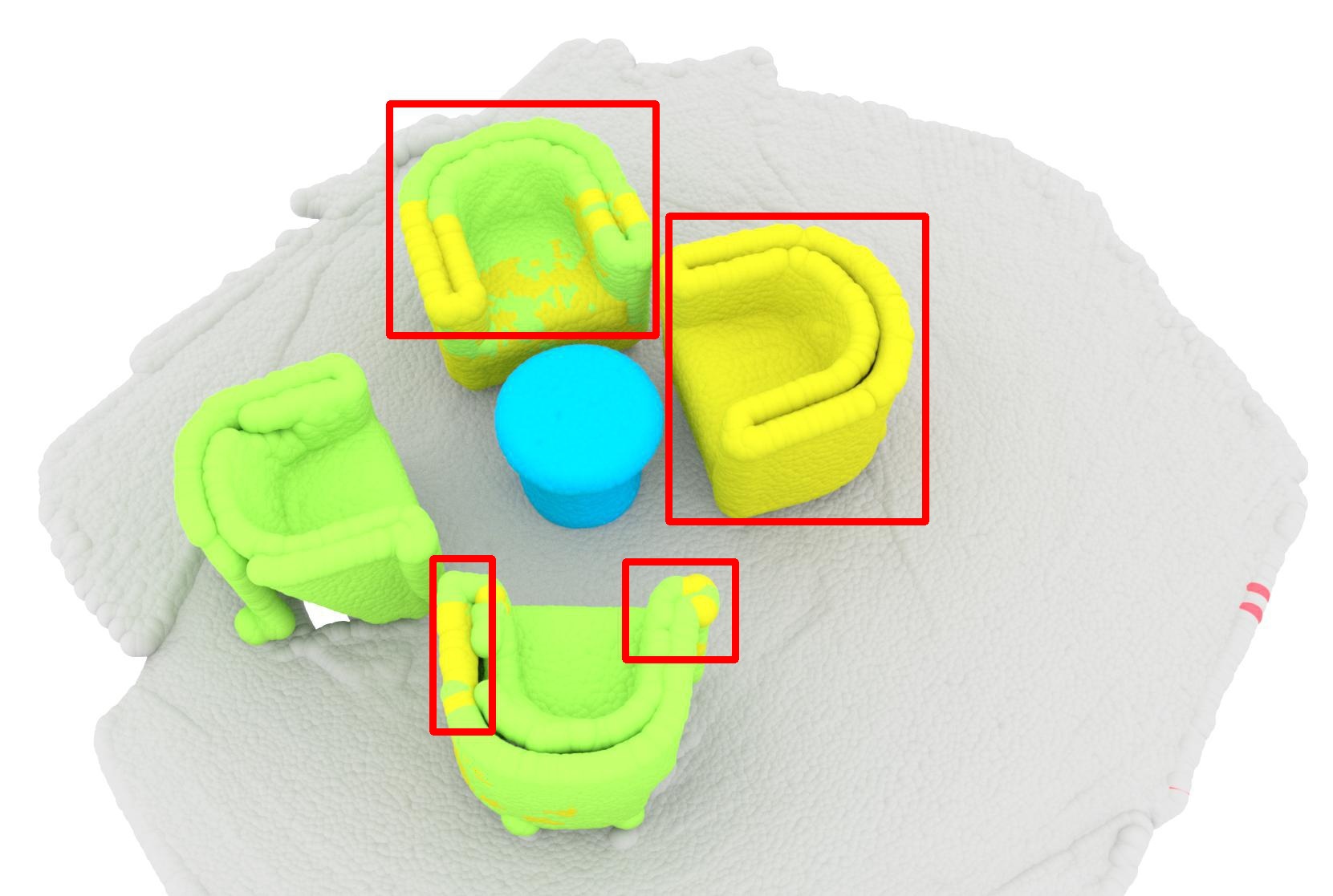}
			\caption*{}
		\end{subfigure}
		\hfill
		\begin{subfigure}[t]{0.24\linewidth}
			\centering
			\includegraphics[width=0.99\linewidth]{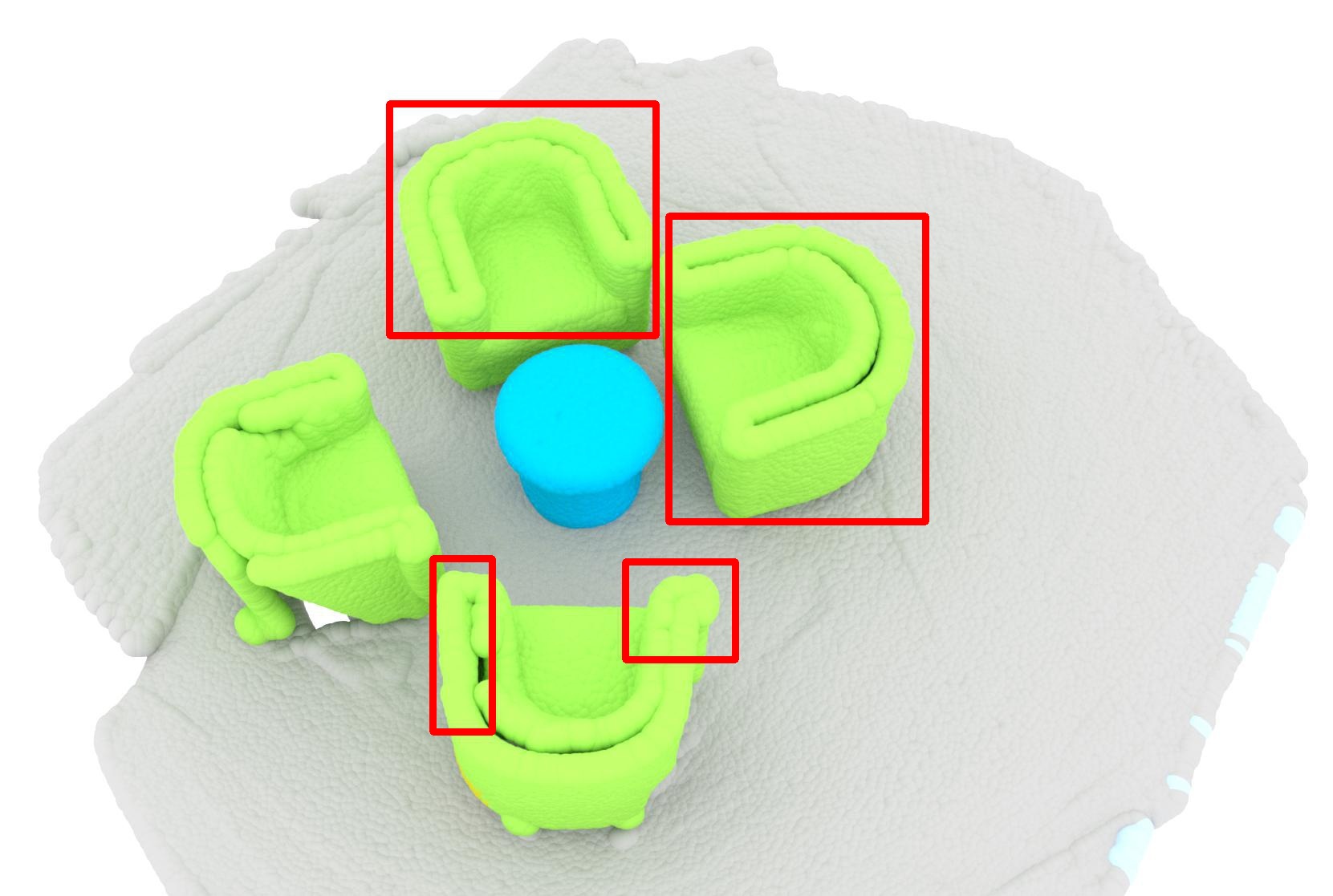}
			\caption*{}
		\end{subfigure}

		\begin{subfigure}[t]{0.24\linewidth}
			\centering
			\includegraphics[width=0.99\linewidth]{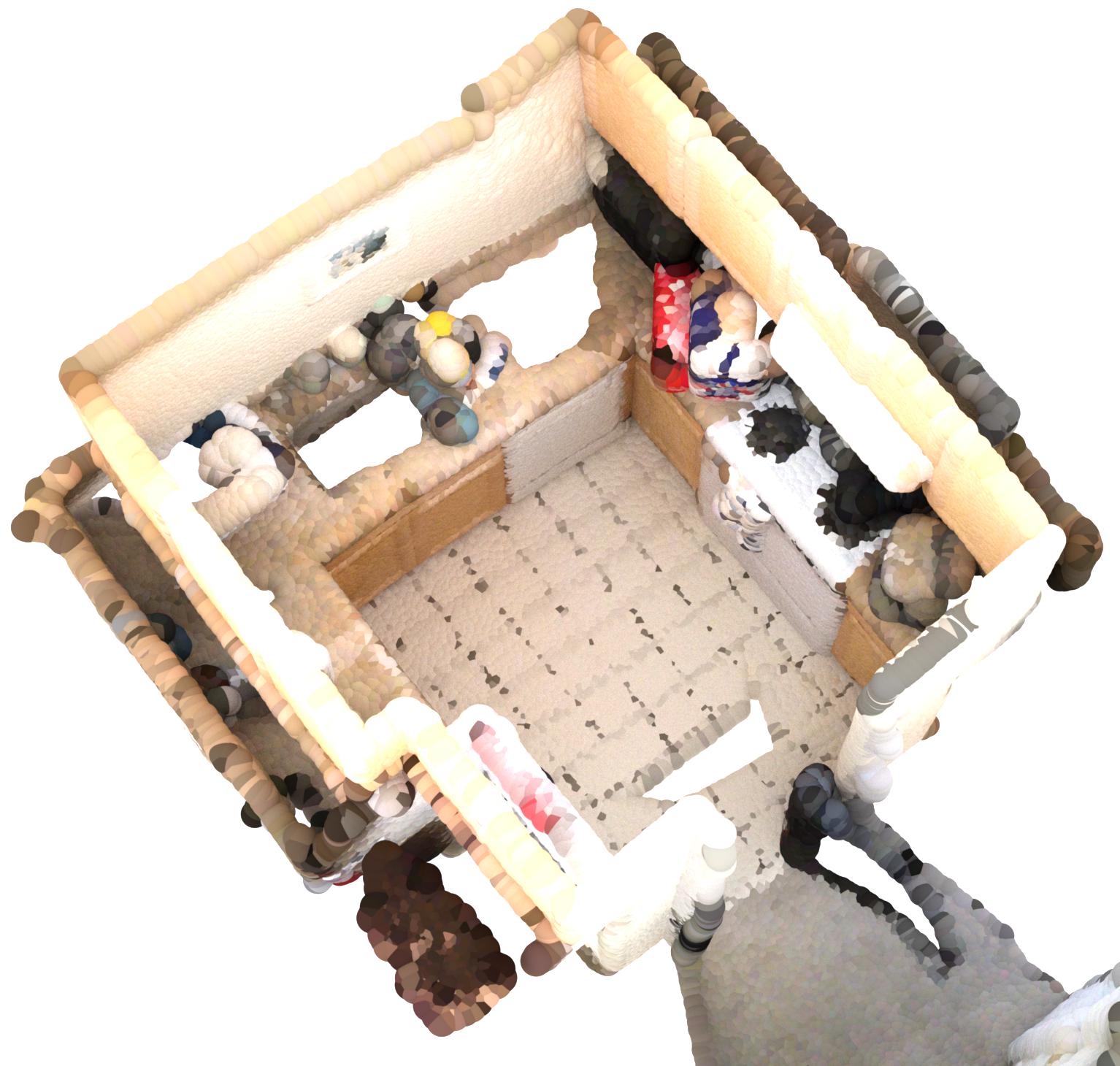}
			\caption*{}
		\end{subfigure}
		\hfill
		\begin{subfigure}[t]{0.24\linewidth}
			\centering
			\includegraphics[width=0.99\linewidth]{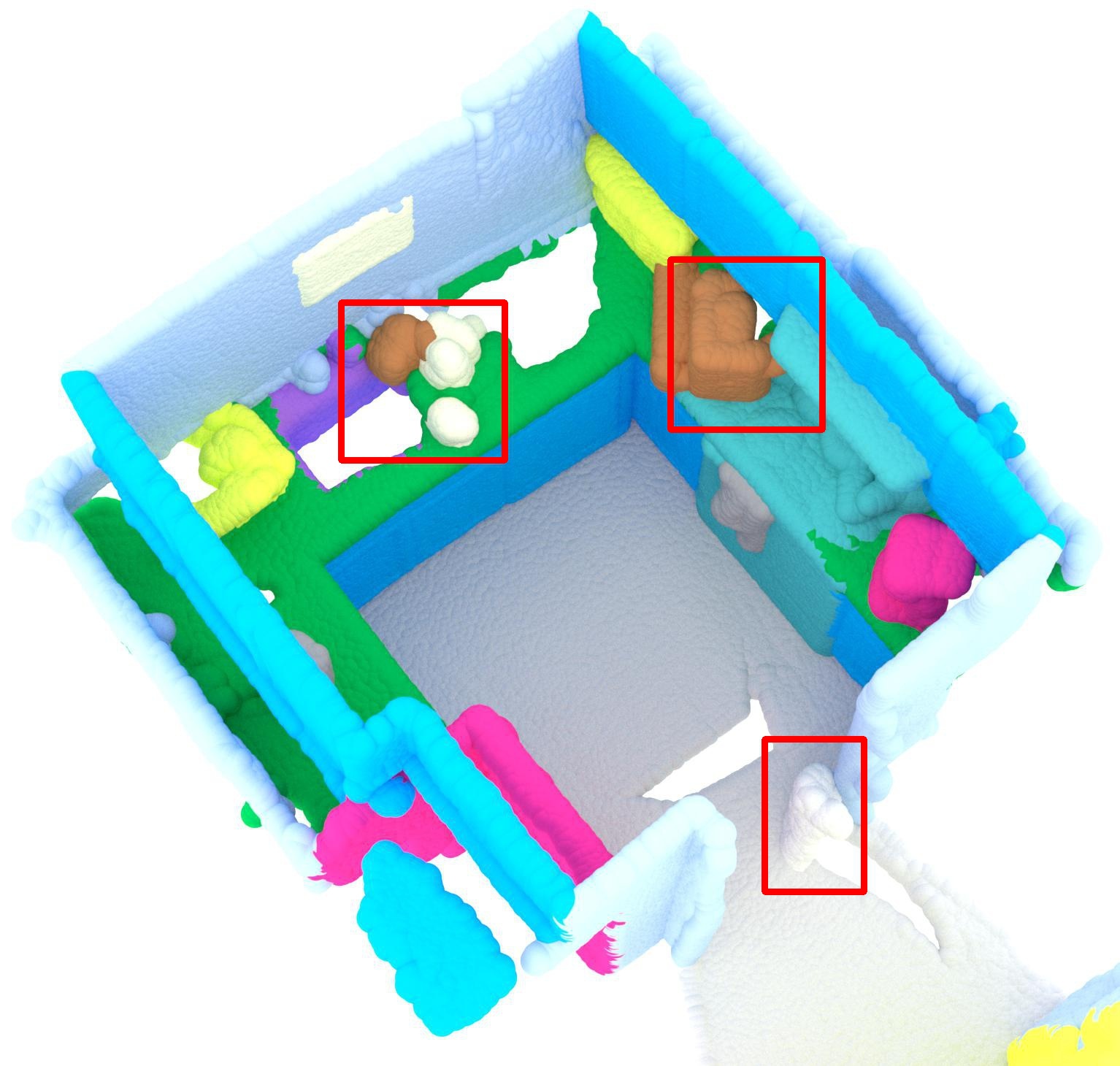}		
			\caption*{}
		\end{subfigure}
		\hfill
		\begin{subfigure}[t]{0.24\linewidth}
			\centering
			\includegraphics[width=0.99\linewidth]{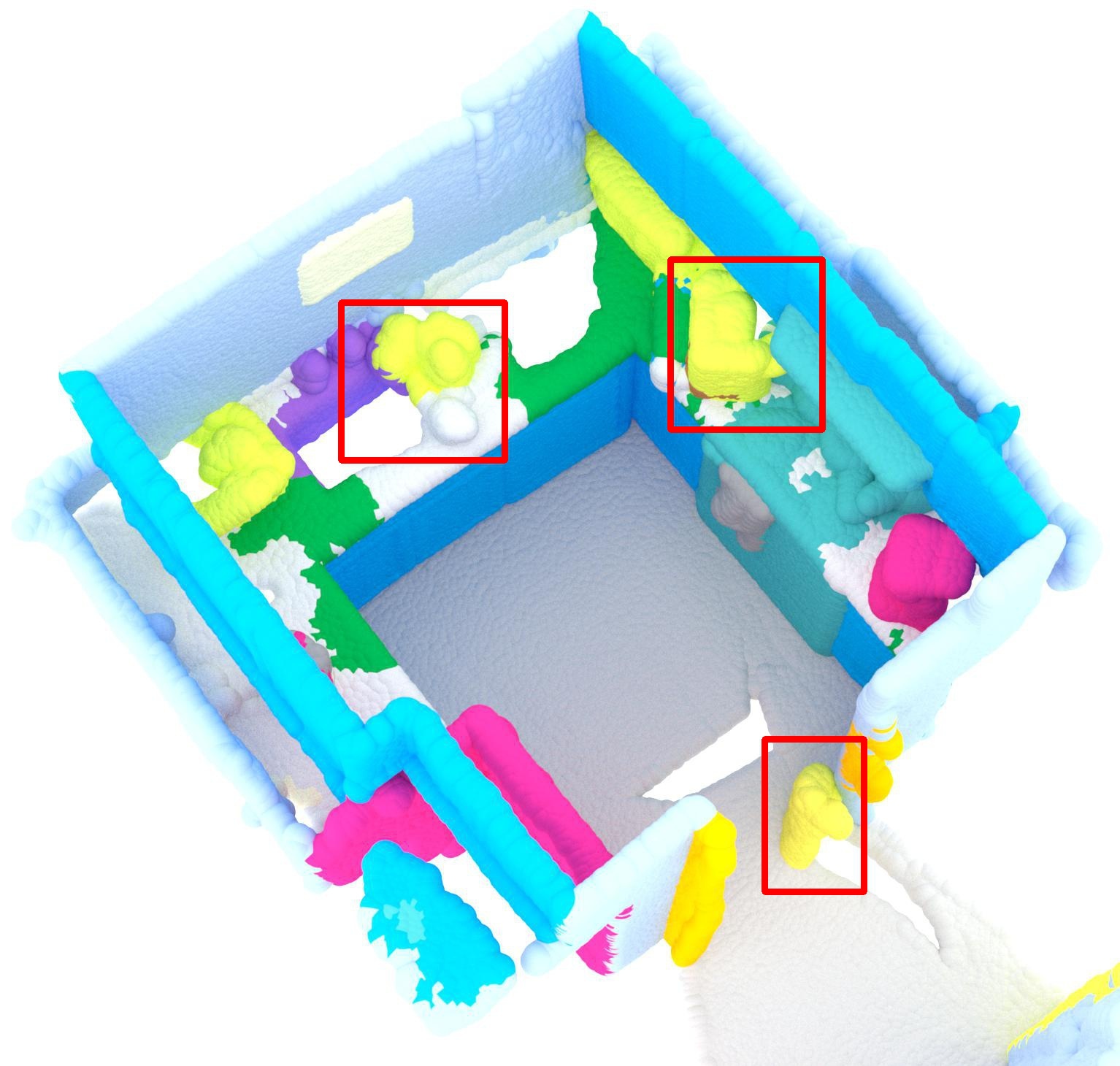}
			\caption*{}
		\end{subfigure}
		\hfill
		\begin{subfigure}[t]{0.24\linewidth}
			\centering
			\includegraphics[width=0.99\linewidth]{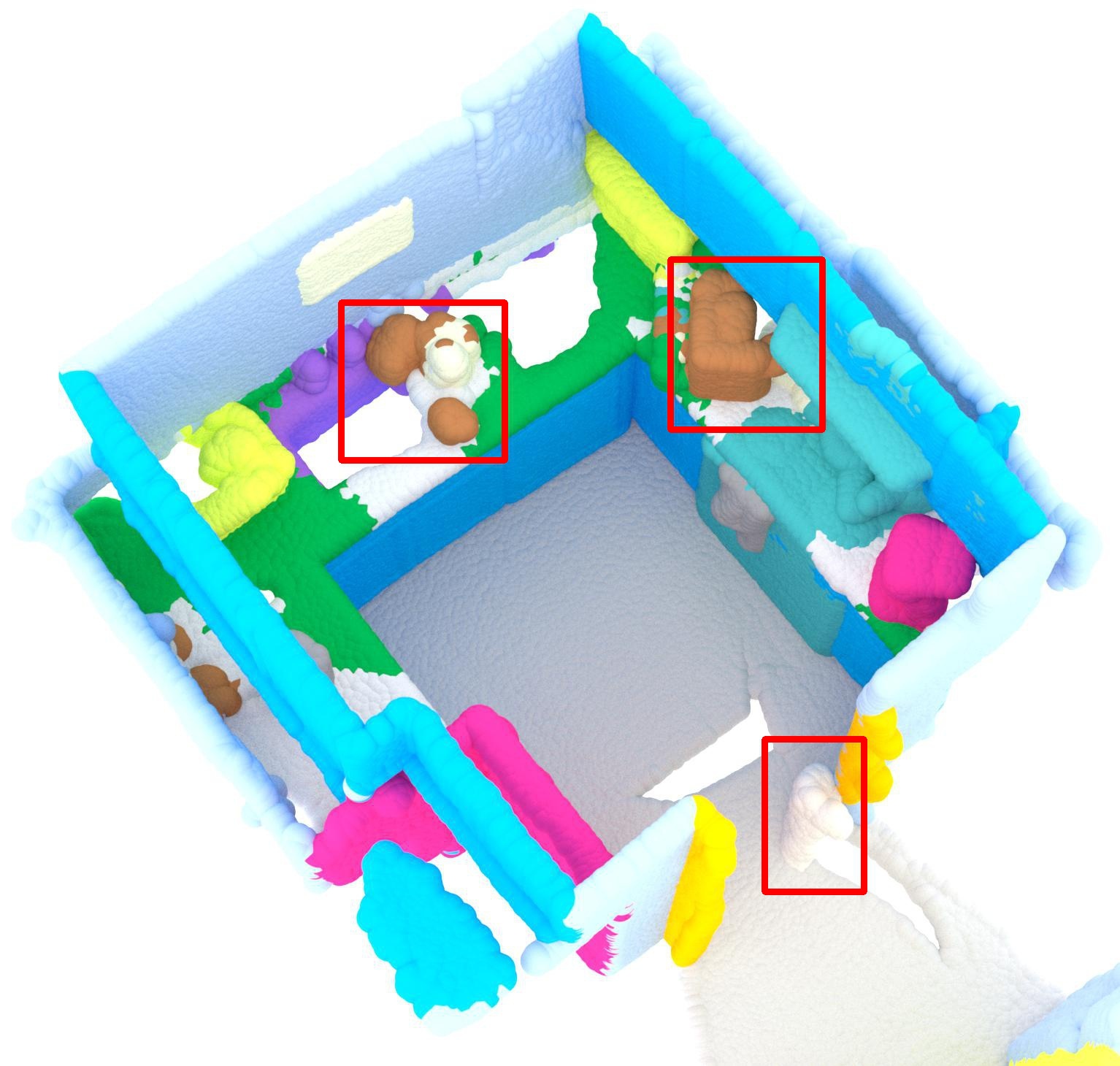}
			\caption*{}
		\end{subfigure}

		\begin{subfigure}[t]{0.24\linewidth}
			\centering
			\includegraphics[width=0.99\linewidth]{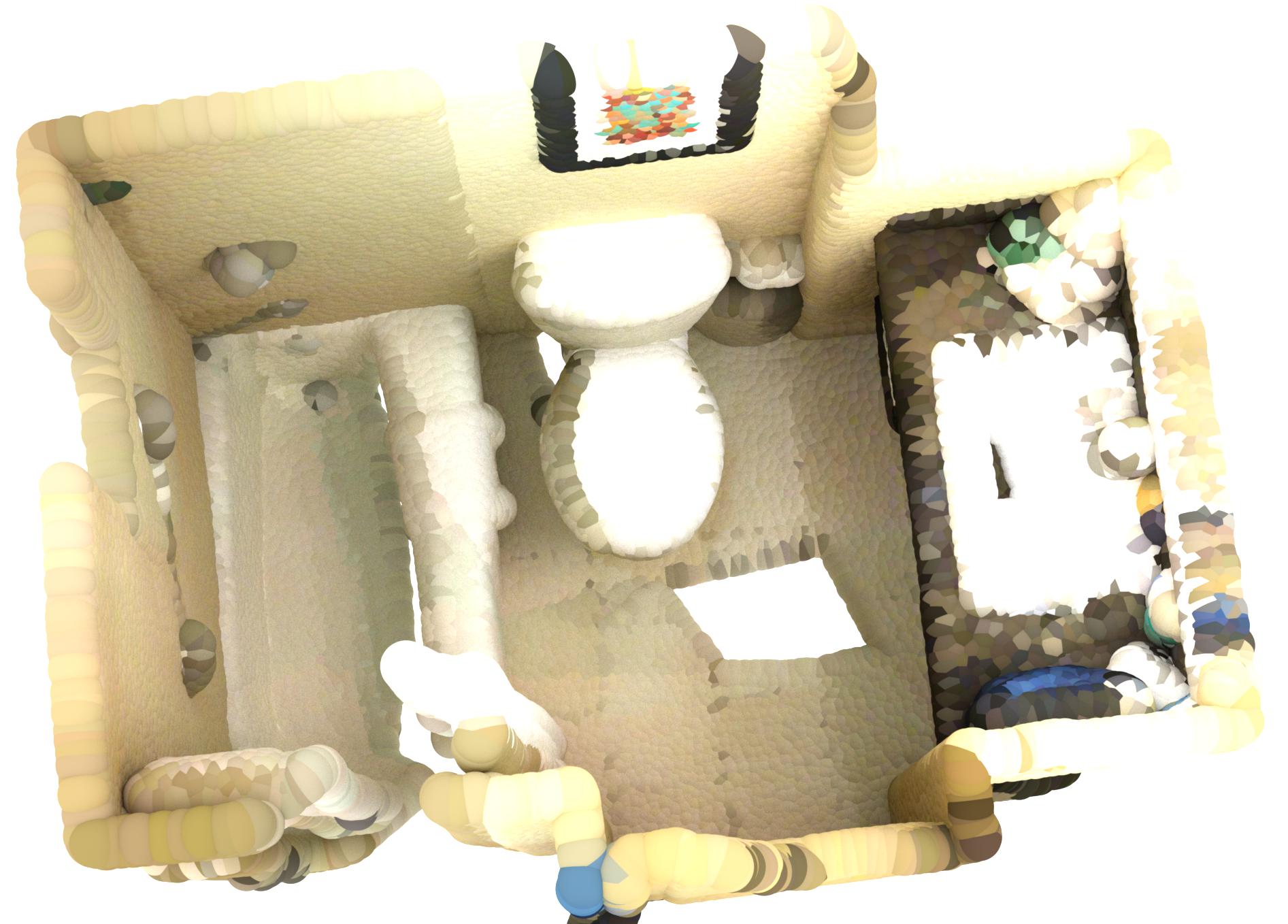}
			\caption*{}
		\end{subfigure}
		\hfill
		\begin{subfigure}[t]{0.24\linewidth}
			\centering
			\includegraphics[width=0.99\linewidth]{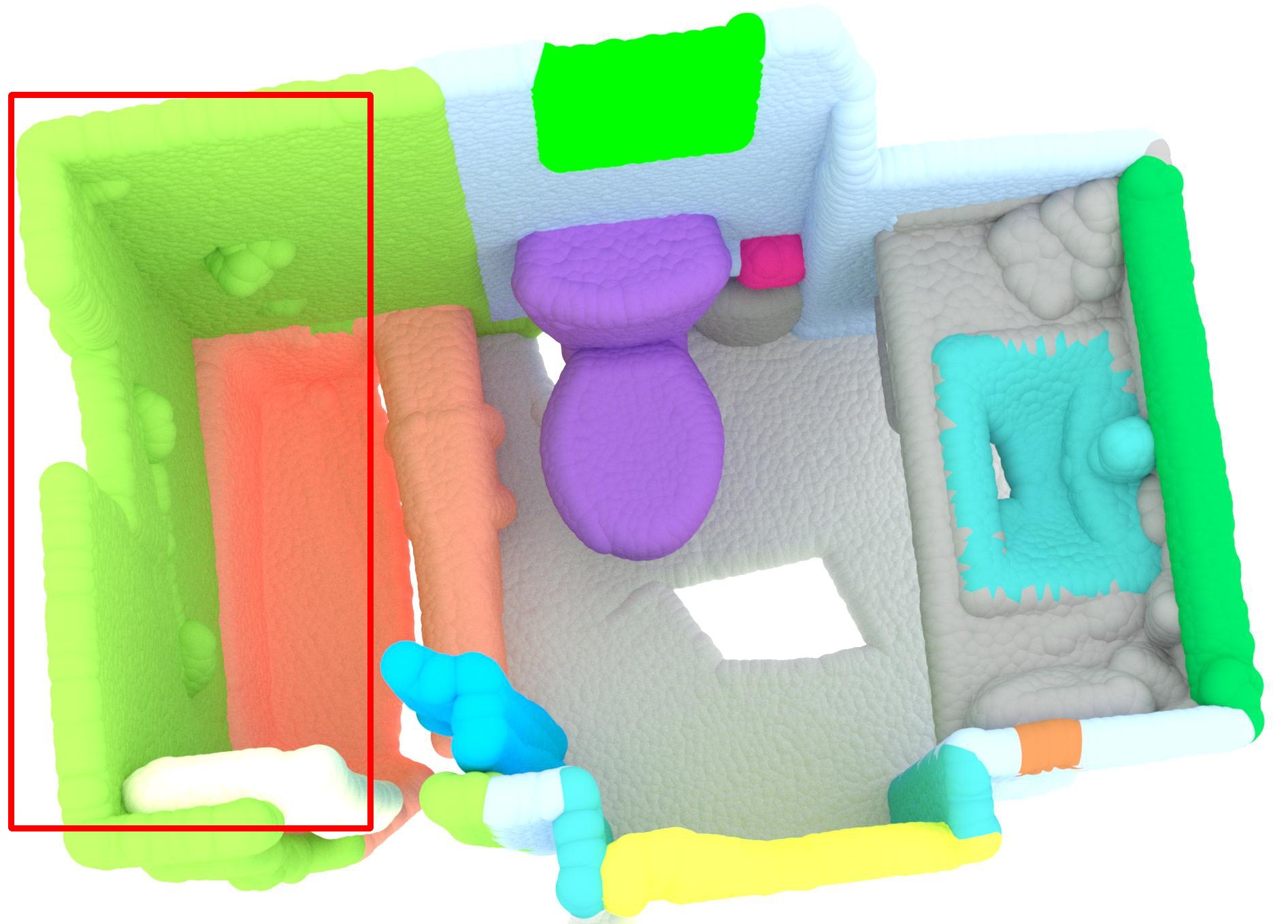}		
			\caption*{}
		\end{subfigure}
		\hfill
		\begin{subfigure}[t]{0.24\linewidth}
			\centering
			\includegraphics[width=0.99\linewidth]{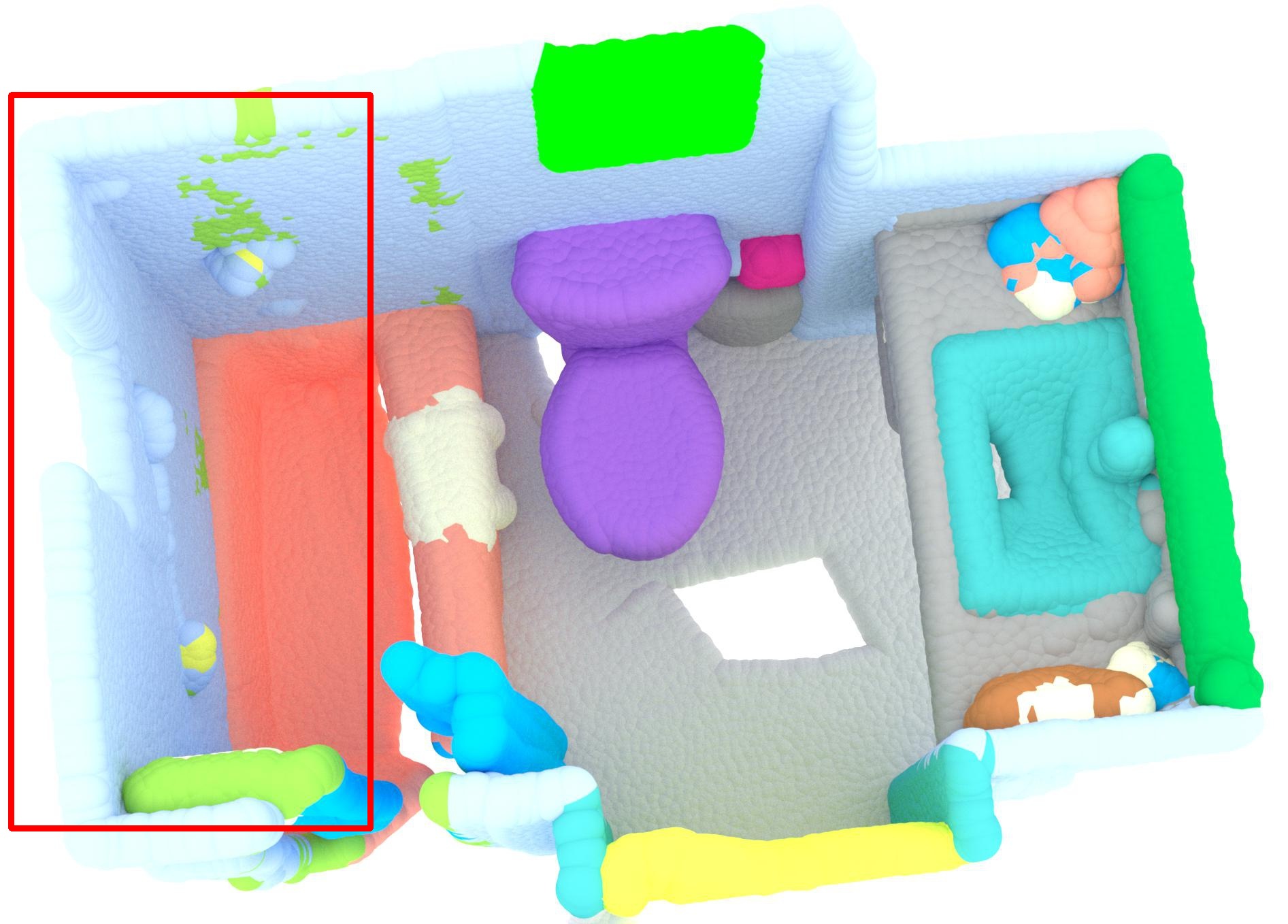}
			\caption*{}
		\end{subfigure}
		\hfill
		\begin{subfigure}[t]{0.24\linewidth}
			\centering
			\includegraphics[width=0.99\linewidth]{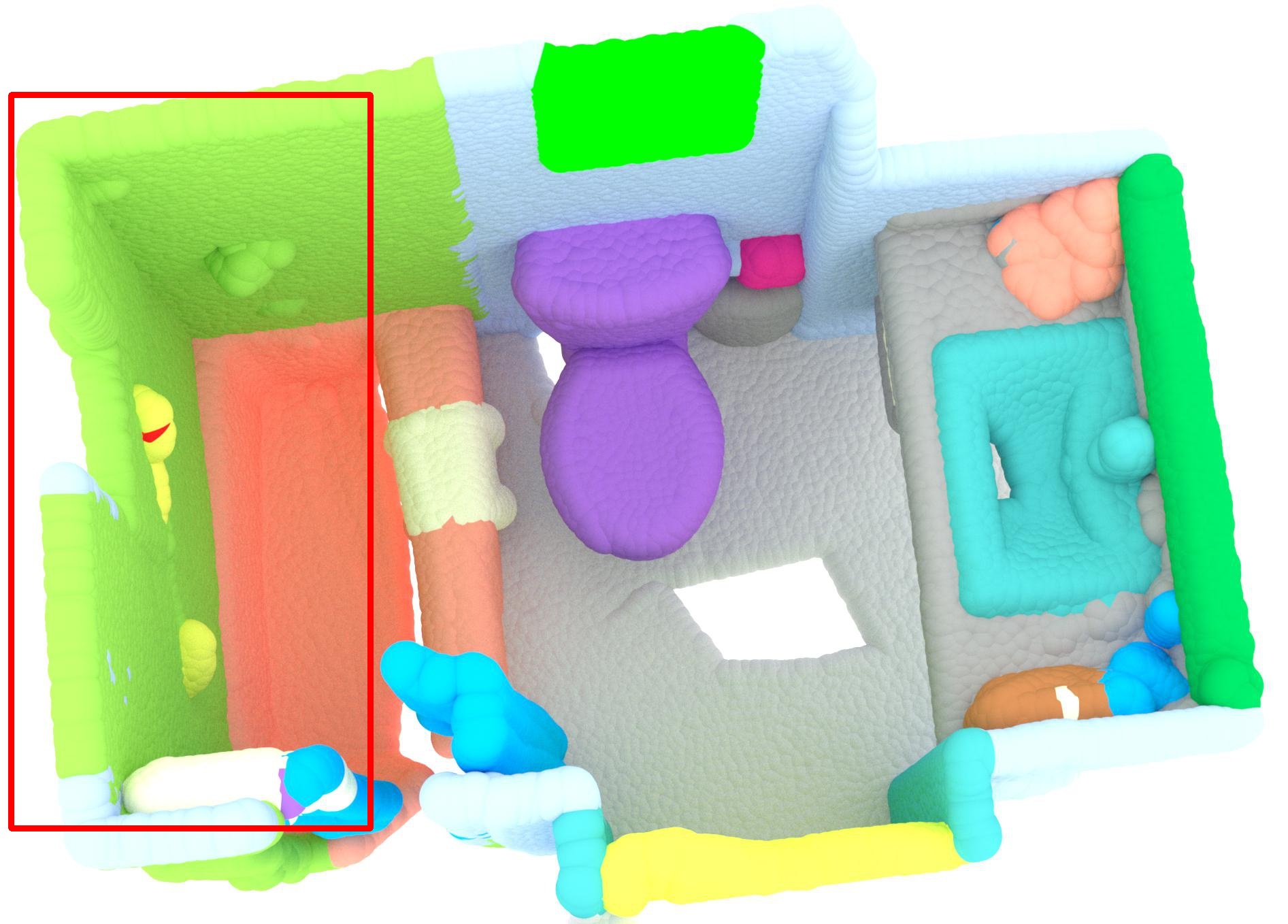}
			\caption*{}
		\end{subfigure}
		
		\begin{subfigure}[t]{0.24\linewidth}
			\centering
			\includegraphics[width=0.99\linewidth]{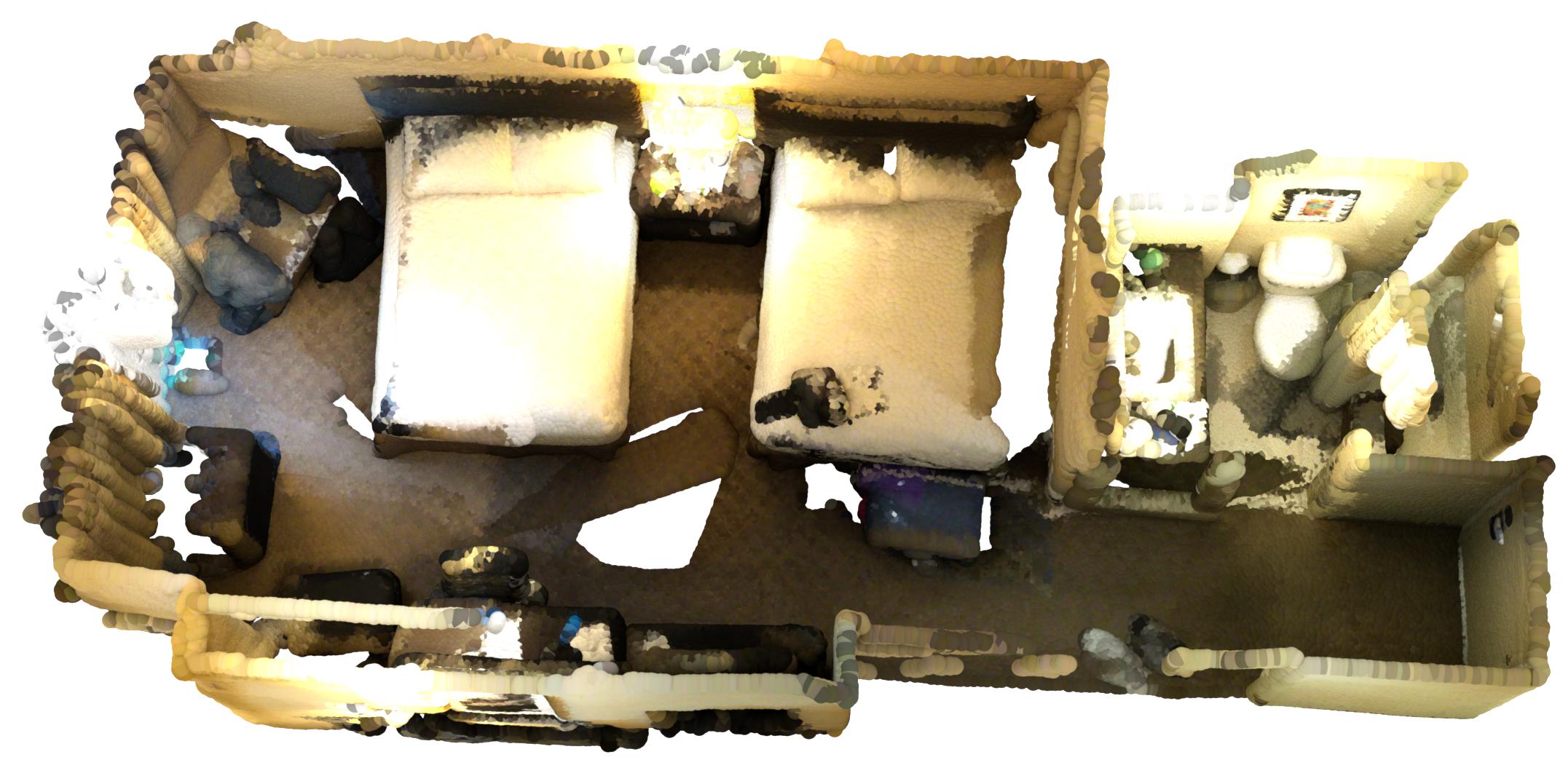}
			\caption{Input Point Cloud}
		\end{subfigure}
		\hfill
		\begin{subfigure}[t]{0.24\linewidth}
			\centering
			\includegraphics[width=0.99\linewidth]{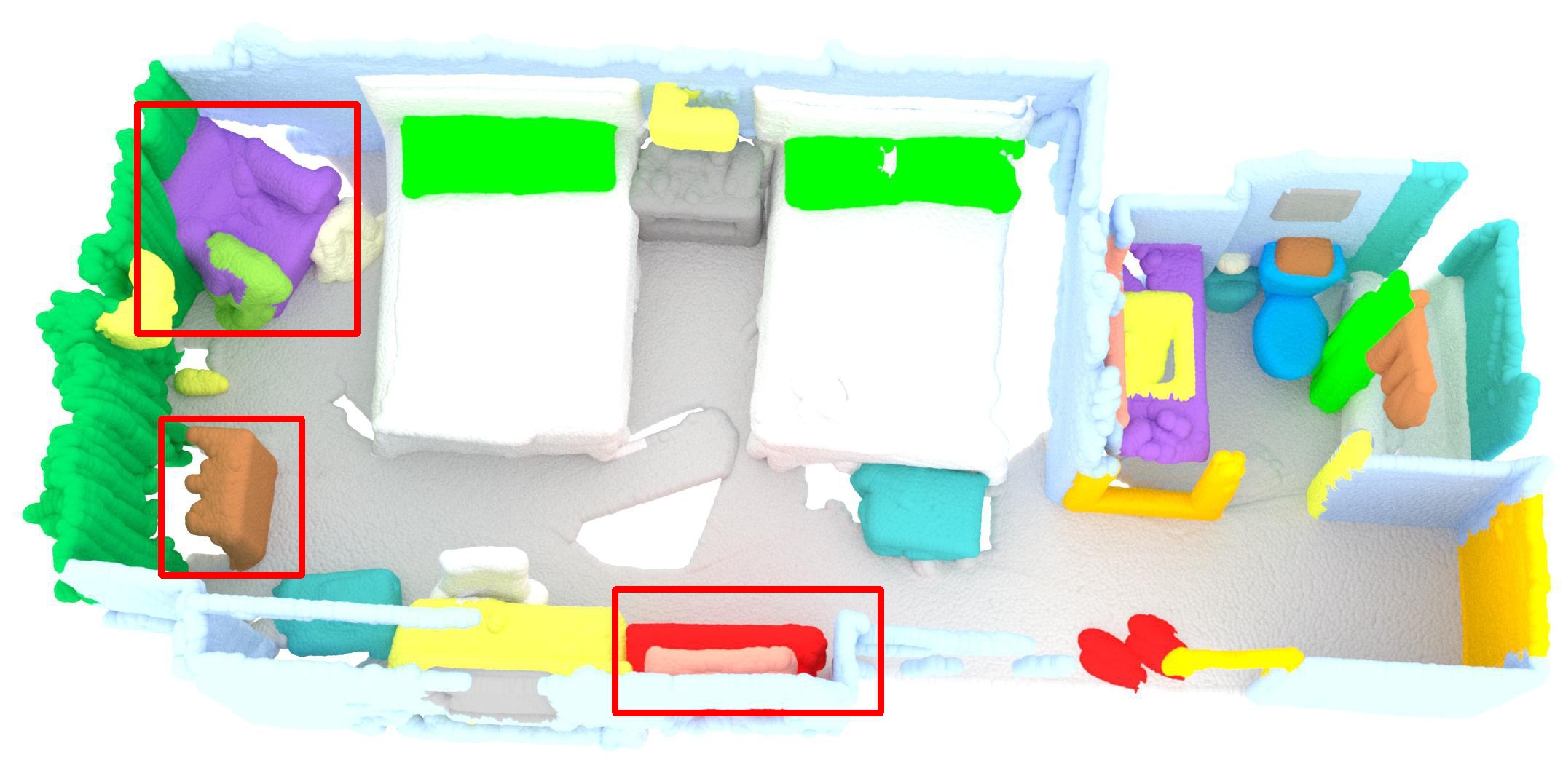}		
			\caption{Groud Truth}
		\end{subfigure}
		\hfill
		\begin{subfigure}[t]{0.24\linewidth}
			\centering
			\includegraphics[width=0.99\linewidth]{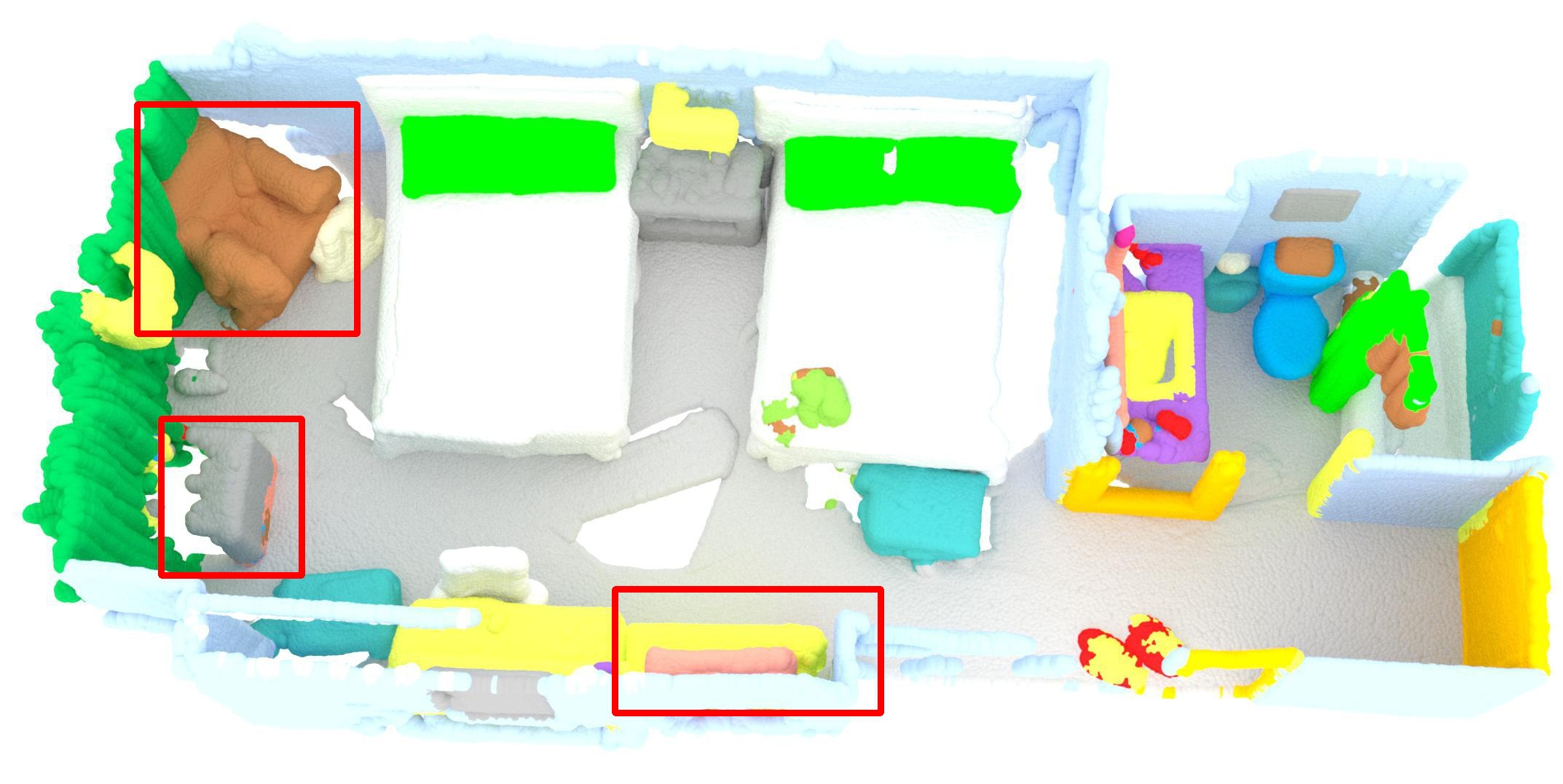}
			\caption{Baseline}
		\end{subfigure}
		\hfill
		\begin{subfigure}[t]{0.24\linewidth}
			\centering
			\includegraphics[width=0.99\linewidth]{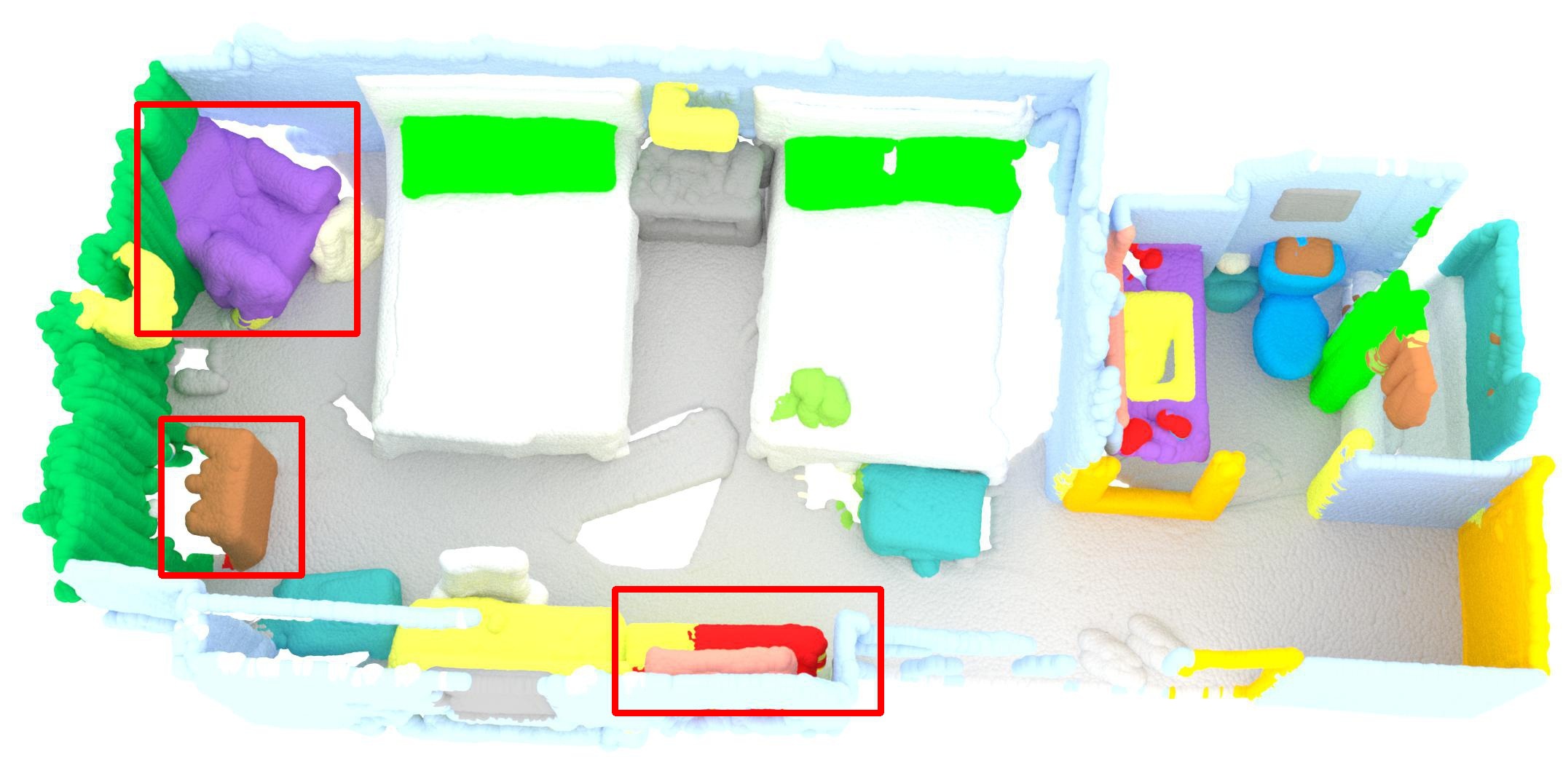}
			\caption{\ours{}}
		\end{subfigure}

		\caption{Visualization comparisons between the CE baseline and \ours{} on ScanNet200.}
		\vspace{-10pt}
		\label{scannet200}
	\end{center}
\end{figure*}

\begin{figure*}[h]
	\centering
	\resizebox{1.0\linewidth}{!}{
		\begin{tabular}{@{\hspace{0.5mm}}c@{\hspace{0.5mm}}c@{\hspace{0.5mm}}c@{\hspace{0.5mm}}c@{\hspace{0.5mm}}c@{\hspace{0.5mm}}c@{\hspace{0.0mm}}}
			\rotatebox[origin=c]{90}{\scriptsize{(a) Input Image}} 
			\includegraphics[width=0.16\linewidth,height=0.8in,valign=m]{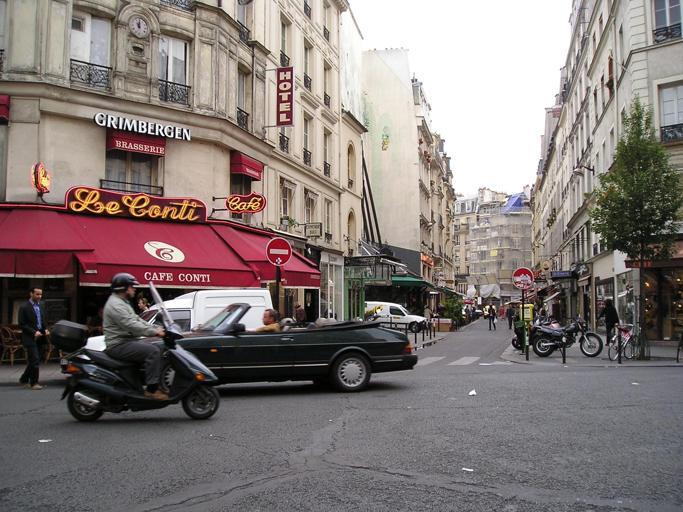}&
			\includegraphics[width=0.16\linewidth,height=0.8in,valign=m]{ade_vis/input_1145.jpg}&
			\includegraphics[width=0.16\linewidth,height=0.8in,valign=m]{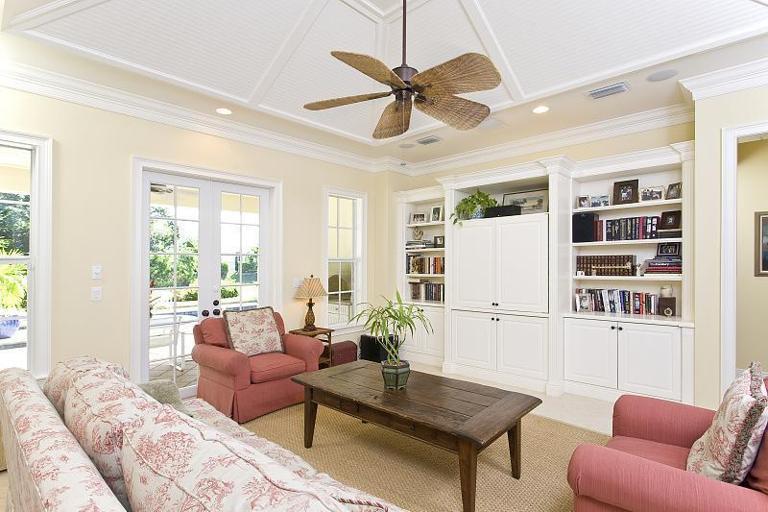}&
			\includegraphics[width=0.16\linewidth,height=0.8in,valign=m]{ade_vis/input_1671.jpg}&
			\includegraphics[width=0.16\linewidth,height=0.8in,valign=m]{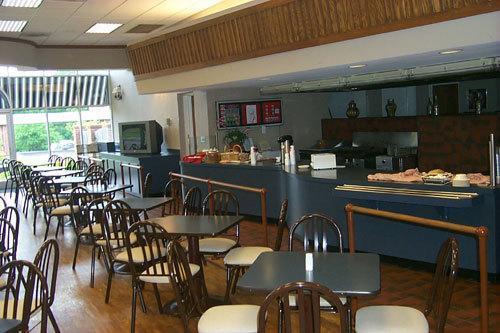}&
			\includegraphics[width=0.16\linewidth,height=0.8in,valign=m]{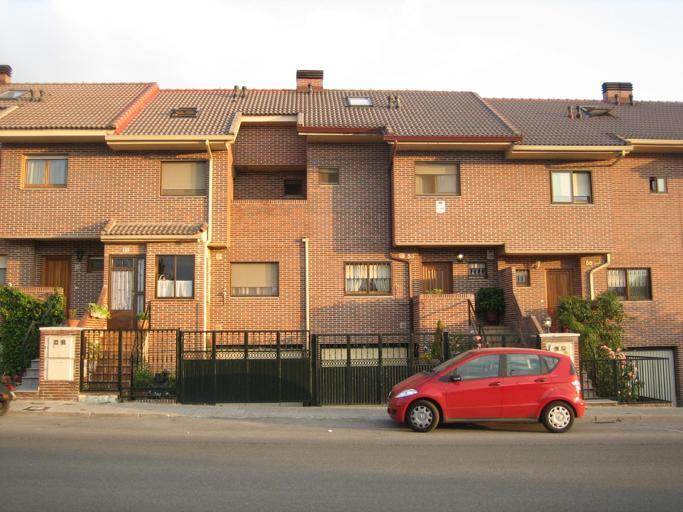} \\
			\addlinespace[3pt]
			\rotatebox[origin=c]{90}{\scriptsize{(b) Ground Truth}}
			\includegraphics[width=0.16\linewidth,height=0.8in,valign=m]{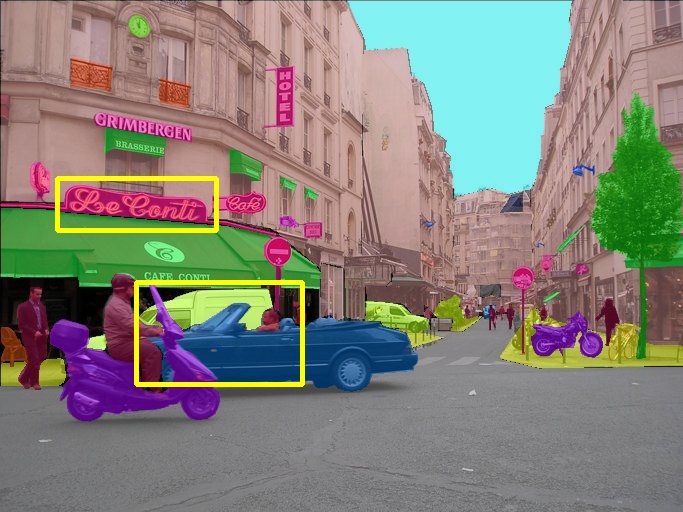}&
			\includegraphics[width=0.16\linewidth,height=0.8in,valign=m]{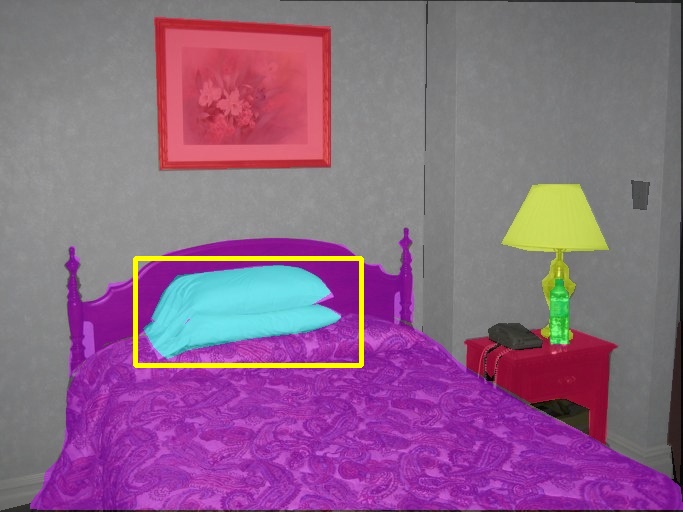}&
			\includegraphics[width=0.16\linewidth,height=0.8in,valign=m]{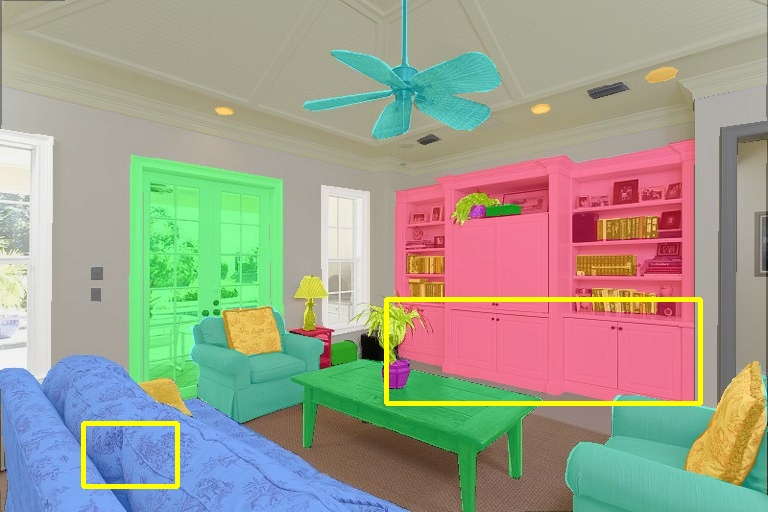}&
			\includegraphics[width=0.16\linewidth,height=0.8in,valign=m]{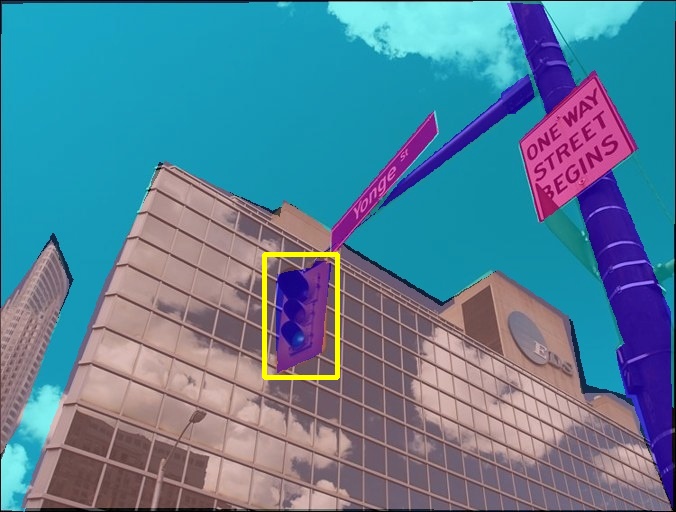}&
			\includegraphics[width=0.16\linewidth,height=0.8in,valign=m]{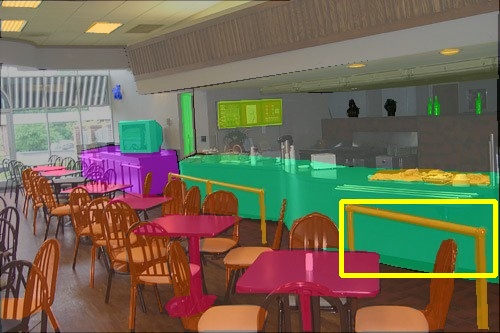}&
			\includegraphics[width=0.16\linewidth,height=0.8in,valign=m]{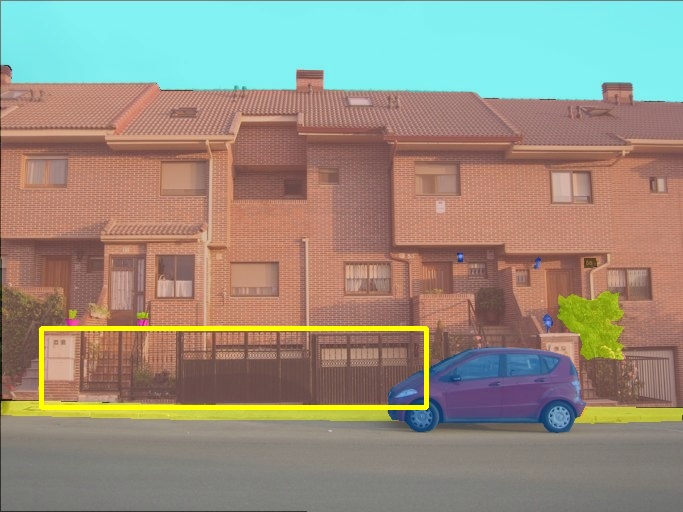} \\
			\addlinespace[3pt]
			\rotatebox[origin=c]{90}{\scriptsize{(c) Baseline}}
			\includegraphics[width=0.16\linewidth,height=0.8in,valign=m]{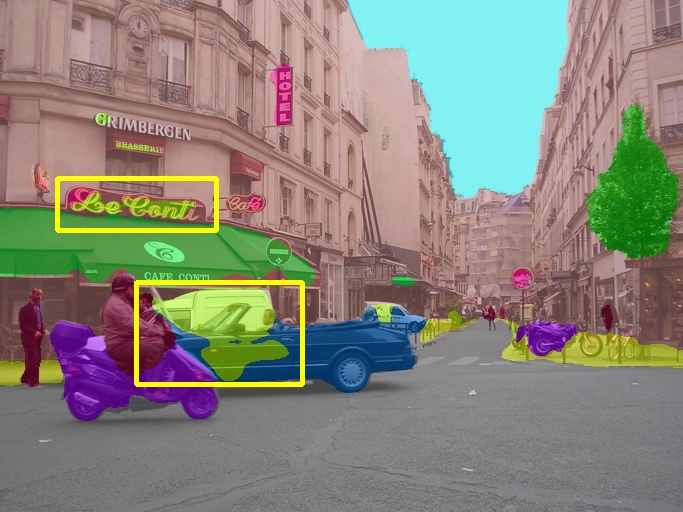}&
			\includegraphics[width=0.16\linewidth,height=0.8in,valign=m]{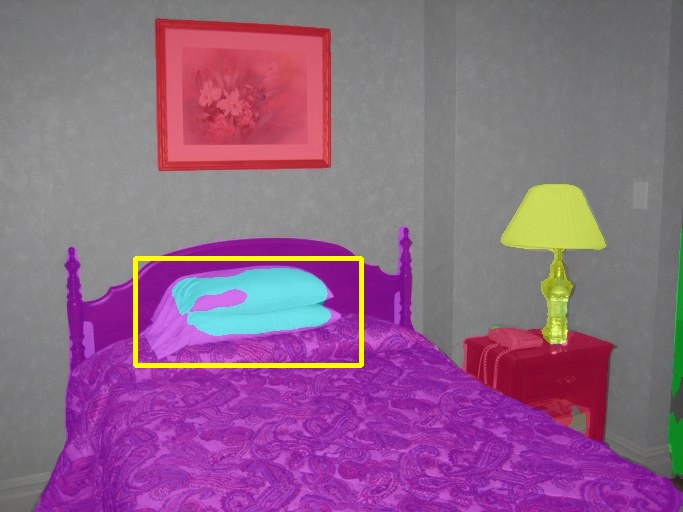}&
			\includegraphics[width=0.16\linewidth,height=0.8in,valign=m]{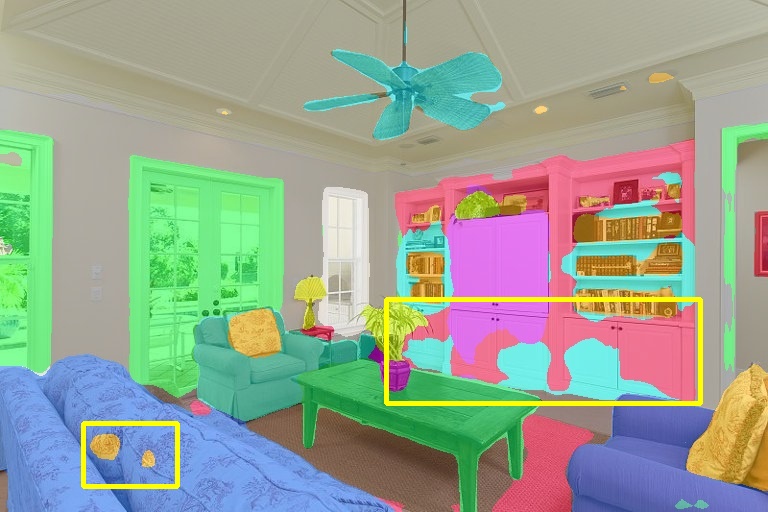}&
			\includegraphics[width=0.16\linewidth,height=0.8in,valign=m]{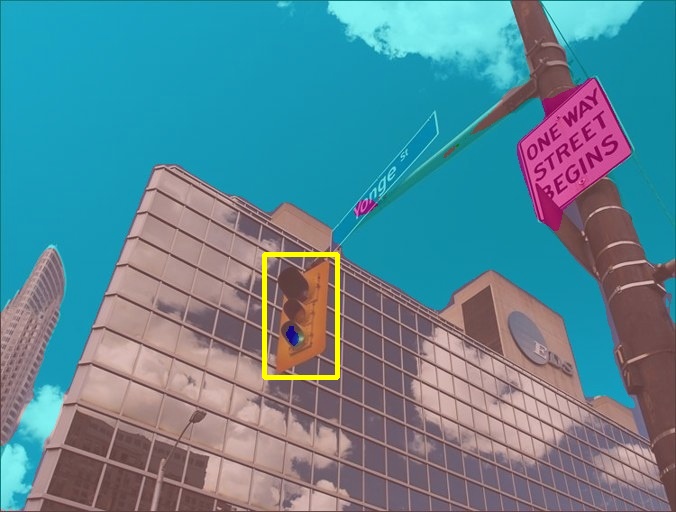}&
			\includegraphics[width=0.16\linewidth,height=0.8in,valign=m]{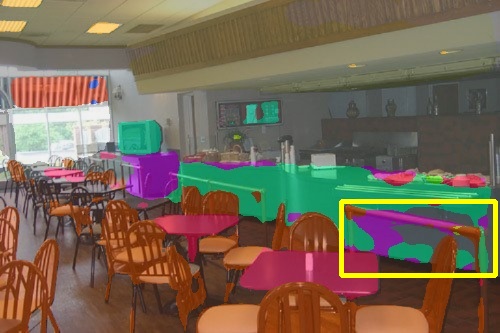}&
			\includegraphics[width=0.16\linewidth,height=0.8in,valign=m]{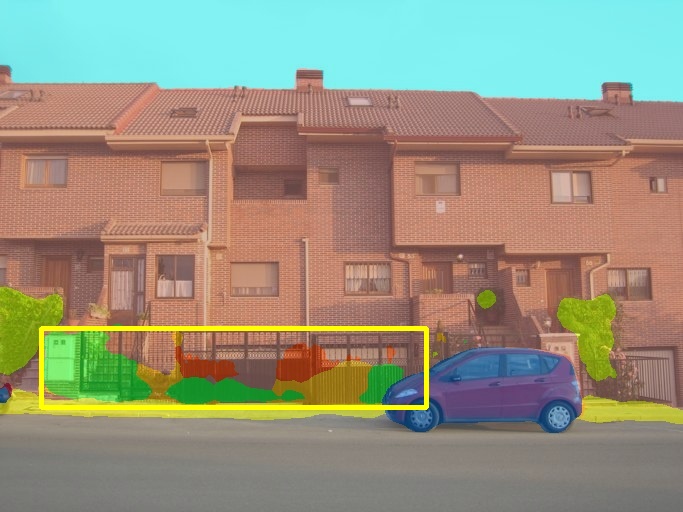} \\
			\addlinespace[3pt]
			\rotatebox[origin=c]{90}{\scriptsize{(d) \ours{}}}
			\includegraphics[width=0.16\linewidth,height=0.8in,valign=m]{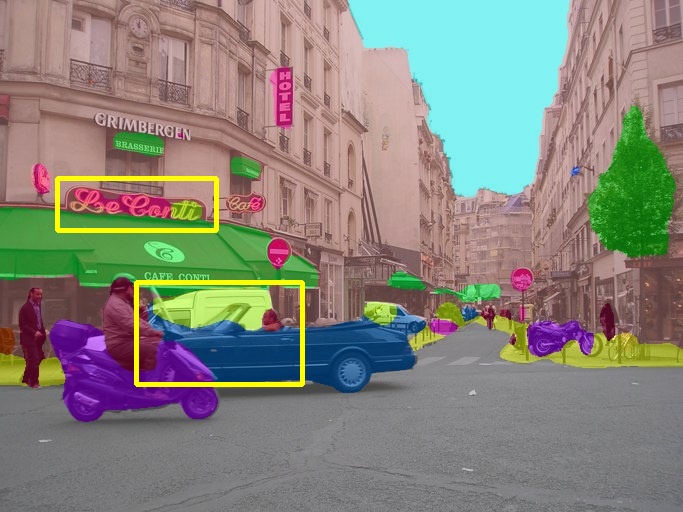}&
			\includegraphics[width=0.16\linewidth,height=0.8in,valign=m]{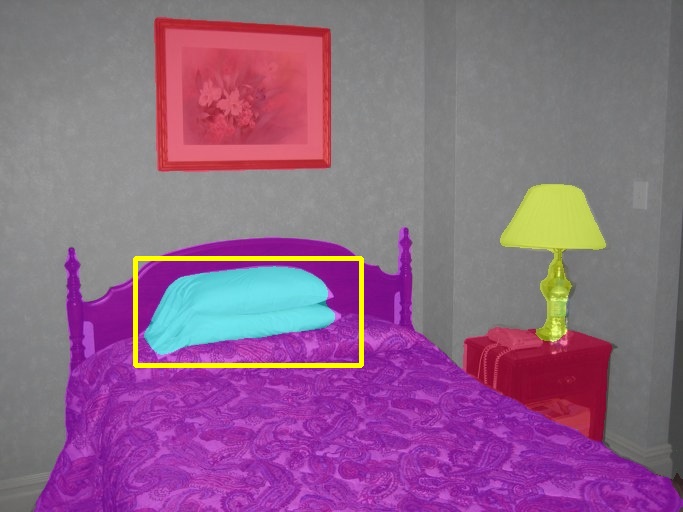}&
			\includegraphics[width=0.16\linewidth,height=0.8in,valign=m]{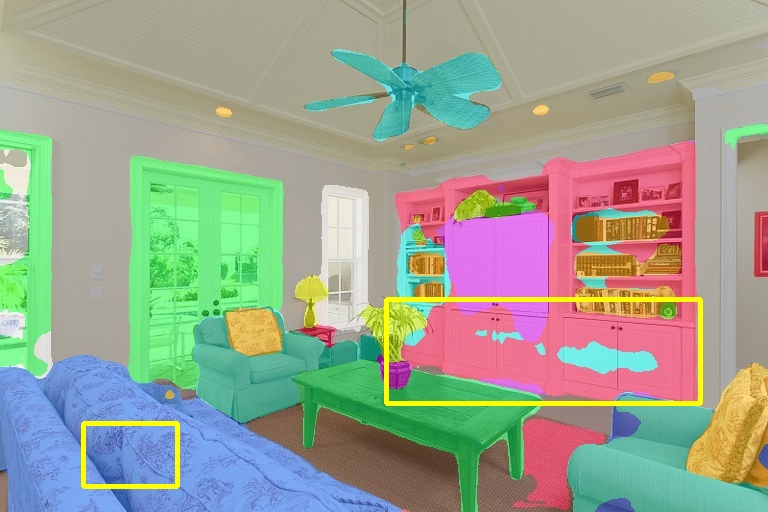}&
			\includegraphics[width=0.16\linewidth,height=0.8in,valign=m]{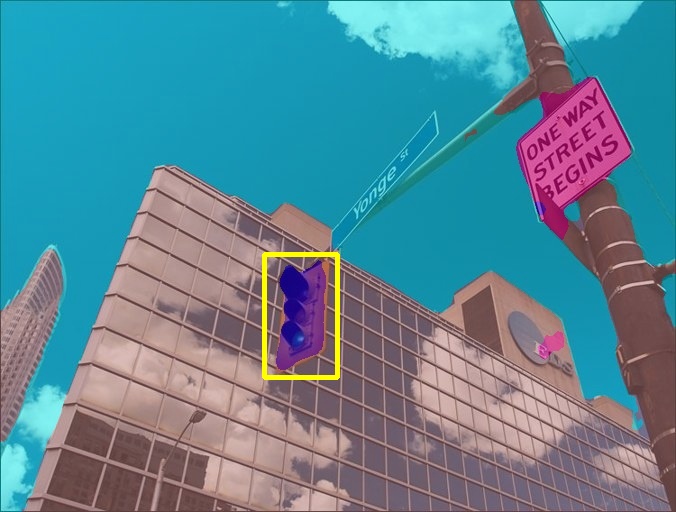}&
			\includegraphics[width=0.16\linewidth,height=0.8in,valign=m]{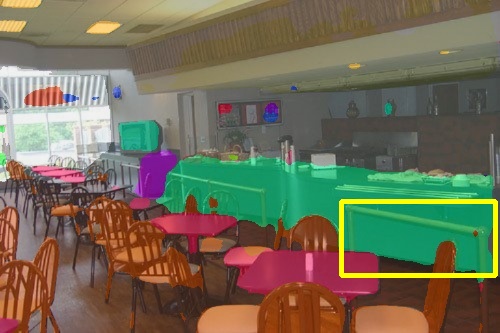}&
			\includegraphics[width=0.16\linewidth,height=0.8in,valign=m]{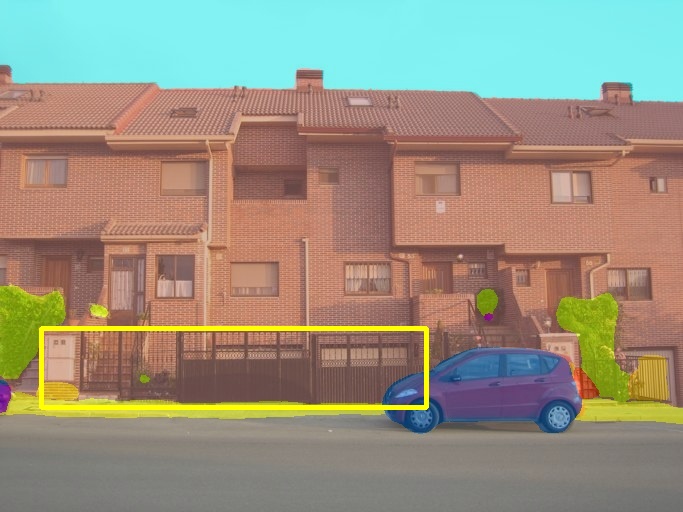} \\
	\end{tabular}}
	\caption{Visualization comparisons between the CE baseline and \ours{} on ADE20K.}
	\label{fig:visual_comparison_ade20k_v1}
\end{figure*}

\begin{figure*}[h]
	\centering
	\resizebox{1.0\linewidth}{!}{
		\begin{tabular}{@{\hspace{0.5mm}}c@{\hspace{0.5mm}}c@{\hspace{0.5mm}}c@{\hspace{0.5mm}}c@{\hspace{0.5mm}}c@{\hspace{0.5mm}}c@{\hspace{0.5mm}}c@{\hspace{0.5mm}}c@{\hspace{0.0mm}}}
			\rotatebox[origin=c]{90}{\scriptsize{(a) Input Image}} 
			\includegraphics[width=0.16\linewidth,height=0.8in,valign=m]{coco_vis/input_570756.jpg}&
			\includegraphics[width=0.16\linewidth,height=0.8in,valign=m]{coco_vis/input_446522.jpg}&
			\includegraphics[width=0.16\linewidth,height=0.8in,valign=m]{coco_vis/input_522713.jpg}&
			\includegraphics[width=0.16\linewidth,height=0.8in,valign=m]{coco_vis/input_562581.jpg}&
			\includegraphics[width=0.16\linewidth,height=0.8in,valign=m]{coco_vis/input_568290.jpg}&
			\includegraphics[width=0.16\linewidth,height=0.8in,valign=m]{coco_vis/input_569972.jpg} \\
			\addlinespace[3pt]
			\rotatebox[origin=c]{90}{\scriptsize{(b) Ground Truth}}
			\includegraphics[width=0.16\linewidth,height=0.8in,valign=m]{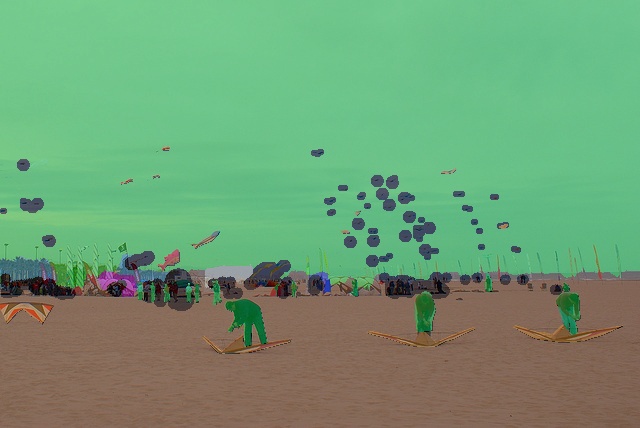}&
			\includegraphics[width=0.16\linewidth,height=0.8in,valign=m]{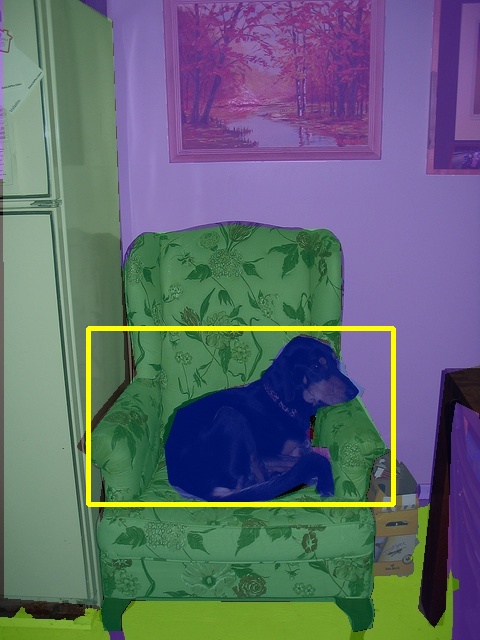}&
			\includegraphics[width=0.16\linewidth,height=0.8in,valign=m]{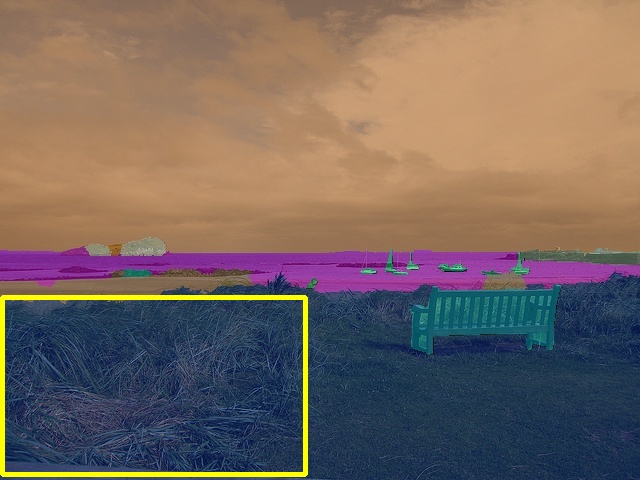}&
			\includegraphics[width=0.16\linewidth,height=0.8in,valign=m]{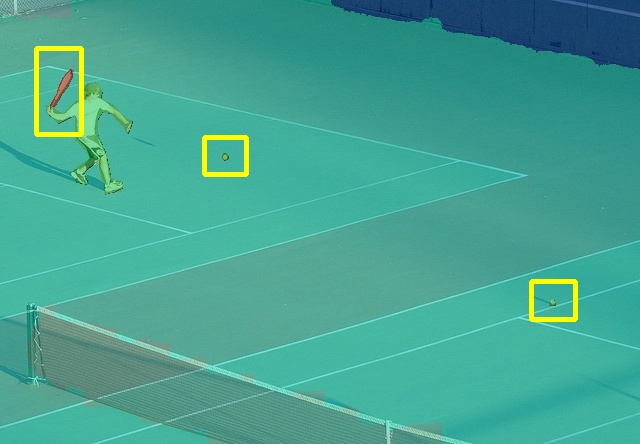}&
			\includegraphics[width=0.16\linewidth,height=0.8in,valign=m]{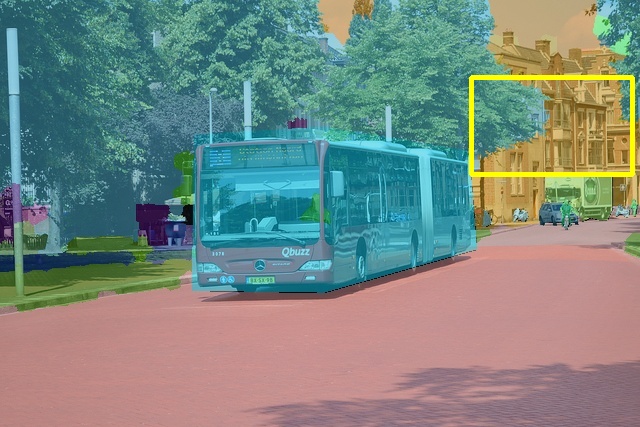}&
			\includegraphics[width=0.16\linewidth,height=0.8in,valign=m]{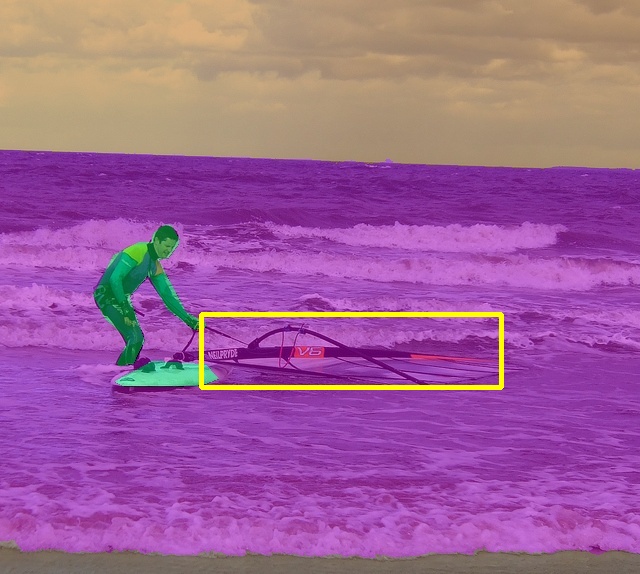} \\
			\addlinespace[3pt]
			\rotatebox[origin=c]{90}{\scriptsize{(c) Baseline}}
			\includegraphics[width=0.16\linewidth,height=0.8in,valign=m]{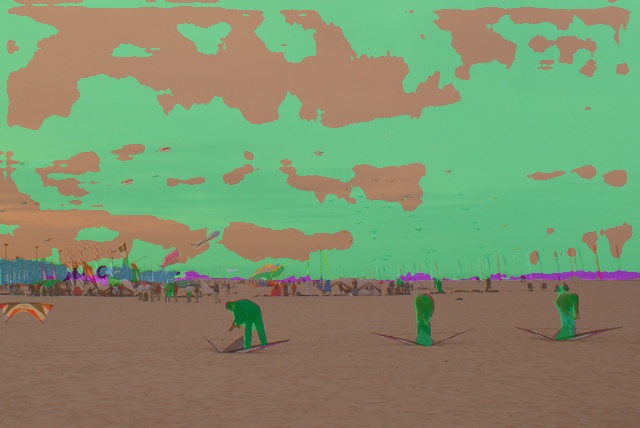}&
			\includegraphics[width=0.16\linewidth,height=0.8in,valign=m]{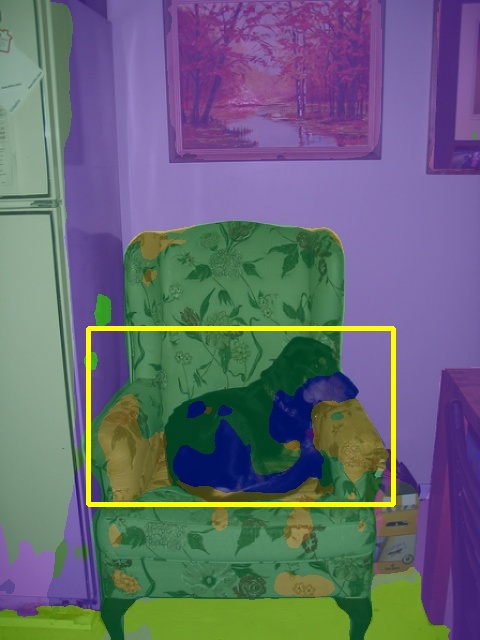}&
			\includegraphics[width=0.16\linewidth,height=0.8in,valign=m]{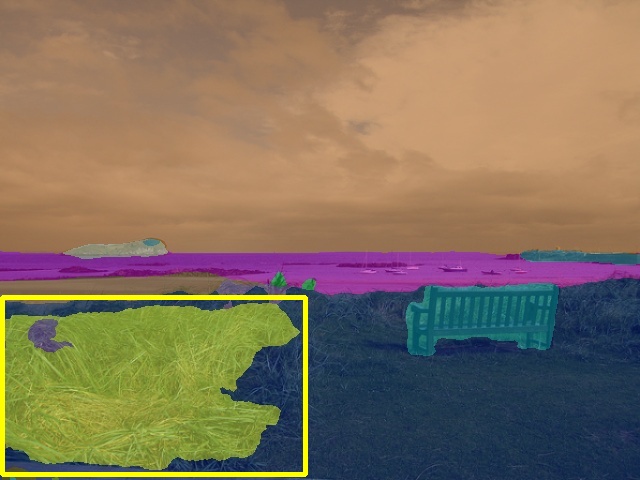}&
			\includegraphics[width=0.16\linewidth,height=0.8in,valign=m]{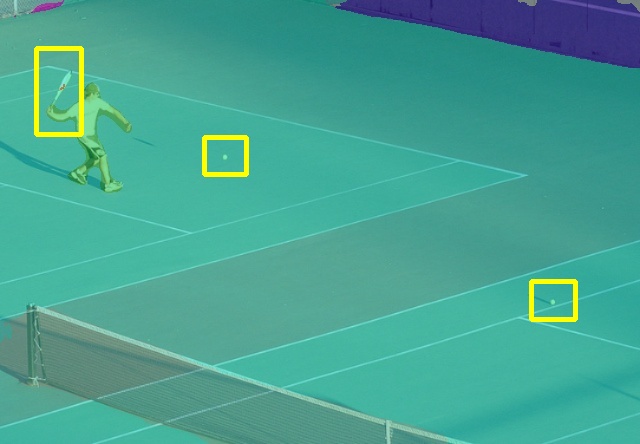}&
			\includegraphics[width=0.16\linewidth,height=0.8in,valign=m]{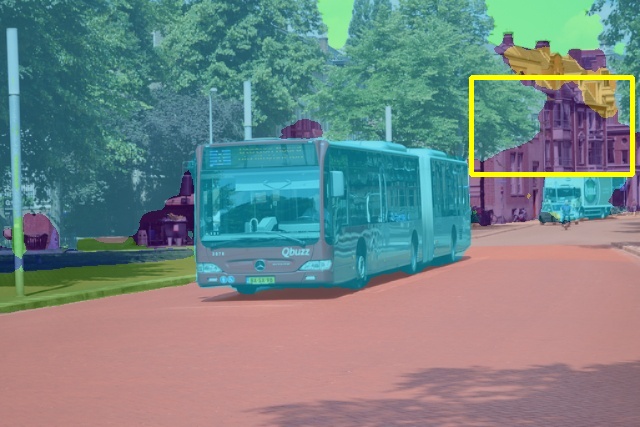}&
			\includegraphics[width=0.16\linewidth,height=0.8in,valign=m]{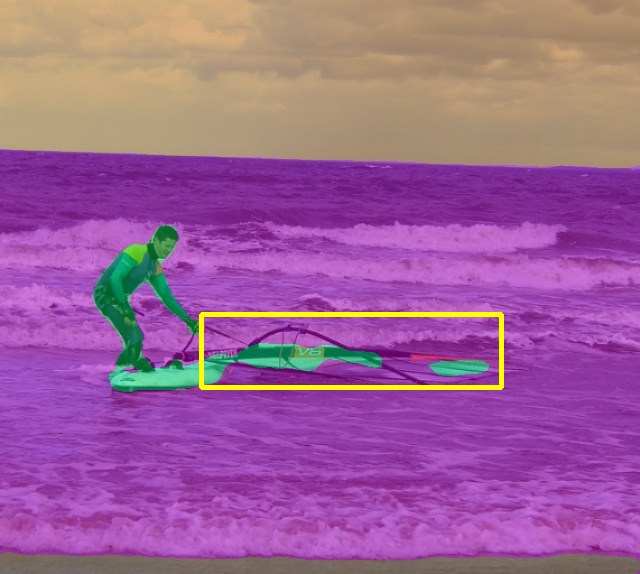} \\
			\addlinespace[3pt]
			\rotatebox[origin=c]{90}{\scriptsize{(d) \ours{}}}
			\includegraphics[width=0.16\linewidth,height=0.8in,valign=m]{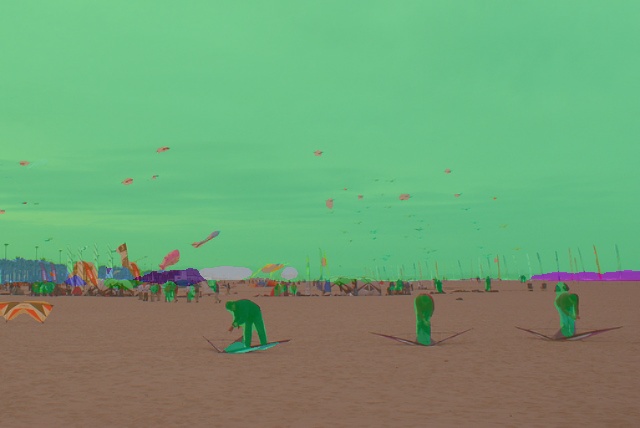}&
			\includegraphics[width=0.16\linewidth,height=0.8in,valign=m]{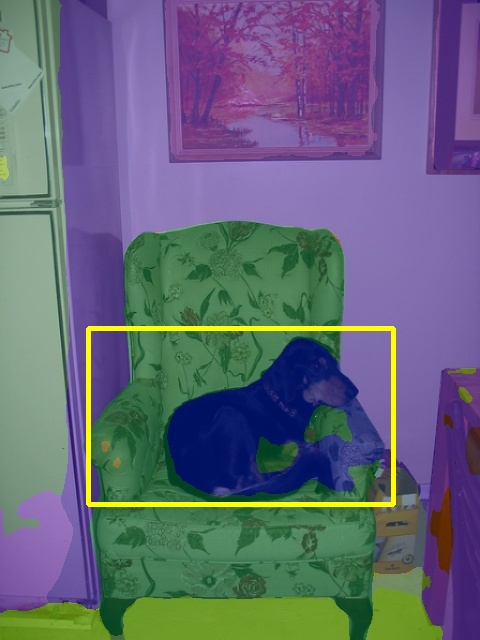}&
			\includegraphics[width=0.16\linewidth,height=0.8in,valign=m]{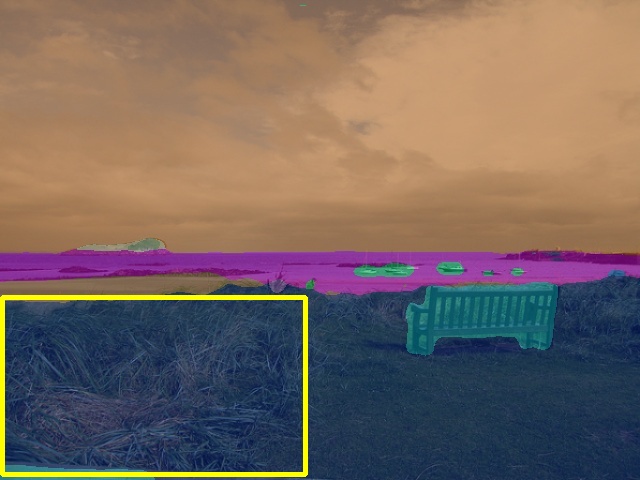}&
			\includegraphics[width=0.16\linewidth,height=0.8in,valign=m]{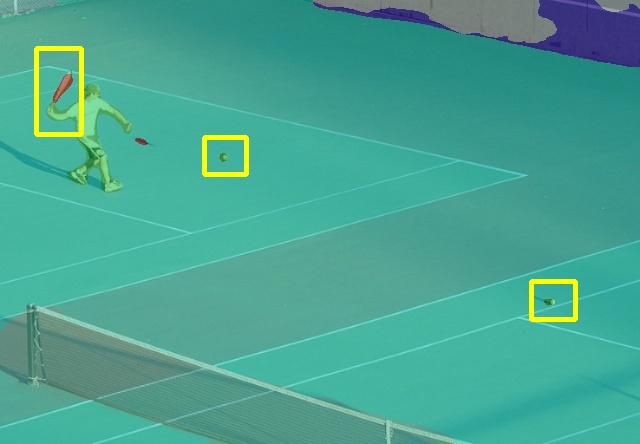}&
			\includegraphics[width=0.16\linewidth,height=0.8in,valign=m]{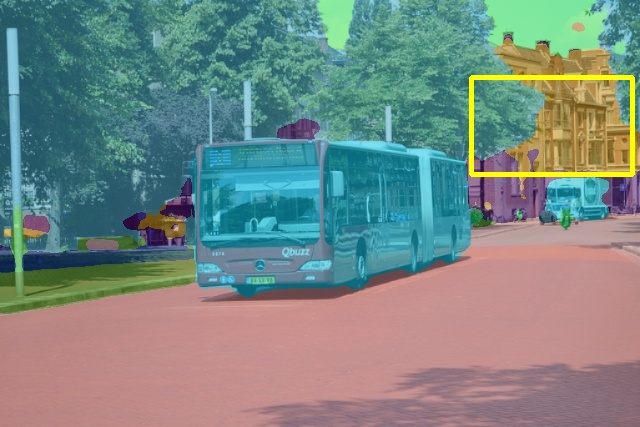}&
			\includegraphics[width=0.16\linewidth,height=0.8in,valign=m]{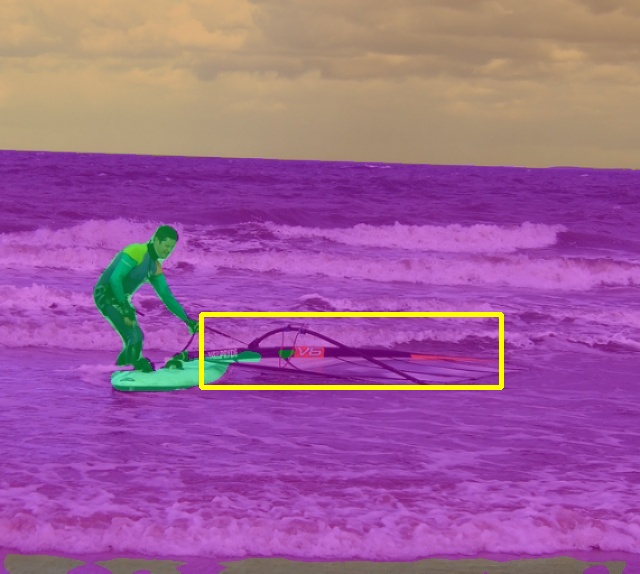} \\
	\end{tabular}}
	\caption{Visualization comparisons between the CE baseline and \ours{} on COCO-Stuff164K.}
	\label{fig:visual_comparison_cocostuff_v1}
\end{figure*}

\clearpage
\section{Limitation Analysis}\label{sec_limit}

\subsection{Simple Imbalance Cases} 
This work presents a new regularization for imbalanced semantic segmentation. However, when the class number of the dataset is small enough or the imbalanced severity is negligible, the performance of \ours{} is quite incremental and even degraded. Here we use Cityscapes and ScanNet v2, two datasets as examples. ScanNet v2 just includes 20 classes, and Cityscapes only contains 30 classes. 
Following the above analysis, we compute the imbalanced factors on both datasets. As shown in Table~\ref{if_comp}, the center imbalanced ratios for these two datasets are much smaller than those on ScanNet200, ADE20K, and COCO-Stuff164K. Under simple imbalance cases, the improvement of \ours{} is quite limited. The detailed performance results are shown in Table~\ref{city} and Table~\ref{scannet}.
\begin{table*}[h]
	\begin{center}
		\begin{minipage}[t]{0.6\linewidth}
			\centering
			\resizebox{0.9\linewidth}{!}
			{
				\begin{tabular}{y{60}y{60}y{60}y{60}}
					\toprule
					\textbf{Method}  &Backbone  &{mIoU} (s.s.) &{mIoU} (m.s.) \\
					\midrule
					DLV3P       &ResNet-18  &76.3 &77.9                  \\
					+ \ours{}           &ResNet-18  &77.5\cgaphl{+}{1.2} &79.2\cgaphl{+}{1.3} \\
					%UPerNet &            &ResNet-50  &42.0 &42.8 \\
					%UPerNet & + \ours{}          &ResNet-50  &43.3 &44.0\cgaphl{+}{1.2} \\
					DLV3P       &ResNet-50  &79.8 &81.4                 \\
					+ \ours{}     &ResNet-50  &80.3\cgaphl{+}{0.5} &81.5\cgaphl{+}{0.1} \\
					\bottomrule
				\end{tabular}
			}
			\caption{Performance on Cityscapes.}
			\label{city}
		\end{minipage}
		\hspace{5pt}
		\begin{minipage}[t]{0.35\linewidth}
			\centering
			\resizebox{0.95\linewidth}{!}{
				\begin{tabular}{y{90}y{60}}
					\toprule
					\textbf{Method} & {mIoU} \\
					\midrule
					PointTransformer~\cite{zhao2020point}       &70.6                  \\
					SparseConvNet~\cite{graham20183d}  &69.3 \\ 
					MinkowskiNet~\cite{choy20194d}       &73.0                 \\
					+ \ours{}   &73.7\cgaphl{+}{0.7} \\
					\bottomrule
			\end{tabular}}
			\caption{Performance on ScanNet v2.}
			\label{scannet}
		\end{minipage}\\
		\vspace{10pt}
		\begin{minipage}[t]{0.6\linewidth}
			\centering
			\resizebox{0.88\linewidth}{!}{
				\begin{tabular}{y{90}y{75}y{75}}
					\toprule
					\textbf{Method}  &Training   &Inference \\
					\midrule
					DLV3P (ResNet-101)      &0.725s  &0.294s       \\
					+ \ours{}           &0.858s\cgaphll{+}{18.3\%}  &0.294s \\
					\midrule
					MinkowskiNet-34      &5.844s  &4.856s          \\
					+ \ours{}     &6.441s\cgaphll{+}{10.2\%}  &4.856s \\
					\bottomrule
 			\end{tabular}}
			\caption{Training and inference time (batch-level) comparison.}
			\label{train}
		\end{minipage}
		\hspace{5pt}
		\begin{minipage}[t]{0.35\linewidth}
			\centering
			\resizebox{0.97  \linewidth}{!}
			{
				\begin{tabular}{y{90}y{60}}
					\toprule
					\textbf{Dataset}   & Imbal. Factor \\
					\midrule
					Cityscapes (point)       &373      \\
					Cityscapes (center)       &\textbf{20} \\
					\midrule
					ScanNet v2 (point)         &116           \\
					ScanNet v2 (center)       &\textbf{11} \\
					\bottomrule
				\end{tabular}
			}
			\caption{Imbalanced factor comparison.}
			\label{if_comp}
		\end{minipage}
	\end{center}
\end{table*}

\subsection{Training Time and Inference Time Analysis}

As we discussed in Sec.~\ref{sec_c2r}, \ours{} can be easily integrated into any off-the-shelf segmentation architecture. 
In the evaluation of our method, we only preserve the point/pixel recognition branch. It means that just the point/pixel classifier is preserved, while the center regularization branch is discarded. Therefore, the evaluation of \ours{} is very efficient. We list the training and inference time results in Table~\ref{train}. All experiments are conducted on the Nvidia GeForce 2080Ti GPU. For DLV3P, we use ResNet-101 as the backbone with a batch of two images. For MinkowskiNet-34, we set the batch size to four. Although our method can greatly improve the imbalance performance and is efficient in inference, the introduction of the additional center regularization branch increases the training time cost by about 10 to 20\%.

\end{document}